\documentclass[utf8]{frontiersSCNS} % for Science, Engineering and Humanities and Social Sciences articles
\usepackage{url,hyperref,lineno,microtype,subcaption}
\usepackage[onehalfspacing]{setspace}
\usepackage[ruled,linesnumbered]{algorithm2e}
\usepackage{natbib}
\usepackage{booktabs}
\usepackage[margin=1in]{geometry}
\usepackage{xcolor, colortbl}
\usepackage{amsmath}
\usepackage{amssymb} 

\DeclareMathOperator*{\argmax}{arg\,max}

\usepackage{graphicx}
\usepackage{amsthm}

\theoremstyle{definition}
\newtheorem{definition}{Definition}[section]
\def\keyFont{\fontsize{8}{11}\helveticabold }
\def\firstAuthorLast{Abhishek Ghose {et~al.}} %use et al only if is more than 1 author
\def\Authors{Abhishek Ghose\,$^{1,*}$, Balaraman Ravindran\,$^{2}$}
\def\Address{$^{1}$Dept. of Computer Science and Engineering, IIT Madras, India \\
$^{2}$Dept. of Computer Science and Engineering, Robert Bosch Centre for Data Science and AI, IIT Madras, India  }
\def\corrAuthor{-}

\def\corrEmail{cs15d004@smail.iitm.ac.in, abhishek.ghose.82@gmail.com}

\begin{document}
\onecolumn
\firstpage{1}

\title[Interpretability with Accurate Small Models]{Interpretability with Accurate Small Models} 

\author[\firstAuthorLast ]{\Authors} %This field will be automatically populated
\address{} %This field will be automatically populated
\correspondance{} %This field will be automatically populated

\extraAuth{}% If there are more than 1 corresponding author, comment this line and uncomment the next one.
%\extraAuth{corresponding Author2 \\ Laboratory X2, Institute X2, Department X2, Organization X2, Street X2, City X2 , State XX2 (only USA, Canada and Australia), Zip Code2, X2 Country X2, email2@uni2.edu}

\maketitle

\begin{abstract}
Models often need to be constrained to a certain size for them to be considered interpretable. For example, a decision tree of depth 5 is much easier to understand than one of depth 50. Limiting model size, however, often reduces accuracy. We suggest a practical technique that minimizes this trade-off between interpretability and classification accuracy. This enables an arbitrary learning algorithm to produce highly accurate small-sized models. Our technique identifies the training data distribution to learn from that leads to the highest accuracy for a model of a given size. 

We represent the training distribution as a combination of sampling schemes.
Each scheme is defined by a parameterized probability mass function applied to the segmentation produced by a decision tree.
An Infinite Mixture Model with Beta components is used to represent a combination of such schemes. The mixture model parameters are learned using Bayesian Optimization.  Under simplistic assumptions, we would need to optimize for $O(d)$ variables for a distribution over a $d$-dimensional input space, which is cumbersome for most real-world data. However, we show that our technique significantly reduces this number to a \emph{fixed set of eight variables} at the cost of relatively cheap preprocessing. 
The proposed technique is flexible: it is \emph{model-agnostic}, i.e.,  it may be applied to the learning algorithm for any model family, and it admits a general notion of model size. We demonstrate its effectiveness using multiple real-world datasets to construct decision trees, linear probability models and gradient boosted models with different sizes. We observe significant improvements in the F1-score in most instances, exceeding an improvement of $100\%$ in some cases.

%%% Leave the Abstract empty if your article does not require one, please see the Summary Table for full details.
%\section{}
%For full guidelines regarding your manuscript please refer to \href{http://www.frontiersin.org/about/AuthorGuidelines}{Author Guidelines}.

%As a primary goal, the abstract should render the general significance and conceptual advance of the work clearly accessible to a broad readership. References should not be cited in the abstract. Leave the Abstract empty if your article does not require one, please see \href{http://www.frontiersin.org/about/AuthorGuidelines#SummaryTable}{Summary Table} for details according to article type. 

\tiny
 \keyFont{ \section{Keywords:} ML, interpretable machine learning, bayesian optimization, infinite mixture models, density estimation} %All article types: you may provide up to 8 keywords; at least 5 are mandatory.
\end{abstract}

\section{Introduction}

As Machine Learning (ML) becomes pervasive in our daily lives, there is an increased desire to know how models reach specific decisions. In certain contexts this might not be important as long as the ML model itself works well, e.g., in product or movie recommendations. But for certain others, such as medicine and healthcare \citep{Caruana:2015:IMH:2783258.2788613, Ustun2016}, banking\footnote{\url{https://blogs.wsj.com/cio/2018/05/11/bank-of-america-confronts-ais-black-box-with-fraud-detection-effort/}}, defence applications\footnote{\url{https://www.darpa.mil/program/explainable-artificial-intelligence}} and law enforcement\footnote{\url{https://www.propublica.org/article/machine-bias-risk-assessments-in-criminal-sentencing}, \url{https://www.propublica.org/article/how-we-analyzed-the-compas-recidivism-algorithm}} model transparency is an important concern.Very soon, regulations governing digital interactions might necessitate interpretability \citep{Goodman2017EuropeanUR}.

All these factors have generated a lot of interest around ``model understanding''. Approaches in the area may be broadly divided into two categories:
\begin{enumerate}
    \item \emph{Interpretability}: build models that are inherently easy to interpret, e.g., rule lists \citep{Letham2013InterpretableCU, Angelino:2017:LCO:3097983.3098047}, decision trees \citep{cart93, Quinlan:1993:CPM:152181, Quinlan:c5}, sparse linear models \citep{Ustun2016}, decision sets \citep{Lakkaraju:2016:IDS:2939672.2939874}, pairwise interaction models that may be linear \citep{Lim2015} or additive \citep{Lou:2013:AIM:2487575.2487579}. %, neural networks that identify snippets that most influence classification \citep{bastings-etal-2019-interpretable}.
    \item \emph{Explainability}: build tools and techniques that allow for explaining black box models, e.g., locally interpretable models such as LIME, Anchors \citep{Ribeiro:2016:WIT:2939672.2939778, AAAI1816982}, visual explanations for Convolutional Neural Networks such as Grad-CAM  \citep{8237336}, influence functions \citep{pmlr-v70-koh17a}, feature attribution based on Shapley values \citep{NIPS2017_7062, pmlr-v97-ancona19a}.
\end{enumerate}

Our work addresses the problem of interpretability by providing a way to increase accuracy of existing models that are considered interpretable. 

Interpretable models are preferably small in \emph{size}: this is referred to as low \emph{explanation complexity} in \citet{ DBLP:journals/corr/abs-1711-07414}, is seen as a form of \emph{simulability} in \citet{Lipton:2018:MMI:3236386.3241340}, is a motivation for \emph{shrinkage methods}  \cite[Section~3.4]{hastie_09_elements-of.statistical-learning}, and is often otherwise listed as a desirable property for interpretable models \citep{Ribeiro:2016:WIT:2939672.2939778, Lakkaraju:2016:IDS:2939672.2939874, Angelino:2017:LCO:3097983.3098047} . For instance, a decision tree of $depth=5$ is easier to understand than one of $depth=50$. Similarly, a linear model with $10$ non-zero terms might be easier to comprehend than one with $50$ non-zero terms. This indicates an obvious problem: an interpretable model is often small in its size, and since model size is usually inversely proportional to the bias, a model often sacrifices accuracy for interpretability.

We propose a technique to minimize this tradeoff for any model family; thus our approach is \emph{model agnostic}. Our technique adaptively samples the provided training data, and identifies a sample on which to learn a model of a given size; the property of this sample being that it is optimal in terms of the accuracy of the constructed model. What makes this strategy practically valuable is that the accuracy of this model may often be significantly higher than one learned on the training data as-is, especially when the model size is small.

Let,
\begin{enumerate}
    \item $accuracy(M, p)$ be the classification accuracy of model $M$ on data represented by the joint distribution $p(X, Y)$ of instances $X$ and labels $Y$. We use the term ``accuracy'' as a generic placeholder for a measure of model correctness. This may specifically measure \emph{F1-score}, \emph{AUC}, \emph{lift}, etc., as needed.    
    \item $train_\mathcal{F}(p, \eta)$ produce a model obtained using a specific training algorithm, e.g., CART \citep{cart93}, for a given model family $\mathcal{F}$, e.g., decision trees, where the model size is fixed at $\eta$, e.g., trees with $depth=5$. The training data is represented by the joint distribution $p(X, Y)$ of instances $X$ and labels $Y$.
\end{enumerate}

If we are interested in learning a classifier of size $\eta$ for data with distribution $p(X,Y)$, our technique produces the \emph{optimal training distribution} $p^*_\eta(X,Y)$ such that:

\begin{align}
\label{eqn:objective1}
     p^*_\eta = \argmax_q accuracy(train_\mathcal{F}(q, \eta), p)
\end{align}
Here $q(X, Y)$ ranges over all possible distributions over the data $(X, Y)$.

Training a model on this optimal distribution produces a model that is at least as good as training on the original distribution $p$:  

\begin{align}
\label{eqn:objective2}
    accuracy(train_\mathcal{F}(p, \eta), p) \leq accuracy(train_\mathcal{F}(p^*_\eta, \eta), p)
\end{align}

Furthermore, the relationship in Equation \ref{eqn:objective2} may be separated into two regimes of operation. A model trained on $p^*_\eta$ outperforms one trained on the original distribution $p$ up to a model size $\eta'$, with both models being comparably accurate beyond this point:

\begin{align}
    \text{For } \eta \leq \eta', accuracy(train_\mathcal{F}(p, \eta), p) < accuracy(train_\mathcal{F}(p^*_\eta, \eta), p) \label{eqn:objective_breakup_a} \\
    \text{For } \eta > \eta', accuracy(train_\mathcal{F}(p, \eta), p) = accuracy(train_\mathcal{F}(p^*_\eta, \eta), p) \label{eqn:objective_breakup_b} 
\end{align}
\label{eqn:size}

% Since our technique relies on sampling\footnote{This approximate relationship is explained in Equation \ref{eqn:approx_label_dist_explained}, Section \ref{sec:preprocess_dt}.}:
% \begin{equation}
%     p(Y|X) \approx p^*_\eta(Y|X), \forall \eta
% \end{equation}
% %    &p(X, Y) \neq p'_\eta(X, Y), \text{ when } \eta \leq \eta^*\\
% %    &p(Y|X) = p'(Y|X), \forall \eta 
% %\end{align}

Our key contributions in this work are:
\begin{enumerate}
    \item Postulating that the optimal training distribution may be different than the test distribution. \textbf{This challenges the conventional wisdom that the training and test data must come from the same distribution}, as in the LHS of Equations \ref{eqn:objective2}, \ref{eqn:objective_breakup_a}, \ref{eqn:objective_breakup_b}. 
    \item Providing a model-agnostic and practical adaptive sampling based technique that exploits this effect to learn small models, that often possess higher accuracy compared to using the original distribution.
    
    \item Demonstrating the effectiveness of our technique with different learning algorithms, $train_\mathcal{F}()$, and multiple real world datasets. Note that our benchmark is not a specific algorithm that learns small models; the value of our approach is in its being \emph{model-agnostic}: it works with arbitrary learners.

    \item We show that learning the distribution, $p^*_\eta$, in the $d$ dimensions of the data, may be decomposed into a relatively cheap preprocessing step that depends on $d$, followed by a core optimization step \emph{independent} of $d$: the optimization is over a \textbf{fixed set of eight variables}. This makes our technique scalable. %Contrast this with a standard distribution learning approach like \emph{Gaussian Mixture Models (GMM)}, which needs to optimize over at least $O(d)$ variables even under strong simplifying assumptions.

\end{enumerate}

We do not impose any constraints on the specification of $train_\mathcal{F}()$ for it to create interpretable models; our technique may be used with any model family. But the fact that we see increased accuracy \emph{up to} a model size ($\eta'$ in Equation \ref{eqn:objective_breakup_a}), makes the technique \emph{useful} in setups where small sized models are preferred. Applications requiring interpretability are an example of this. There may be others, such as \emph{model compression}, which we have not explored, but briefly mention in Section \ref{ssec:extn}.

%Note that our benchmark is not a specific algorithm that can learn small models; the value of our approach is in its \emph{generality}: it works with arbitrary learners, and as such we may categorize it as a ``wrapper algorithm''. Some other examples within this category are various commonly used methods for feature selection \citep{John:1994:IFS:3091574.3091590, Kohavi:1997:WFS:270613.270627, wrapper_review7745366} and \emph{MetaCost} \citep{Domingos:1999:MGM:312129.312220}, a method to make classifiers cost-sensitive.

\section{Overview}
This section provides an overview of various aspects of our work: we impart some intuition for why we expect the train and test distributions to differ for small-sized models, describe where our technique fits into a model building workflow, mention connections to previous work and establish our notation and terminology.
\subsection{Intuition}
\label{ssec:intuition}
\begin{figure}[h]
    \centering
\includegraphics[height=0.5\textwidth]{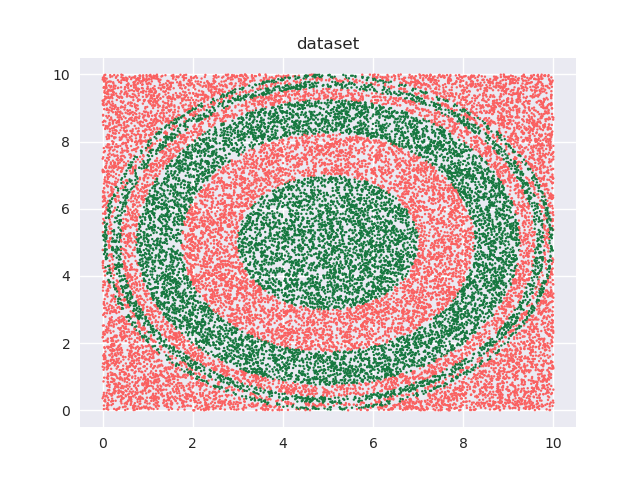}
    \caption{Binary class dataset for classification.}
    \label{fig:circle_dataset}
\end{figure}

Let's begin with a quick demonstration of how modifying the training distribution can be useful. We have the binary class data, shown in Figure \ref{fig:circle_dataset}, that we wish to classify with decision trees with $depth = 5$. %We refer to the data distribution provided as the \emph{original distribution}. %We always evaluate our models on \emph{test} datasets (and \emph{validation} datasets, where applicable) drawn from the original distribution.

\begin{figure}[ht]
    \centering
\includegraphics[scale=0.38]{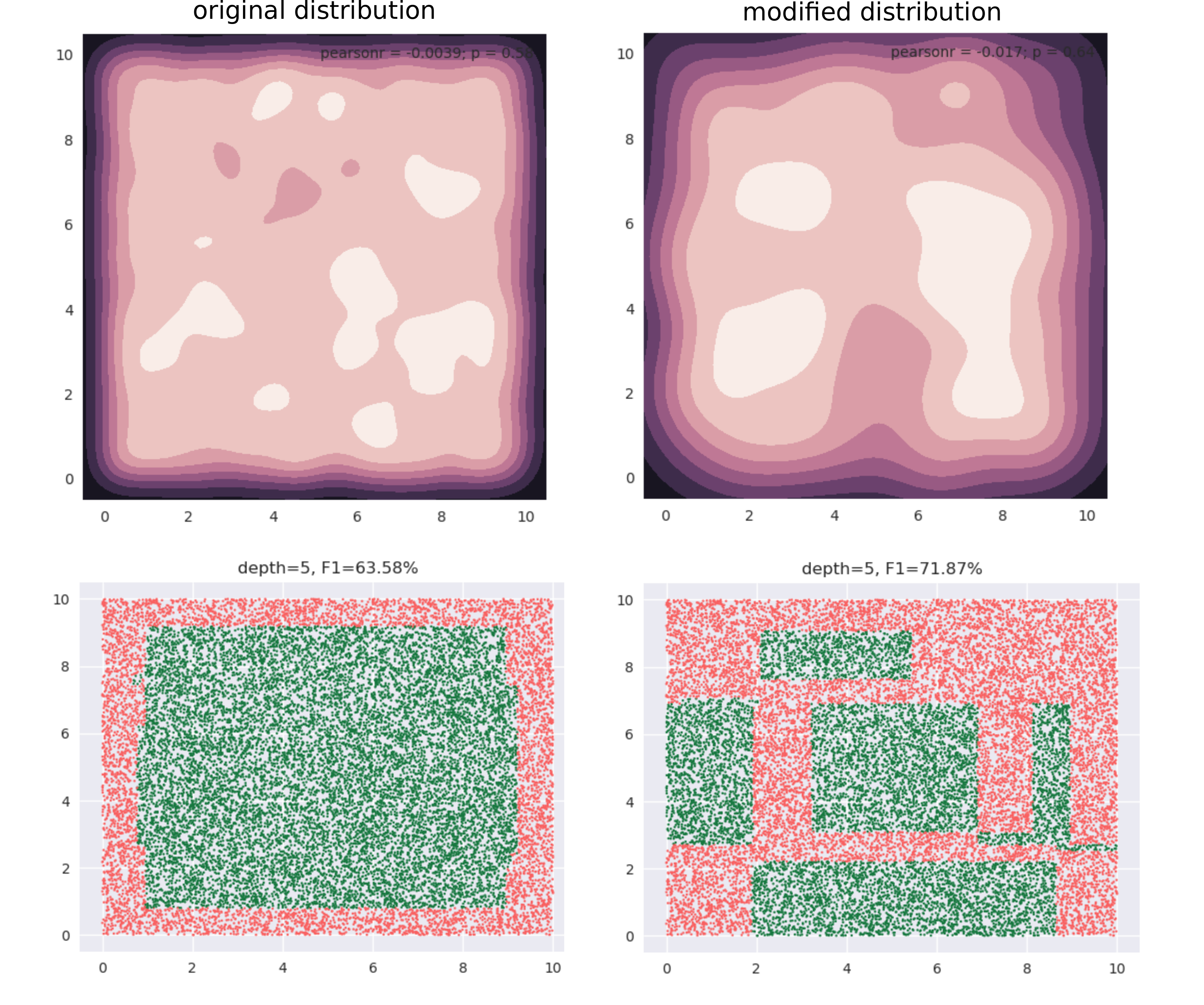}
    \caption{Changing the input distribution can significantly affect model accuracy.}
    \label{fig:CART_diff_densities}
\end{figure}

Our training data is a subset of this data (not shown). The training data is approximately uniformly distributed in the input space - see the 2D \emph{kernel density} plot in the top-left panel in Figure \ref{fig:CART_diff_densities}. The bottom-left panel in the figure shows the regions of a decision tree with $depth=5$ learns, using the CART algorithm. The top-right panel shows a modified distribution of the data (now the density seems to be relatively concentrated away from the edge regions of the input space), and the corresponding decision tree with $depth=5$, also learned using CART, is visualized in the bottom-right panel. Both decision trees used the same learning algorithm and possess the same depth. As we can see, the $F1$ scores are significantly different: $63.58\%$ and $71.87\%$ respectively.  

Where does this additional accuracy come from?

All classification algorithms use some heuristic to make learning tractable, e.g.:
\begin{itemize}
\item Decision Trees - one step lookahead (note that the CART tree has a significantly smaller number of leaves than the possible $2^5=32$, in our example).
\item Logistic Regression - local search, e.g., \emph{Stochastic Gradient Descent (SGD)}.
\item Artificial Neural Networks (ANN) - local search, e.g., SGD, \emph{Adam}. 
\end{itemize}

Increasing the size allows for offsetting the shortcomings of the heuristic by adding parameters to the model till it is satisfactorily accurate: increasing depth, terms, hidden layers or nodes per layer.
Our hypothesis is, in restricting a model to a small size, this potential gap between the \emph{representational} and \emph{effective} capacities becomes pronounced.
%Restricting the size exposes this gap between the \emph{effective} and \emph{representational} capacities, when it exists. 
%This gives us room to squeeze in additional performance. We achieve this by changing the density in a way to better focus the small model on regions of the input space that are most valuable in terms of accuracy.
%Changing the density in a way to better focus the small model on regions of the input space that are most valuable in terms of accuracy.
In such cases, modifying the data distribution guides the heuristic to focus learning on regions of the input space that are valuable in terms of accuracy. We are able to empirically demonstrate this effect for DTs in Section \ref{sssec:dt_results}.

\subsection{Workflow}

\begin{figure}[ht]
    \centering
\includegraphics[scale=0.45]{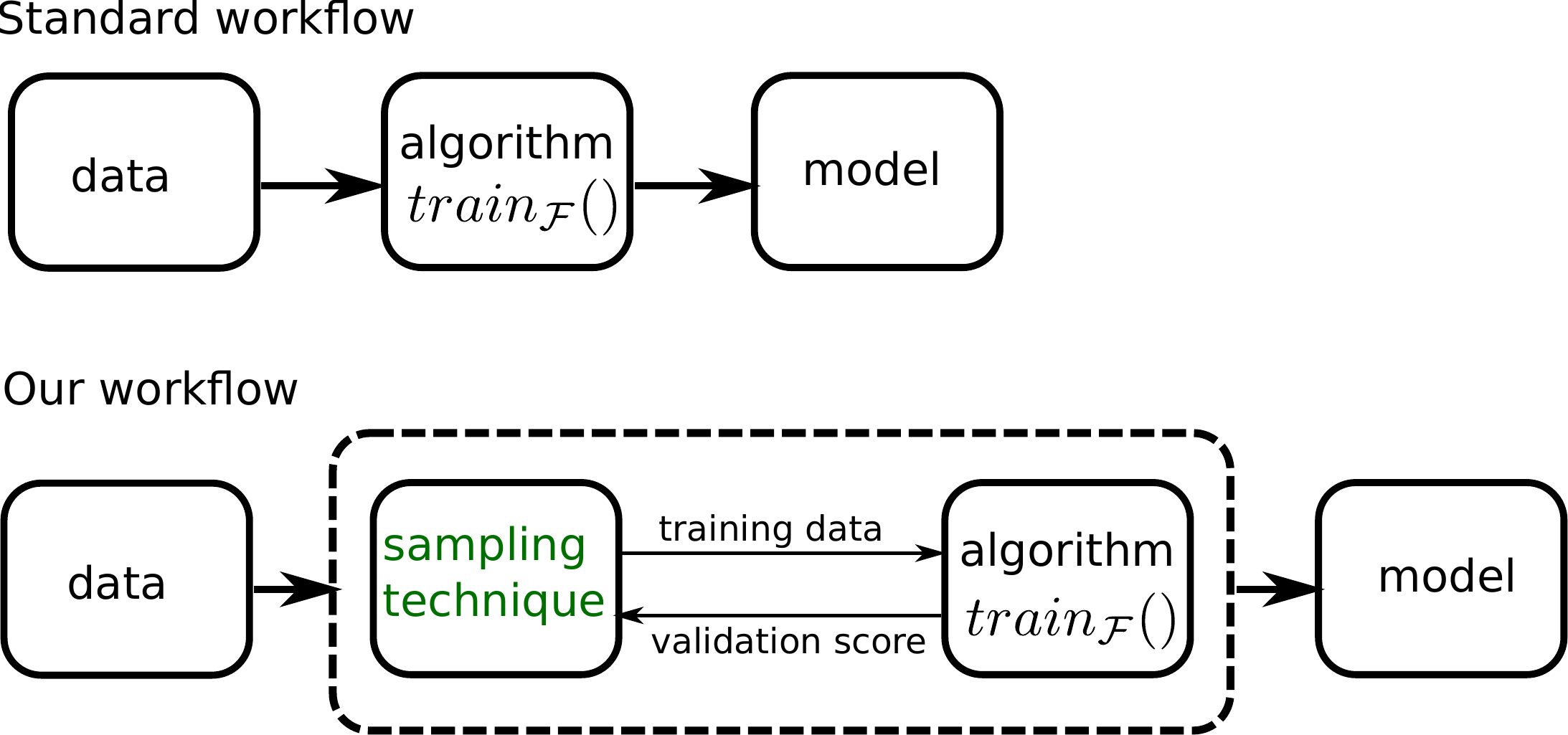}
    \caption{Our workflow compared with the standard workflow.}
    \label{fig:flowchart}
\end{figure}

Figure \ref{fig:flowchart} shows how our sampling technique modifies the model building workflow. In the standard workflow, we feed the data into a learning algorithm, $train_{\mathcal{F}}()$, to obtain a model. In our setup, the data is presented to a system, represented by the dashed box, that is comprised of both the learning algorithm and our sampling technique.

This system produces the final model in an iterative fashion: the sampling technique (or \emph{sampler}) produces a sample using its current distribution parameters, that is used by the learning algorithm to produce a model. This model is evaluated on a validation dataset and the validation score is conveyed back to the sampler. This information is used to modify the distribution parameters and generate a new training sample for the algorithm, and so on, till we reach a stopping criteria. The criteria we use is a specified number of iterations - we refer to this as our \emph{budget}. The best model produced within the budget, as measured by the validation score, is our final model, and the corresponding distribution is presented as the ideal training distribution.

\subsection{Previous Work}

We are aware of no prior work that studies the relationship between data distribution and accuracy in the small model regime. In terms of the larger view of modifying the training distribution to influence learning, parallels may be drawn to the following methodologies:

\begin{enumerate}
    \item When learning on data with class imbalance, using a different train distribution compared to test via over/under-sampling \citep{Japkowicz:2002:CIP:1293951.1293954}, is a commonly used strategy. Seen from this perspective, we are positing that modifying the original distribution is helpful in a wider set of circumstances, i.e., when there is no imbalance, as in Figure \ref{fig:circle_dataset}, but the model is restricted in size.
    \item Among popular techniques, \emph{Active Learning} \citep{settles2009active, Dasgupta:2011:TFA:1959886.1960197} probably bears the strongest resemblance to our approach. However, our problem is different in the following key respects:
    \begin{enumerate}
        \item In active learning, we don't know the labels of most or all of the data instances, and there is an explicit label acquisition cost that must be accounted for. In contrast, our work targets the traditional supervised learning setting where the joint distribution of instances and labels is approximately known through a fixed set of samples drawn from that distribution.
        \item Because there is a label acquisition cost, learning from a small subset of the data such that the resulting model approximates one learned on the complete dataset, is strongly incentivized. This economy in sample size is possibly the most common metric used to evaluate the utility of an active learner. This is different from our objective, where we are not interested in minimizing training data size, but in learning small-sized models. Further, we are interested in \emph{outperforming} a model learned on the complete data.
        
    \end{enumerate}
    \item \emph{Coreset} construction techniques \citep{2017arXiv170306476B,Munteanu2018} seek to create a ``summary'' weighted sample of a dataset with the property that a model learned on this dataset approximates one learned on the complete dataset. Here too, the difference in objectives is that we focus on small models, ignore training data size, and are interested in outperforming a model learned on the complete data. 
\end{enumerate}

This is not to say that the tools of analysis from the areas of active learning or coreset identification cannot be adapted here; but current techniques in these areas do not solve for our objective.

%Using a distribution different from the one provided to learn from, is not unprecedented. We see this commonly in the case of learning on data with class imbalance where over/under-sampling is used \citep{Japkowicz:2002:CIP:1293951.1293954}. Seen from this perspective, we are positing that modifying the original distribution is helpful in a wider set of circumstances. 

%We are aware of no prior work that studies the relationship between data density and accuracy in the small model regime. In terms of the larger view of modifying density to influence learning, parallels may be drawn to \emph{active learning} \citep{settles2009active}, and the identification of \emph{coresets} \citep{2017arXiv170306476B,Munteanu2018}.

\subsection{Terminology and Notation}
Let's begin with the notion of ``model size''.
Even though there is no standard notion of size across model families, or even within a model family, we assume the term informally denotes model attribute(s) with the following properties:
\begin{enumerate}
    \item $size \propto bias^{-1}$
    \item Smaller the size of a model, easier it is to interpret.
\end{enumerate}
As mentioned earlier, only property 1 is strictly required for our technique to be applicable; property 2 is needed for interpretability.

Some examples of model size are depth of decision trees, number of non-zero terms in a linear model and number of rules in a rule set. 

In practice, a model family may have \emph{multiple} notions of size depending upon the modeler, e.g., depth of a tree or the number of leaves. The size might even be determined by multiple attributes in conjunction, e.g., maximum depth of each tree \emph{and} number of boosting rounds in the case of a \emph{gradient boosted model (GBM)}. It is also possible that while users of a model might agree on a definition of size they might disagree on the \emph{value} for the size up to which the model stays interpretable. For e.g., are decision trees interpretable up to a depth of $5$ or $10$? Clearly, the definition of size and its admissible values might be subjective. Regardless, the discussion in this paper remains valid as long as the notion of size exhibits the properties above. With this general notion in mind, we say that interpretable models are typically \emph{small}.

Here are the notations we use:
\begin{enumerate}
    \item The matrix $X \in \mathbb{R}^{N \times d}$ represents an ordered collection of $N$ input feature vectors, each of which has $d$ dimensions. We assume individual feature vectors $x_i \in \mathbb{R}^{d \times 1}$ to be column vectors, and hence the $i^{th}$ row of $X$ represents $x_i^T$. We occasionally treat $X$ as a set and write $x_i \in X$ to denote the feature vector $x_i$ is part of the collection $X$.
    
    An ordered collection of $N$ labels is represented by the vector $Y \in \mathbb{R}^N$.
    
    We represent a dataset with $N$ instances with the tuple $(X, Y)$, where $X \in \mathbb{R}^{N \times d}$,  $Y \in \mathbb{R}^N$, and the label for $x_i$ is $Y_i, \text{ where } 1 \leq i \leq N$.
    
    \item The element at the $p^{th}$ row and $q^{th}$ column indices of a matrix $A$ is denoted by $[A]_{pq}$.
    
    \item We refer to the joint distribution $p(X, Y)$ from which a given dataset was sampled, as the \emph{original distribution}. In the context of learning a model and predicting on a held-out dataset, we distinguish between the \emph{train}, \emph{validation} and \emph{test} distributions. In this work, the train distribution may or may not be identical to the original distribution, which would be made clear by the context, but the validation and test distributions are \emph{always} identical to the original distribution.
    \item The terms \emph{pdf} and \emph{pmf} denote \emph{probability density function} and \emph{probability mass function} respectively. The term ``probability distribution'' may refer to either, and is made clear by the context. A distribution $p$, parameterized by $\theta$, defined over the variable $x$, is denoted by $p(x; \theta)$.
    \item 
    We use the following terms introduced before:
    \begin{itemize}
        \item $accuracy(M, p)$ is the classification accuracy of model $M$ on data represented by the joint distribution $p(X, Y)$ of instances $X$ and labels $Y$. We often overload this term to use a dataset instead of distribution. In this case, we write $accuracy(M, (X,Y))$ where $(X, Y)$ is the dataset.
        \item $train_\mathcal{F}(p, \eta)$ produces a model obtained using a specific training algorithm for a model family $\mathcal{F}$, where the model size is fixed at $\eta$. This may also be overloaded to use a dataset, and we write: $train_{\mathcal{F}}((X, Y), \eta)$.
    \end{itemize}
    
    \item We denote the depth of a tree $T$ by the function $depth(T)$.
    \item $\mathbb{R}$, $\mathbb{Z}$ and $\mathbb{N}$ denote the sets of \emph{reals}, \emph{integers} and \emph{natural numbers} respectively. 
\end{enumerate}

\qed

The rest of the paper is organized as follows: in Section \ref{sec:methodology} we describe in detail two formulations of the problem of learning the optimal density. Section \ref{sec:expts} reports experiments we have conducted to evaluate our technique. It also presents our analysis of the results. We conclude with Section \ref{sec:discuss} where we discuss some of the algorithm design choices and possible extensions of our technique.

\section{Methodology}
\label{sec:methodology}

In this section we describe our sampling technique. We begin with a intuitive formulation of the problem in Section \ref{ssec:naive} to illustrate challenges with a simple approach. This also allows us to introduce the relevant mathematical tools. Based on our understanding here, we propose a much more efficient approach in Section \ref{sec:preprocess_dt}.

\subsection{A Naive Formulation}
\label{ssec:naive}
We phrase the problem of finding the ideal density (for the learning algorithm) as an optimization problem. We represent the density over the input space with the \emph{pdf} $p(x;\Psi)$, where $\Psi$ is a parameter vector. Our optimization algorithm runs for a budget of $T$ time steps. %We loosely express sampling from $p(x; \Psi)$ as $x \sim \Psi$. 
Algorithm \ref{alg:naive_soln} lists the execution steps.

\SetKw{KwBy}{by}
\begin{algorithm}
 \KwData{Learning algorithm $train_{\mathcal{F}}()$, size of model $\eta$, data $(X,Y)$, iterations $T$}
 \KwResult{Optimal density $\Psi^*$, accuracy on test set $s_{test}$ }
Create stratified subsets $(X_{train}, Y_{train}), (X_{val}, Y_{val}), (X_{test}, Y_{test})$ from $(X,Y)$\;
\For {$t\gets1$ \KwTo $T$}{
   $\Psi_{t} \gets suggest(s_{t-1}, ... s_1, \Psi_{t-1}, ..., \Psi_1)$\ \tcp{$suggest()$ is described below} 
   $(X_t, Y_t) \gets$ sample $N_s$ points from $(X_{train}, Y_{train})$ based on $p(X_{train}; \Psi_t)$\;
   %Let $w(x_i) \gets \Psi_t(x_i)$ be the sampling weight for $(x_i, y_i) \in (X_{train}, Y_{train})$\;
  %$(X_t, Y_t) \gets$ sample $N$ points with replacement from $X_{train}$, using $w(x_i)$\;  
  %$(X_t, Y_t) \sim \Psi_{t}$\;
  $M_t \gets train_{\mathcal{F}}((X_t, Y_t), \eta) $ \;
  $s_t \gets accuracy(M_t, (X_{val}, Y_{val}))$ \;
 }
$t^* \gets \argmax_t{\{s_1, s_2, ..., s_{T-1}, s_T\}}$\;
$\Psi^* \gets \Psi_{t^*}, M^* \gets M_{t^*}$\;
%$(X^*, Y^*) \gets$ sample $N_s$ points from $(X_{train}, Y_{train})$ based on $p(X_{train}; \Psi^*)$\;
%$M^* \gets train_{\mathcal{F}}((X^*, Y^*), \eta) $ \;
$s_{test} \gets accuracy(M^*,(X_{test}, Y_{test}))$\;
\Return $\Psi^*$, $s_{test}$
\caption{Naive formulation}
\label{alg:naive_soln}
\end{algorithm}

In Algorithm \ref{alg:naive_soln}:
\begin{enumerate}
    %\item Our objective function is denoted by $accuracy()$. This may specifically measure \emph{F1-score}, \emph{AUC}, \emph{lift} etc as needed. It is evaluated on the \emph{validation set} $X_{val}$. The result of a call to $accuracy()$ is represented by $s$.  
    %\item $train()$ denotes the learning algorithm we are interested in, e.g., a decision tree or a sparse linear model learner.
    \item $suggest()$ is a call to the optimizer at time $t$, that accepts past validation scores $s_{t-1}, ... s_1$ and values of the density parameter $\Psi_{t-1}, ..., \Psi_1$. These values are randomly initialized for $t=1$. Note that not all optimizers require this information, but we refer to a generic form of optimization that makes use of the entire history.
    \item In Line 4, a sampled dataset $(X_t, Y_t)$ comprises of instances $x_i \in X_{train}$, and their corresponding labels $y_i \in Y_{train}$. Denoting the sampling weight of an instance $x_i$ as $w(x_i)$, we use $w(x_i) \propto p(x_i;\Psi_t), \forall x_i \in X_{train}$.  
    
    The sampling in Line 10 is analogous.
    \item Although the training happens on a sample drawn based on $\Psi_t$, the validation dataset $(X_{val}, Y_{val})$ isn't modified by the algorithm and always reflects the original distribution. Hence, $s_t$ represents the accuracy of a model on the original distribution. 
    %\item For a sample $X_t$, where do the $Y_t$ come from? For now, we assume that we have kept aside a training set $(X_{train}, Y_{train})$, from which we draw samples based on $p(x;\Psi_t)$.
    \item In the interest of keeping the algorithm simple to focus on the salient steps/challenges, we defer a discussion of the sample size $N_s$ to our improved formulation in Section \ref{sec:preprocess_dt}.
\end{enumerate}

Algorithm \ref{alg:naive_soln} represents a general framework to discover the optimal density within a time budget $T$. We refer to this as a ``naive'' algorithm, since within our larger philosophy of discovering the optimal distribution, this is the most direct way to do so. It uses $accuracy()$ as both the objective and fitness function, where the score $s_t$ is the fitness value for current parameters $\Psi_t$. It is easy to see here what makes our technique model-agnostic: the arbitrary learner $train_{\mathcal{F}}()$ helps define the fitness function but there are no assumptions made about its form.  While conceptually simple, clearly the following key implementation aspects dictate its usefulness in practice:
\begin{enumerate}
    \item The optimizer to use for $suggest()$.
    \item The precise representation of the $pdf$ $p(x;\Psi)$.
\end{enumerate}

We look at these next.

\subsubsection{Optimization}
\label{sec:opt}
The fact that our objective function is not only a black-box, but is also noisy, makes our optimization problem hard to solve, especially within a budget $T$. The quality of the optimizer $suggest()$ critically influences the utility of Algorithm \ref{alg:naive_soln}.

We list below the characteristics we need our optimizer to possess:

\begin{enumerate}
        \item \textbf{Requirement 1: it should be able to work with a black-box objective function.} Our objective function is $accuracy()$, which depends on a model produced by $train_{\mathcal{F}}()$. The latter is an input to the algorithm and we make no assumptions about its form. The cost of this generality is that $accuracy()$ is a black-box function and our optimizer needs to work without knowing its smoothness, amenability to gradient estimation etc.
        
        \item \textbf{Requirement 2: should be robust against noise.} Results of $accuracy()$ may be noisy. There are multiple possible sources of noise, e.g.: 
        \begin{enumerate}
            \item The model itself is learned on a sample $(X_t, y_t)$.
            \item The classifier might use a local search method like SGD whose final value for a given training dataset depends on various factors like initialization, order of points, etc.
        \end{enumerate}
        
         \item \textbf{Requirement 3: minimizes calls to the objective function.} The acquisition cost of a fitness value $s_t$ for a solution $\Psi_t$ is high: this requires a call to $accuracy()$, which in turn calls $train_{\mathcal{F}}()$. Hence, we want the optimizer to minimize such calls, instead shifting the burden of computation to the optimization strategy. The number of allowed calls to $accuracy()$  is often referred to as the \emph{fitness evaluation budget}.
\end{enumerate}

Some optimization algorithms that satisfy the above properties to varying degrees are the class of \emph{Bayesian Optimization (BO)} \citep{Brochu2010ATO,7352306} algorithms; evolutionary algorithms such as \emph{Covariance Matrix Adaptation Evolution Strategy (CMA-ES)}  \citep{Hansen:2001:CDS:1108839.1108843, hansen2004ecm} and \emph{Particle Swarm Optimization (PSO)} \citep{PSO, Parsopoulos01particleswarm}; heuristics based algorithms such as \emph{Simulated Annealing} \citep{Kirkpatrick1983, Gelfand1989SimulatedAW, Gutjahr1996}; bandit-based algorithms such as \emph{Parallel Optimistic Optimization} \citep{NIPS2015_5721} and \emph{Hyperband} \citep{Li:2017:HNB:3122009.3242042}. 

We use BO here since it has enjoyed substantial success in the area of \emph{hyperparameter optimization}, e.g., \citet{Bergstra:2011:AHO:2986459.2986743, NIPS2012_4522, NIPS2018_7917, pmlr-v97-dai19a}, where the challenges are similar to ours. %It is also an active area of research.

While a detailed discussion of BO techniques is beyond the scope of this paper (refer to \cite{Brochu2010ATO, 7352306} for an overview), we briefly describe why they meet our requirements: BO techniques build their own model of the response surface over multiple evaluations of the objective function; this model serves as a \emph{surrogate} (whose form is known) for the actual black-box objective function. The BO algorithm relies on the surrogate alone for optimization, bypassing the challenges in directly working with a black-box function (Requirement 1 above). %The model gets better with successive iterations of the BO algorithm as an increasing number of evaluations help refine the surrogate to better approximate the original objective function. 
The surrogate representation is also probabilistic; this helps in quantifying uncertainties in evaluations, possibly arising due to noise, making for robust optimization (Requirement 2). Since every call to $suggest()$ is informed by this model, the BO algorithm methodically focuses on only the most promising regions in the search space, making prudent use of its fitness evaluation budget (Requirement 3).
%For more details refer \cite{Brochu2010ATO}.

%Additionally, BO algorithms can naturally work with \emph{box constraints} - as we shall see in the next section, this works well within our setup.

The family of BO algorithms is fairly large and continues to grow \citep{Hutter:2011:SMO:2177360.2177404, Bergstra:2011:AHO:2986459.2986743, NIPS2012_4522, Wang:2013:BOH:2540128.2540383, Gelbart:2014:BOU:3020751.3020778, Snoek:2015:SBO:3045118.3045349, Hernandez-Lobato:2016:GFC:2946645.3053442, pmlr-v70-rana17a, Levesque2017BayesianOF, ijcai2017-291, BO_noisy, abo_NIPS2018_7838, NIPS2018_7917, pmlr-v97-nayebi19a, pmlr-v97-alvi19a, pmlr-v97-dai19a}. We  use the \emph{Tree Structured Parzen Estimator (TPE)} algorithm \citep{Bergstra:2011:AHO:2986459.2986743} since it scales linearly with the number of evaluations (the runtime complexity of a naive BO algorithm is \emph{cubic} in the number of evaluations - see \citet{7352306}) and has a popular and mature library: \emph{Hyperopt} \citep{Bergstra:2013:MSM:3042817.3042832}. %, and we observed it to be reasonably fast in our experiments co.

\subsubsection{Density Representation}
\label{sec:density}
The representation of the $pdf$, $p(x;\Psi)$ is the other key ingredient in Algorithm \ref{alg:naive_soln}. The characterestics we are interested in are:
\begin{enumerate}
\item \textbf{Requirement 1: It must be able to represent an arbitrary density function}. This is an obvious requirement since we want to \emph{discover} the optimal density.
\item \textbf{Requirement 2: It must have a fixed set of parameters}. This is for convenience of optimization, since most optimizers cannot handle the \emph{conditional parameter spaces} that some $pdf$ representations use. A common example of the latter is the popular \emph{Gaussian Mixture Model (GMM)}, where the number of parameters increases linearly with the number of mixture components.   

This algorithm design choice allows for a larger scope of being able to use different optimizers in Algorithm \ref{alg:naive_soln}; there are many more optimizers that can handle fixed compared to conditional parameter spaces. And an optimizer that works with the latter, can work with a fixed parameter space as well.
\footnote{The optimizer we use, TPE, \emph{can} handle conditional spaces. However, as mentioned, our goal is flexibility in implementation.}
%assuming $k$ components in $d$ dimensions, and allowing the covariance matrix to be an identity matrix, we have $O(kd)$ density parameters; thus, the total number of parameters depends on the value of $k$. 

\end{enumerate}

The \emph{Infinite Gaussian Mixture Model (IGMM)} \citep{Rasmussen:1999:IGM:3009657.3009736}, a \emph{non-parametric Bayesian} extension to the standard GMM, satisfies these criteria. It side-steps the problem of explicitly denoting the number of components by representing it using a \emph{Dirichlet Process (DP)}. The DP is characterized by a \emph{concentration parameter} $\alpha$, which determines both the number of components (also known as \emph{partitions} or \emph{clusters}) and association of a data point to a specific component. The parameters for these components are not directly learned, but are instead drawn from prior distributions; the parameters of these prior distributions comprises our fixed set of variables (Requirement 2). We make the parameter $\alpha$ part of our optimization search space, so that the appropriate number of components maybe discovered; this makes our $pdf$ flexible (Requirement 1).

We make a few modifications to the IGMM for it to better fit our problem. This doesn't change its compatibility to our requirements. Our modifications are:
\begin{enumerate}
    \item Since our data is limited to a ``bounding box'' within $\mathbb{R}^d$ (this region is easily found by determining the \emph{min} and \emph{max} values across instances in the provided dataset, for each dimension, ignoring outliers if needed), we replace the Gaussian mixture components with a multivariate generalization of the $Beta$ distribution. We pick $Beta$ since it naturally supports bounded intervals. 
    In fact, we may treat the data as lying within the unit hypercube $[0,1]^d$ without loss of generality, and with the understanding that the features of an instance are suitably scaled in the actual implementation.
    
    Using a bounded interval distribution provides the additional benefit that we don't need to worry about infeasible solution regions in our optimization.  
    
    \item Further, we assume \emph{independence} across the $d$ dimensions as a starting point. We do this to minimize the number of parameters, similar to using a diagonal covariance matrix in GMMs. 
    
    Thus, our $d$-dimensional generalization of the $Beta$ is essentially a set of $d$ $Beta$ distributions, and every component in the mixture is associated with such a set. For $k$ mixture components, we have $k \times d$ $Beta$ distributions in all, as against $k$ $d$-dimensional Gaussians in an IGMM.  
    
    \item A $Beta$ distribution uses two positive valued \emph{shape parameters}. Recall that we don't want to learn these parameters for each of the $k \times d$ $Beta$ distributions (which would defeat our objective of a fixed parameter space); instead we sample these from prior distributions. We use $Beta$ distributions for our priors too: each shape parameter is drawn from a corresponding prior $Beta$ distribution. 
    
    Since we have assumed that the dimensions are independent, we have two prior $Beta$ for the shape parameters \emph{per dimension}. We obtain the parameters $\{A_j, B_j\}$ of a $Beta$ for dimension $j, 1 \leq j \leq d$, by drawing $A_j \sim Beta(a_j, b_j)$ and $B_j \sim Beta(a'_j, b'_j)$, where $\{a_j, b_j\}$ and $\{a'_j, b'_j\}$ are the shape parameters of the priors.
    
    There are a total of $4d$ prior parameters, with $4$ prior parameters $\{a_j, b_j, a'_j, b'_j\}$ per dimension $j, 1 \leq j \leq d$.
    
\end{enumerate}

 We refer to this mixture model as an \emph{Infinite Beta Mixture Model (IBMM)}\footnote{We justify this name by noting that there is more than one multivariate generalization of the $Beta$: the \emph{Dirichlet distribution} is a popular one, but there are others, e.g., \cite{article_alt_beta}}. 
For $d$ dimensional data, we have $\Psi = \{\alpha, a_1, b_1, a'_1, b'_1,  ..., a_d, b_d, a'_d, b'_d\}$. This is a total of $4d+1$ parameters. 
 %Accounting for the concentration parameter, $\alpha$, and the prior parameters, we have a total of $4d+1$ parameters. 

Algorithm \ref{alg:sample_ibmm_naive} shows how we sample $N_t$ points from $(X,Y)$ using the IBMM.

\SetKw{KwBy}{by}
\begin{algorithm}
 \KwData{number of points to sample $N_s$, dataset $(X,Y), X \in \mathbb{R}^{N \times d}, Y \in \mathbb{R}^N$}
 \KwResult{$(X_t,Y_t)$, $X_t \in \mathbb{R}^{N_s \times d}, Y_t \in \mathbb{R}^{N_s}$}
 $X_t=[\;], Y_t=[\;]$\;
 $\{(c_1, n_1), (c_2, n_2), ..., (c_k, n_k)\} \gets $ partition $N_s$ using the $DP$\tcp{Here $\sum_{i=1}^k n_i=N_s$.}
 \For {$i\gets1$ \KwTo $k$}{
    \tcp{Get the $Beta$ parameters for component $c_i$}
    \For {$j\gets1$ \KwTo $d$}{
        $A_{ij} \sim Beta(a_j, b_j)$\;
        $B_{ij} \sim Beta(a'_j, b'_j)$\;
    }
    \For {$l\gets1$ \KwTo $N$}{
        $p(x_l|c_i) \gets \prod_{j=1}^d Beta(x_{lj}|A_{ij},B_{ij})$ \;
    }
    $X_{ti} \gets$ sample $n_i$ points from $X$ based on $p(x_l|c_i)$\;
    $Y_{ti} \gets$ labels corresponding to $X_{ti}$ from $Y$\;
    $X_t \gets \begin{bmatrix} X_t\\ X_{ti} \end{bmatrix}, Y_t \gets \begin{bmatrix} Y_t\\ Y_{ti} \end{bmatrix}$\;
 }
 
\Return $(X_t, Y_t)$
\caption{Sampling using IBMM}
\label{alg:sample_ibmm_naive}
\end{algorithm}

We first determine the partitioning of the number $N_s$, induced by the $DP$ (line 2). We use \emph{Blackwell-MacQueen} sampling \citep{blackwell1973} for this step. This gives us $k$ components, denoted by $c_i, 1\leq i \leq k$, and the corresponding number of points $n_i, 1\leq i \leq k$ to be assigned to each component. We then sample points one component at a time: we draw the $Beta$ parameters per dimension - $A_{ij}, B_{ij}$ - from the priors (lines 4-6), followed by constructing sampling weights $p(x_l|c_i), \forall x_l \in X$ assuming independent dimensions (line 9).

We emphasize here that we use the IBMM \emph{purely for representational convenience}. All the $4d+1$ parameters are learned by the optimizer, and we ignore the standard associated machinery for estimation or inference. These parameters \emph{cannot} be learned from the data since our fundamental hypothesis is that the optimal distribution is different from the original distribution. 

%For further details about this form of the \emph{pdf}, the interested reader is referred to \citet{Rasmussen:1999:IGM:3009657.3009736}.

\subsection{Challenges}
The primary challenge with this formulation is the size of the search space. 
We have successfully tried out Algorithm \ref{alg:naive_soln} on small toy datasets as proof-of-concept, but for most real world datasets, optimizing over $4d+1$ variables leads to an impractically high run-time even using a fast optimizer like TPE.

One could also question the independence assumption for dimensions, but that doesn't address the problem of the number of variables: learning a $pdf$ directly in $d$ dimensions would require \emph{at least} $O(d)$ optimization variables. In fact, a richer assumption makes the problem worse with $O(d^2)$ variables to represent inter-dimension interactions.

%Also the component parameters themselves come from Beta priors.
%A \emph{Infinite beta Mixture Model (IBMM)} achieves this. This is our minor modification of the standard \emph{Infinite Gaussian Mixture Model (IGMM)} \parencite{Rasmussen:1999:IGM:3009657.3009736}. We replace the Gaussians with Beta distributions. Also the component parameters themselves come from Beta priors.
%There are few standard ways to derive the number of components 

\subsection{An Efficient Approach using Decision Trees}
\label{sec:preprocess_dt}
%Clearly, the key problem to address here is the size of the parameter search space. 

We begin by asking if we can prune the search space in some fashion. 
Note that we are solving a \emph{classification} problem, measured by $accuracy()$; however the IBMM only indirectly achieves this goal by searching the complete space $\Psi$. The search presumably goes through distributions with points from only one class, no points close to any or most of the class boundary regions, etc; distributions that decidedly result in poor fitness scores. Is there a way to exclude such ``bad'' configuration values from the search space?

One strategy would be to first determine where the class boundaries lie, and \emph{penalize} any density $\Psi_t$ that doesn't have at least some overlap with them. This is a common optimization strategy used to steer the search trajectory away from bad solutions. However, implementation-wise, this leads to a new set of challenges:
\begin{enumerate}
    \item How do we determine, and then represent, the location of class boundaries?
    \item What metric do we use to appropriately capture our notion of overlap of $\Psi_t$ and these locations?
    \item How do we efficiently execute the previous steps? After all, our goal is to either (a) reduce the number of optimization variables OR (b) significantly reduce the size of the search space for the current $O(d)$ variables.
\end{enumerate}

We offer a novel resolution to these challenges that leads to an efficient algorithm by making the optimization ``class boundary sensitive''.

\begin{figure}[h]
    \centering
\includegraphics[scale=0.27]{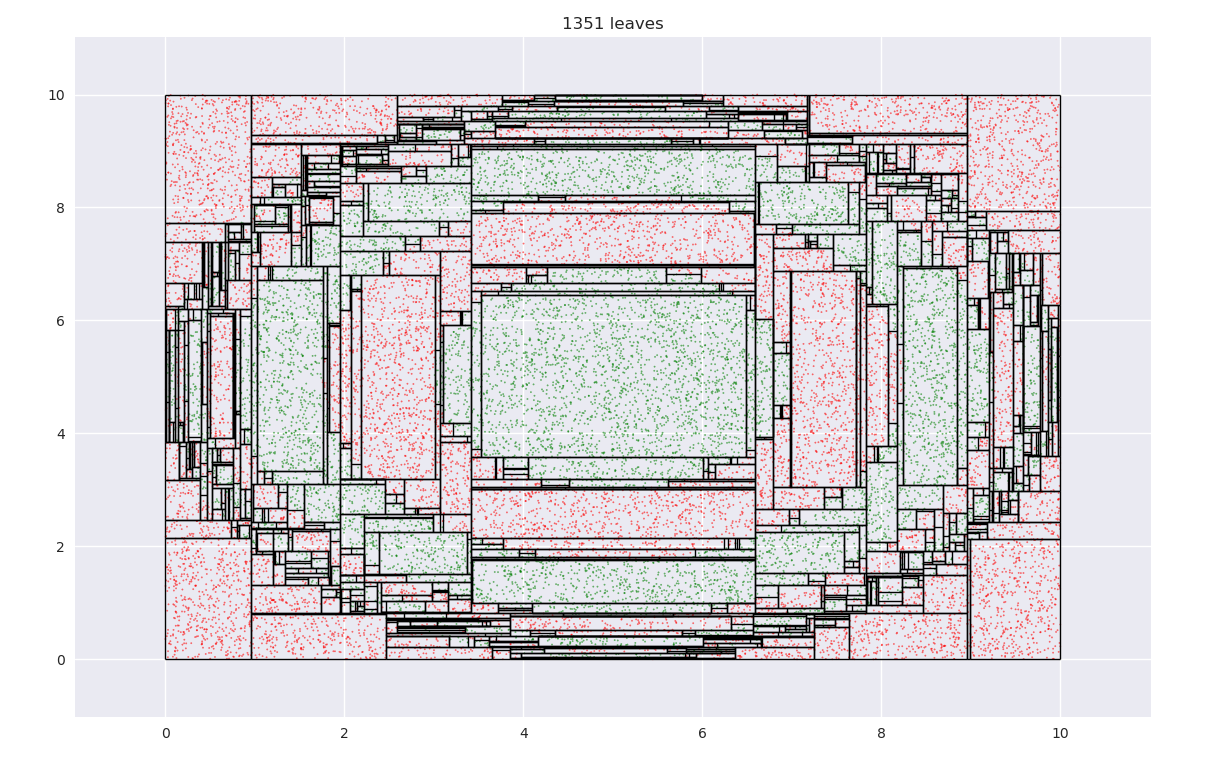}
    \caption{Tessellation of space produced by leaves of a decision tree.}
    \label{fig:dt_fragment}
\end{figure}

Our key insight is an interesting property of decision trees (DT). A DT fragments its input space into axis-parallel rectangles. Figure \ref{fig:dt_fragment} shows what this looks like when we learn a tree using CART on the dataset from Figure \ref{fig:circle_dataset}. Leaf regions are shown with the rectangles with the black edges.

\begin{figure}[h!]
    \centering
\includegraphics[scale=0.45]{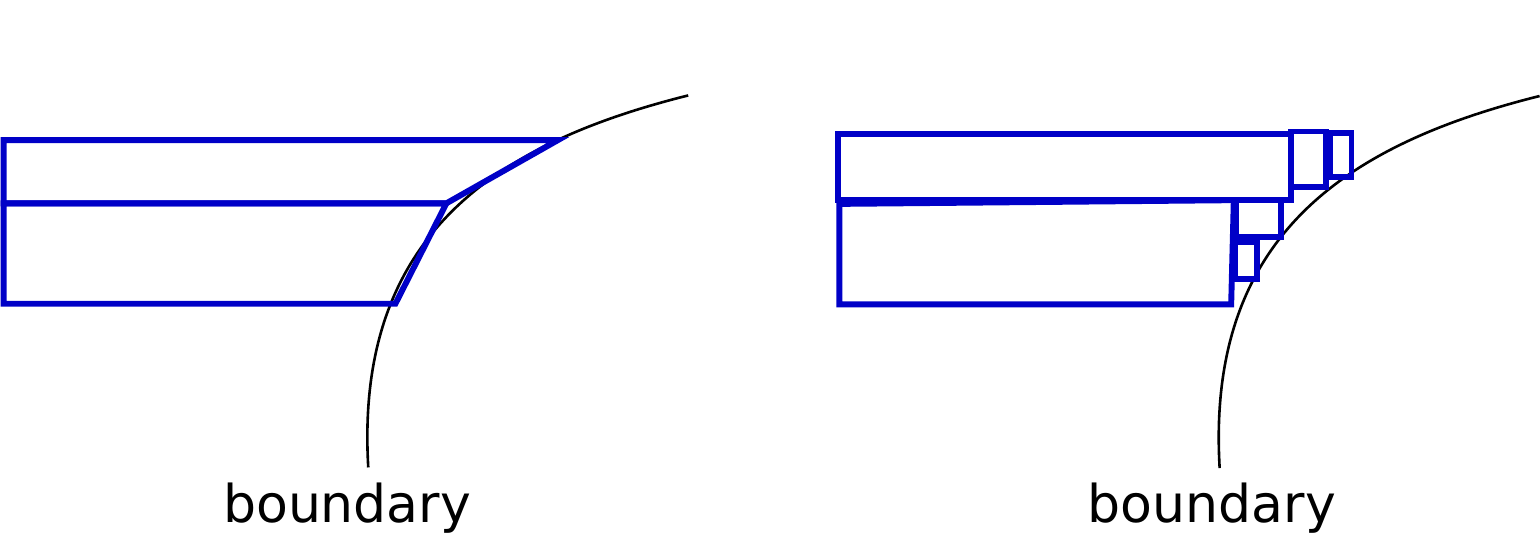}
    \caption{We see leaves of small areas because DTs are forced to approximate curvature with them.}
    \label{fig:why_small_leaves}
\end{figure}

Note how regions with relatively small areas almost always occur near boundaries. This happens here since none of the class boundaries are axis-parallel, and the DT, in being constrained in representation to axis-parallel rectangles, must use multiple small rectangles to approximate the curvature of the boundary. This is essentially \emph{piecewise linear approximation} in high dimensions, with the additional constraint that the ``linear pieces'' be axis-parallel. Figure \ref{fig:why_small_leaves} shows a magnified view of the interaction of leaf edges with a curved boundary. The first panel shows how hypothetical trapezoid leaves might closely approximate boundary curvature. However, since the DT may only use axis-parallel rectangles, we are led to multiple small rectangles as an approximation, as shown in the second panel.

We exploit this geometrical property; in general, leaf regions with relatively small areas (volumes, in higher dimensions) produced by a DT, represent regions close to the boundary. Instead of directly determining an optimal \emph{pdf} on the input space, we now do the following:
\begin{enumerate}
    \item Learn a DT, with no size restrictions, on the data $(X_{train}, Y_{train})$. Assume the tree produces $m$ leaves, where the region encompassed by a leaf is denoted by $R_i, 1 \leq i \leq m $. 
    
    \item Define a \emph{pmf} over the leaves, that assigns mass to a leaf in inverse proportion to its volume. Let $L \in \{1, 2, ..., m\}$ be a random variable denoting a leaf. Our \emph{pmf} is $P_L(i) = P(L=i) =f(R_i) \text{, where } f(R_i) \propto vol(R_i)^{-1}$.
    
    The probability of sampling outside any $R_i$ is set to $0$.
    
    \item To sample a point, sample a leaf first, based on the above \emph{pmf}, and then sample a point from within this leaf assuming a uniform distribution:
    \begin{enumerate}
        \item Sample a leaf, $i \sim P_L$.
        \item Sample a point within this leaf, $x \sim \mathcal{U}(R_i)$.
        \item Since leaves are characterized by low entropy of the label distribution, we assign the majority label of leaf $i$, denoted by $label(i)$, to the sampled point $x$.
    \end{enumerate} 
    
     Assuming we have $k$ unique labels, $label(i)$ is calculated as follows: 
     
     Let $S_i = \{y_j: y_j \in Y_{train}, x_j \in X_{train}, x_j \in R_i\}$. Then,
    \begin{align}
        &label(i) =  \argmax_k \hat{p}_{ik}\\
        \text{where,\;  } &\hat{p}_{ik} = \frac{1}{|S_i|} \sum_{S_i} I(y_j=k)
    \end{align}
    \label{eqn:majority_label}
    Note here that because of using $\mathcal{U}(R_i)$ we may generate points $x \notin X_{train}$. Also, since a point $x \in R_i \cap X_{train}$ gets \emph{assigned} $label(i)$, the conditional distribution of labels approximately equals the original distribution:
    
    \begin{equation}    \label{eqn:approx_label_dist_explained}
    p(Y_t|X_t) \approx p(Y_{train}|X_{train})
    \end{equation}
\end{enumerate}
\theoremstyle{definition}

We call such a DT a \emph{density tree}\footnote{We use this term since this helps us define a \emph{pdf} over the input space $\mathbb{R}^d$. We don't abbreviate this term to avoid confusion with ``DT''. DT always refers to a decision tree in this work, and the term ``density tree'' is used as-is.} which we formally define as follows.
\begin{definition}{} We refer to a DT as a \textbf{density tree} if (a) it is learned on $(X_{train}, Y_{train})$ with no size restrictions (b) there is a \emph{pmf} defined over its leaves s.t. $P_L(i) = P(L=i) =f(R_i) \text{, where } f(R_i) \propto vol(R_i)^{-1}$.
\end{definition}

Referring back to our desiderata, it should be clear how we address some of the challenges:
\begin{enumerate}
    \item The location of class boundaries are naturally produced by DTs, in the form of (typically) low-volume leaf regions.
    \item Instead of penalizing the lack of overlap with such boundary regions, we sample points in way that favors points close to class boundaries. 
    
    Note that in relation to Equation \ref{eqn:objective_breakup_a} (reproduced below), $q$ no longer ranges over all possible distributions; but over a restricted set relevant to the problem:
    \begin{align}
     p^*_\eta = \argmax_q accuracy(train_\mathcal{F}(q, \eta), p)
    \end{align}
    
\end{enumerate}

We visit the issue of efficiency towards the end of this section.

This simple scheme represents our approach at a high-level. However, this in itself is not sufficient to build a robust and efficient algorithm. We consider the following refinements to our approach:

\begin{enumerate}
    \item \textbf{\emph{pmf} at the leaf level}. What function $f$ must we use to construct our \emph{pmf}? One could just use $f(R_i) =  c \cdot vol(R_i)^{-1}$ where $c$ is the normalization constant $c = 1/\sum_{i=1}^m vol(R_i)^{-1}$. However, this quantity changes rapidly with volume. Consider a hypercube with edge-length $a$ in $d$ dimensions; the ratio of the (non-normalized) mass between this and another hypercube with edge-length $a/2$ is $2^d$. Not only is this change drastic, but it also has potential for numeric underflow.
    
    An alternative is to use a function that changes more slowly like the inverse of the length of the diagonal, $f(R_i) = c \cdot diag(R_i)^{-1} \text{ where } c=1/\sum_{i=1}^m diag(R_i)^{-1}$. Since DT leaves are axis-parallel hyperrectangles, $diag(R_i)$ is always well defined. In our hypercube example, the probability masses are $\propto 1/(a \sqrt{d})$ and $\propto 1/(a \sqrt{d}/2)$ when the edge-lengths are $a$ and $a/2$ respectively. The ratio of the non-normalized masses between the two cubes is now $2$. 
    
    This begs the question: is there yet another \emph{pmf} we can use, that is optimal in some sense? Instead of looking for such an optimal \emph{pmf}, we adopt the more pragmatic approach of starting with a ``base'' \emph{pmf} - we use the inverse of the diagonal length - and then allowing the algorithm to modify it, via \emph{smoothing}, to adapt it to the data. 
    
    \item \textbf{Smoothing}. Our algorithm may perform smoothing over the base \emph{pmf} as part of the optimization. We use \emph{Laplace smoothing} \cite[Section~3.4]{jurafsky2019speech}, with $\lambda$ as the smoothing coefficient. This modifies our \emph{pmf} thus: 
    \begin{equation}
    f'(R_i) = c  \Big(f(R_i) + \frac{ \lambda}{m}  \Big)%, 1 \leq i \leq m
    \end{equation}
     Here, $c$ is the normalization constant.
     The optimizer discovers the ideal value for $\lambda$.
    
    We pick Laplace smoothing because it is fast.
    Our framework, however, is general enough to admit a wide variety of options (discussed in Section \ref{ssec:extn}). 
    
    \item \textbf{Axis-aligned boundaries}. A shortcoming of our geometric view is if a boundary is axis-aligned, there are no leaf regions of small volumes along this boundary. This foils our sampling strategy. An easy way to address this problem is to transform the data by rotating or shearing it, and then construct a decision tree. See Figure \ref{fig:effect_of_rotation}. The image on the left shows a DT with two leaves constructed on the data that has an axis-parallel boundary. The image on the right shows multiple leaves around the boundary region, after the data is transformed (the transformation may be noticed at the top left and bottom right regions).
    
    \begin{figure}[h!]
    \centering
    \includegraphics[scale=0.45]{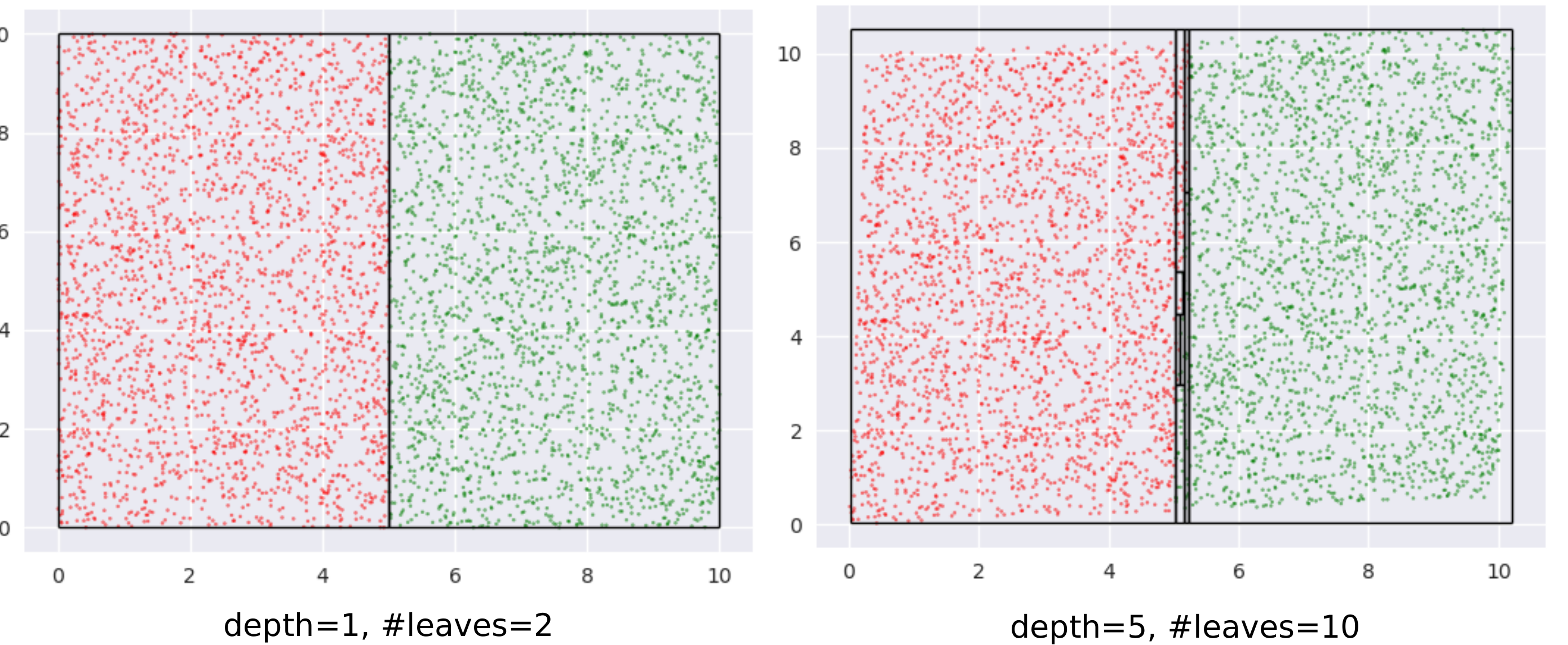}
    \caption{Left: Axis parallel boundaries don't create small regions. Right: This can be addressed by transforming the data. We see an increase in depth and the number of leaves of the density tree in the latter case.}
    \label{fig:effect_of_rotation}
    \end{figure}
    
    The idea of transforming data by rotation is not new \citep{Rodriguez:2006:RFN:1159167.1159358, JMLR:v17:blaser16a}. However, a couple of significant differences in our setup are: 
    \begin{enumerate}
        \item We don't require rotation per se as our specific transformation; any transformation that produces small leaf regions near the boundary works for us.
        \item Since \emph{interpretability in the original input space}  is our goal, we need to transform \emph{back} our sample. This would not be required, say, if our only goal is to increase classification accuracy.
    \end{enumerate}
    The need to undo the transformation introduces an additional challenge: we cannot drastically transform the data since sampled points in the transformed space might be outliers in the original space. Figure \ref{fig:too_much_rotation} illustrates this idea, using the same data as in Figure \ref{fig:effect_of_rotation}. 
    
    \begin{figure}[h]
    \centering
    \includegraphics[scale=0.45]{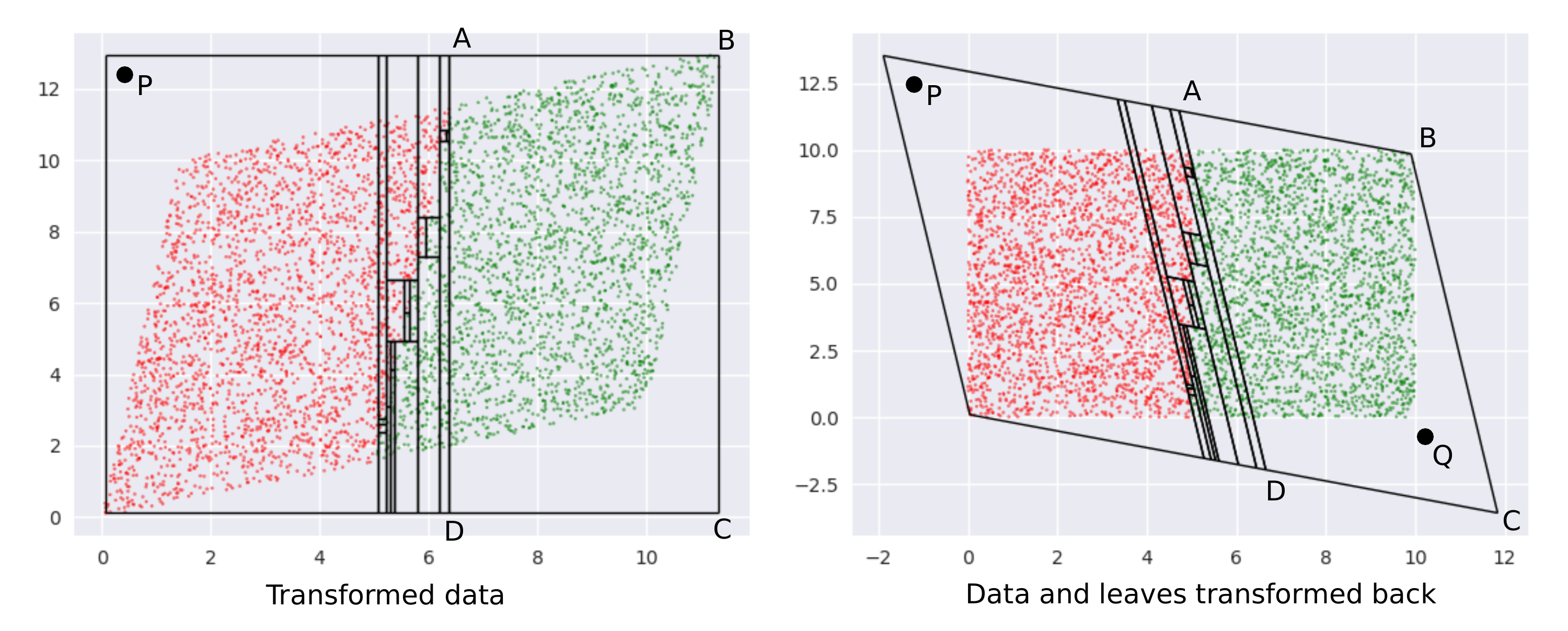}
    \caption{Left: transformed data. Right: the leaves in the inverse transformation contain regions outside the bounding box of the original dataset. See text for a description of points P and Q.}
    \label{fig:too_much_rotation}
    \end{figure}

    The first panel shows leaves learned on the data in the transformed space. Note how the overall region covered by the leaves is defined by the extremities - the top-right and bottom-left corners - of the region occupied by the transformed data. Any point within this rectangle is part of \emph{some} leaf in a DT learned in this space. Consider point $P$ - it is valid for our sampler to pick this. The second panel shows what the training data and leaf-regions look like when they are transformed back to the original space. Clearly, the leaves from the transformed space may not create a tight envelope around the data in the original space, and here, $P$ becomes an outlier. 
    
    Sampling a significant number of outliers is problematic because:
    \begin{enumerate}
        \item The validation and test sets do not have these points and hence learning a model on a training dataset with a lot of outliers would lead to sub-optimal accuracies.
        \item There is no way to \emph{selectively} ignore points like $P$ in their leaf, since we uniformly sample within the entire leaf region. The only way to avoid sampling $P$ is to ignore the leaf containing it (using an appropriate \emph{pmf}); which is not desirable since it also forces us to ignore the non-outlier points within the leaf.
    \end{enumerate}

    Note that we also cannot transform the leaves back to the original space \emph{first} and then sample from them, since (1) we lose the convenience and low runtime of uniform sampling $\mathcal{U}(R_i)$: the leaves are not simple hyperrectangles any more; (2) for leaves not contained within the data bounding box in the original space, we cannot sample from the entire leaf region without risking obtaining outliers again - see point $Q$ in $\overline{ABCD}$, in Figure \ref{fig:too_much_rotation}.
    
    A simple and efficient solution to this problem is to only \emph{slightly} transform the data, so that we obtain the small volume leaves at class boundaries (in the transformed space), but also, all valid samples are less likely to be outliers. This may be achieved by restricting the extent of transformation using a ``near identity'' matrix $A \in \mathbb{R}^{d \times d}$: 
    \begin{align}
        [A]_{pq}&=1, \text{ if } p=q\\
        [A]_{pq}&\sim \mathcal{U}([0, \epsilon]), \text{ if } p\neq q \text{, where }\epsilon \in \mathbb{R}_{> 0} \text{ is a small number. } 
    \end{align}
    With this transformation, we would \emph{still} be sampling outliers, but: 
    \begin{enumerate}
        \item Their numbers are not significant now.
        \item The outliers themselves are close to the data bounding box in the original space.
    \end{enumerate}
   
   These substantially weaken their negative impact on our technique.  
    
    The tree is constructed on $AX$, where $X$ is the original data, and samples from the leaves, $X'_t$, are transformed back with $A^{-1}X'_t$. Figure \ref{fig:effect_of_rotation} is actually an example of such a near-identity transformation.
    
    A relevant question here is how do we know \emph{when} to transform our data, i.e., when do we know we have axis-aligned boundaries? Since this is computationally expensive to determine, we always create multiple trees, each on a transformed version of the data (with different transformation matrices), and uniformly sample from the different trees. It is highly unlikely that \emph{all} trees in this \emph{bagging} step would have axis-aligned boundaries in their respective transformed spaces. Bagging also provides the additional benefit of low variance. 
    
    We denote this bag of trees and their corresponding transformations by $B$. Algorithm \ref{alg:create_bag} details how $B$ is created. Our process is not too sensitive to the choice of epsilon, hence we set $\epsilon=0.2$ for our experiments.
     
\SetKw{KwBy}{by}
\begin{algorithm}
 \KwData{($X_{train}, Y_{train}$), size of bag $n$}
 \KwResult{$B=\{(T_1, A_1), (T_2, A_2), ..., (T_n, A_n)\}$}
%Create test and validation sets: $(X_{val}, y_{val}), (X_{test}, y_{test})$\;
$B=\{\}$\;
\For {$i \gets 1$ \KwTo $n$}{
$\text{Create matrix }A_i \in \mathbb{R}^{d \times d}\text{ s.t. } [A_{i}]_{pq}=1, \text{if } p=q \text{ else } [A_{i}]_{pq}\sim \mathcal{U}([0, \epsilon])$\;
$X'_{train} \gets A_i X_{train}$\;
$T_i \gets \text{learn tree on } (X'_{train}, Y_{train})$\;
$B \gets B \cup \{(T_i, A_i)\}$

 }
\Return $B$
 \caption{Create bag of density trees, $B$}
 \label{alg:create_bag}
\end{algorithm}

    \item \textbf{Selective Generalization}. Since we rely on geometric properties alone to define our \emph{pmf}, all boundary regions receive a high probability mass irrespective of their contribution to classification accuracy. 
    This is not desirable when the classifier is small and must focus on a few high impact regions. In other words, we prioritize all boundaries, but not all of them are valuable for classification; our algorithm needs a mechanism to ignore some of them. We refer to this desired ability of the algorithm as \emph{selective generalization}.
    
    \begin{figure}[h!]
    \centering
    \includegraphics[scale=0.5]{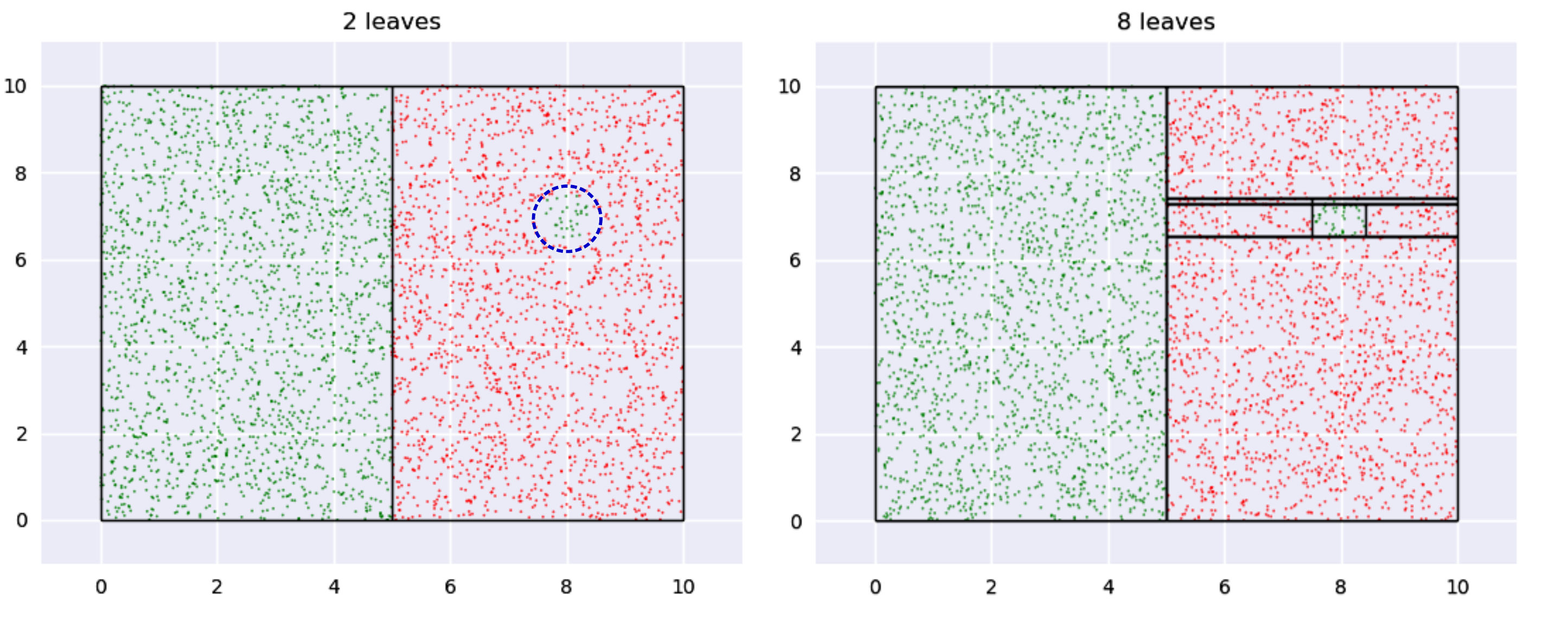}
    \caption{A region of low impact is shown in the first panel with a dashed blue circle. The first tree ignores this while a second, larger, tree creates a leaf for it.}
    \label{fig:unreg_vs_reg}
    \end{figure}

    Figure \ref{fig:unreg_vs_reg} illustrates the problem and suggests a solution. The data shown has a small green region, shown with a dashed blue circle in the first panel, which we may want to ignore if we had to pick between learning its boundary or the relatively significant vertical boundary. The figure shows two trees of different depths learned on the data - leaf boundaries are indicated with solid black lines.  A small tree, shown on the left, automatically ignores the circle boundary, while a larger tree, on the right, identifies leaves around it.
    
    Thus, one way to enable selective generalization is to allow our technique to pick a density tree of appropriate depth.
    
     But a shallow density tree is already part of a deeper density tree! - we can just sample at the depth we need. Instead of constructing density trees with different depths, we learn a ``depth distribution'' over fully grown density trees; drawing a sample from this tells us what fraction of the tree to consider.

    \begin{figure}[h!]
    \centering
    \includegraphics[scale=0.29]{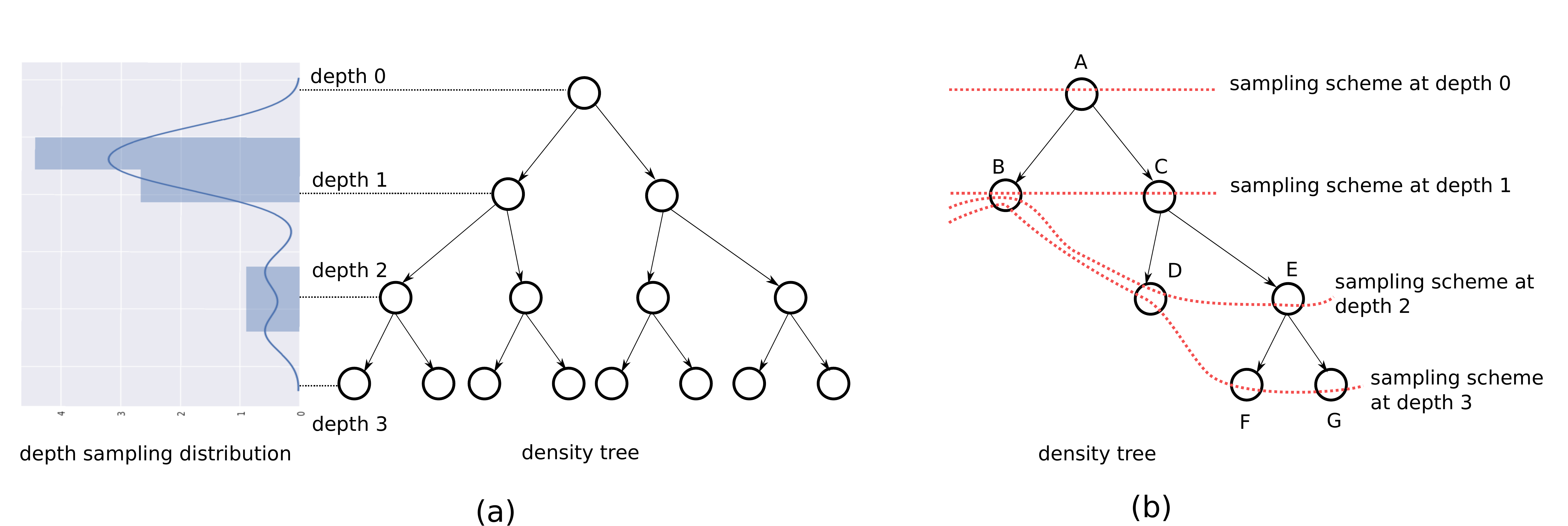}
    \caption{(a) The set of nodes at a depth have an associated \emph{pmf} to sample from (not shown). A depth is picked based on the IBMM. (b) In case of an incomplete binary tree, we use the last available nodes closest to the depth being sampled from, so that the entire input space is represented. The red dotted lines show the nodes comprising the sampling scheme for different depths.}
    \label{fig:depth_schematic}
    \end{figure}

    Figure \ref{fig:depth_schematic}(a) illustrates this idea. The depth distribution is visualized vertically and adjacent to a tree. We sample $r \in [0, 1]$ from the distribution, and scale and discretize it to reflect a valid value for the depth. Let $depth_T()$ be the scaling/discretizing function for a tree $T$. Taking the tree in the figure as our example, $r=0$ implies we sample our data instances from the nodes at $depth_T(r)=0$ i.e. at the \emph{root}, and $r=0.5$ implies we must sample from the nodes at $depth_T(r)=1$. We refer to the \emph{pmf} for the nodes at a depth to be the \emph{sampling scheme} at that depth. $T$ has $4$ sampling schemes - each capturing class boundary information at a different granularity, ranging from the root with no information and the leaves with the most information.  
    
    We use an IBMM for the depth distribution. Similar to the one previously discussed in Section \ref{sec:density}, the depth-distribution has a parameter $\alpha$ for the DP and parameters $\{a, b, a', b'\}$ for its \emph{Beta} priors. The significant difference is we have just one dimension now: the depth. The IBMM is shared across all trees in the bag; Algorithm \ref{alg:bag_trees} provides details at the end of this section.
    
    %A draw from a depth-distribution tells us the depth in the density tree at which we should consider nodes to sample points from. We treat these nodes as our leaves, constructing our smoothed \emph{pmf} over them as before. This distribution is represented as a IBMM on one dimension - the depth. 
    %Figure \ref{fig:depth_schematic}(a) illustrates this idea. Similar to the representation described in Section \ref{sec:density}, the depth-distribution has a parameter $\alpha$ for the DP and parameters for its \emph{Beta} priors: $a, b, a', b'$.
    
    \item \textbf{Revisiting label entropy}. When we sampled only from the leaves of a density tree, we could assign the majority label to the samples owing to the low label entropy. However, this is not true for nodes at intermediate levels - which the depth distribution might lead us to sample from. We deal with this change by defining an \textbf{entropy threshold} $\boldsymbol{E}$. If the label distribution at a node has $entropy \leq E$, we sample uniformly from the region encompassed by the node (which may be a leaf or an internal node) and use the majority label. However, if the $entropy > E$, we sample only among the \emph{training} data instances that the node covers. Like $\epsilon$, our technique is not very sensitive to a specific value of $E$ (and therefore, need not be learned), as long as it is reasonably low: we use $E=0.15$ in our experiments.

    \item \textbf{Incomplete trees}. Since we use CART to learn our density trees, we have binary trees that are always \emph{full}, but not necessarily \emph{complete}, i.e., the nodes at a certain depth alone might not represent the entire input space. To sample at such depths, we ``back up'' to the nodes at the closest depth. Figure \ref{fig:depth_schematic}(b) shows this: at $depth=0$ and $depth=1$, we can construct our \emph{pmf} with only nodes available at these depths, $\{A\}$ and $\{B, C\}$ respectively, and still cover the whole input space. But for $depth=2$ and $depth=3$, we consider nodes $\{B, D, E\}$ and $\{B, D, F, G\}$ respectively. The dotted red line connects the nodes that contribute to the sampling scheme for a certain depth. 
    
    %Note that when we were sampling only from the leaves of a density tree, we decided that we could just assign the majority label to the samples given low label entropy. This is not true at nodes at intermediate levels which may have high label entropies. We deal with this problem by defining an \textbf{entropy threshold} $\boldsymbol{E}$. If the label distribution at a node has $entropy \leq E$, we sample uniformly within the region encompassed by the node (which may be a leaf or an internal node) and assign the majority label to the sampled instances. This potentially \emph{generates} new points. However, if the $entropy > E$, we sample only among the \emph{training} data instances that the node covers. $E$ can be set to a reasonably low value like $0.15$ and need not be learned.
    Algorithm \ref{alg:bag_trees} shows how sampling from $B$ works.

\SetKw{KwBy}{by}
\begin{algorithm}
 \KwData{\# points to sample $N$, bag of density trees $B$, depth distribution $\Psi$, smoothing parameter $\lambda$}
 \KwResult{$(X, Y), X \in \mathbb{R}^{N \times d}, Y \in \mathbb{R}^n$}
%Create test and validation sets: $(X_{val}, y_{val}), (X_{test}, y_{test})$\;
$X=[\;], Y=[\;]$\;
\For {$i\gets1$ \KwTo $N$}{
$r \sim \Psi$ \;
$T, A \gets \text{randomly pick a tree and the corresponding transformation from } B$\;

$\Theta \gets$ construct \emph{pmf}  over the nodes at $depth_T(r)$, smooth with $\lambda$\tcp{back-up if depth is incomplete}
$L \sim \Theta$\ \tcp{$L$ is a node at $depth_T(r)$}
$S_L \gets \{(x_j, y_j): x_j \in X_{train} \cap R_L \text{ and } y_j \in Y_{train}  \text{ is its label} \}$\;

 \eIf{$entropy(L) \leq E$}{
   $x \sim \mathcal{U}(L), y \gets label(L)$ \tcp{$label()$ defined in Equation \ref{eqn:majority_label}}
   }{
   $(x,y) \sim S_L$ \tcp{notation: sample a point from $S_L$}
  }

$X \gets \begin{bmatrix} X\\ A^{-1}x \end{bmatrix},Y \gets \begin{bmatrix} Y\\ y \end{bmatrix} $\;
 }
\Return $(X, Y)$
 \caption{Sampling from a bag of density trees, $B$}
 \label{alg:bag_trees}
\end{algorithm}

\end{enumerate}

\qed
    \begin{figure}[h!]
    \centering
    \includegraphics[scale=0.35]{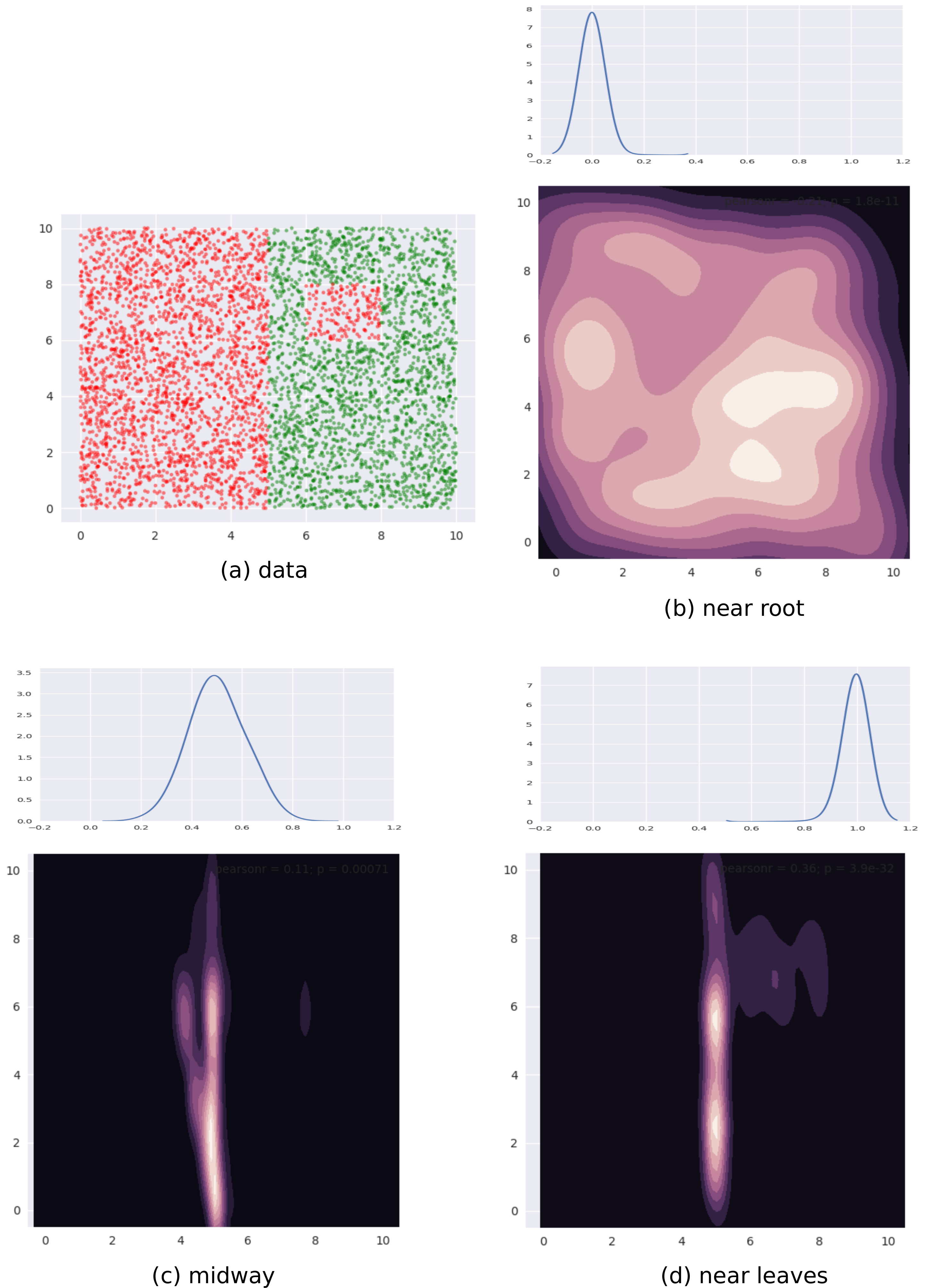}
    \caption{(a) shows our dataset, while (b), (c), (d) show how the sample distribution varies with change of the depth distribution.}
    \label{fig:working_pmf}
    \end{figure}

Figure \ref{fig:working_pmf} illustrates some of the distributions we obtain using our mechanism. Panel (a) shows our data - note, we only have axis-aligned boundaries. In panels (b), (c), (d), we show the depth distribution at the top, going from favoring the root in (b), to nodes halfway along the height of the tree in (c), finally to the leaves in (d). The contour plot visualizes the distributions, where a lighter color indicates relatively higher sample density. We see that in (b), we sample everywhere in the data bounding box. In (c), the larger boundary is identified. In (d), the smaller boundary is also identified. A bag of size $5$ was used and the smoothing coefficient $\lambda$ was held constant at a small value.

This completes the discussion of the salient details of our sampling technique. The optimization variables are summarized below:
\begin{enumerate}
\item $\lambda$, the Laplace smoothing coefficient. 
\item $\alpha$, the DP parameter. 
\item $\{a, b, a', b'\}$, the parameters of the \emph{Beta} priors for the IBMM depth distribution. A component/partition $i$ is characterized by the distribution $Beta(A_i, B_i)$, where $A_i \sim Beta(a, b)$, $B_i \sim Beta(a', b')$. 
\end{enumerate}

The IBMM and its parameters, $\{\alpha, a, b, a', b'\}$, are shared across all trees in the bag $B$, and $\lambda$ is shared across all sampling schemes.

We also introduced two additional parameters: $\epsilon$ and $E$. As mentioned previously, we do not include them in our optimization since our process is largely insensitive to their precise values as long as these are reasonably small. We use $\epsilon=0.2$ and $E=0.15$ for our experiments.

The above parameters exclusively determine how the sampler works. In addition, we propose the following parameters:
\begin{enumerate}
\setcounter{enumi}{3}
\item $N_s \in \mathbb{N}$, sample size. The sample size can have a significant effect on model performance. We let the optimizer determine the best sample size to learn from. We constrain $N_s$ to be larger than the minimum number of points needed for statistically significant results.

Note that we can allow $N_s > |X_{train}|$. This larger sample will be created by either repeatedly sampling points - at nodes where the label $entropy >E$ - or by generating synthetic points, when $entropy \leq E$. 

\item $p_o \in [0, 1]$ - proportion of the sample from the original distribution. Given a  value for $N_s$, we sample $(1-p_o)N_s$ points from the density tree(s) and $p_oN_s$ points (stratified) from our training data $(X_{train}, Y_{train})$. 

Recall that our hypothesis is that learning a distribution helps until a size $\eta'$ (Equation \ref{eqn:size}). Beyond this size, we need to provide a way for the sampler to reproduce the original distribution. While it is possible the optimizer finds a $\Psi_t$ that corresponds to this distribution, we want to make this easier: now the optimizer can simply set $p_o = 1$. Essentially, $p_o$ is way to ``short-circuit'' the discovery of the original distribution.

This variable provides the additional benefit that observing a transition $p_o=0 \to 1$, as the model size increases, would empirically validate our hypothesis.

%See section \ref{sec:impl} for additional details.

\end{enumerate}

We have a total of \textbf{eight optimization variables} in this technique. The variables that influence the sampling behaviour are collectively denoted by $\Psi=\{\alpha, a, b, a', b'\}$. The complete set of variables is denoted by $\Phi = \{\Psi, N_s, \lambda, p_o\}$. 

%Although our bag of density trees, $B$, has more than one tree, the parameter $\Psi$ \emph{is shared}; we sample a value $i \sim \Psi, i \in [0, 1]$, and scale and discretize it to reflect a valid value for depth for the density tree being considered. For instance, if we want to sample from tree $T$ where $depth(T)=10$, $i=0$ implies we must sample at $depth_T(i)=0$ i.e. at the \emph{root}, and $i=0.5$ implies we must sample at $depth_T(i)=5$. 

This is a welcome departure from our naive solution: the number of optimization variables does not depend on the dimensionality $d$ at all! Creating density trees as a preprocessing step gives us a \emph{fixed set of 8 optimization variables} for any data. This makes the algorithm much more efficient than before, and makes it practical to use for real world data.

%\footnote{Note that our running time still depends on the dimensionality since $\mathcal{U}(L)$ runs in $O(d)$.}. 
%Of course, there is a hidden cost we incur in learning the bag of density trees, but the total run time is still much lower than the previous solution.

Algorithm \ref{algo:density_tree_opt} shows how we modify our naive solution to incorporate the new sampler.

\begin{algorithm}[tbh]
 \KwData{Learning algorithm $train_{\mathcal{F}}()$, size of model $\eta$, data $(X,Y)$, number of density trees $n$, iterations $T$}
 \KwResult{$\Phi^*, s_{test}$ }
Create stratified samples $(X_{train}, Y_{train}), (X_{val}, Y_{val}), (X_{test}, Y_{test})$ from $(X, Y)$\;
Construct bag $B$ of $n$ density trees on $(X_{train}, Y_{train})$\;
\For {$t\gets1$ \KwTo $T$}{
   $\Phi_{t} \gets suggest(s_{t-1}, ... s_1, \Phi_{t-1}, ..., \Phi_1)$ \tcp{randomly initialize at $t=1$}
   \tcp{Note: $\Phi_t = \{\Psi_t, N_{s\_t}, \lambda_t, p_{o\_t}\}$ where $\Psi_t = \{\alpha_t, a_t, b_t, a'_t, b'_t\}$.}
$N_o \gets p_{o\_t} \times N_{s\_t}$ \;
$N_{B} \gets N_{s\_t} - N_o$ \;
$(X_o, Y_o) \gets \text{sample } N_o \text{ points from } (X_{train}, Y_{train})$ based on $p(X_{train}; \Psi_t)$\; 

$(X_{dp}, Y_{dp}) \gets \text{sample } N_{B} \text{ points from }B$, using Algorithm \ref{alg:bag_trees}\;

$X_t \gets \begin{bmatrix} X_o\\ X_{dp} \end{bmatrix}, Y_t \gets \begin{bmatrix} Y_o\\ Y_{dp} \end{bmatrix}$\tcp{combine the above samples}

  $M_t \gets train_{\mathcal{F}}((X_t, Y_t), \eta) $\;
  $s_t \gets accuracy(M_t, (X_{val}, Y_{val}))$\;
 }
$t^* \gets \argmax_t{\{s_1, s_2, ..., s_{T-1}, s_T\}}$\;
$\Phi^* \gets \Phi_{t^*}$\;

$(X^*, Y^*) \gets$ sample $N^*_s$ points from $(X_{train}, Y_{train})$ and $B$ based on $p^*_o$\;
$M^* \gets train_{\mathcal{F}}((X^*, Y^*), \eta) $ \;
$s_{test} \gets accuracy(M^*,(X_{test}, Y_{test}))$\;
\Return $\Phi^*$, $s_{test}$
\caption{Adaptive sampling using density trees}
 \label{algo:density_tree_opt}
\end{algorithm}

As before, we discover the optimal $\Phi$ using TPE as the optimizer and $accuracy()$ as the fitness function. We begin by constructing our bag of density trees, $B$, on transformed versions of $(X_{train}, Y_{train})$, as described in Algorithm \ref{alg:create_bag}.
At each iteration in the optimization, based on the current value $p_{o\_t}$, we sample data from $B$ and $(X_{train}, Y_{train})$, train our model on it, and evaluate it on $(X_{val}, Y_{val})$. In our implementation, lines 7-11 are repeated (thrice, in our experiments) and the accuracies are averaged to obtain a stable estimate for $s_t$.

\section{Experiments}
\label{sec:expts}

This section discusses experiments that validate our technique and demonstrate its practical utility. We describe our experimental setup in Section \ref{sec:setup} and present our observations and analysis in Section \ref{sec:obv}. %In Section \ref{sec:setup}, we describe our setup: datasets, model families, parameter settings, etc. Section \ref{sec:obv} presents the results and our analysis.

\subsection{Setup}
\label{sec:setup}
We evaluate Algorithm \ref{algo:density_tree_opt} using $3$ different learning algorithms, i.e., $train_\mathcal{F}()$,  on $13$ real world datasets. We construct models for a wide range of sizes, $\eta$, to comprehensively understand the behavior of the algorithm. For each combination of dataset, learning algorithm and model size, we record the percentage \emph{relative improvement} in the $F1$(macro) score on $(X_{test}, Y_{test})$ compared to the \emph{baseline} of training the model on the original distribution:
\begin{equation*}
    \delta F1 = \frac{100 \times (F1_{new} - F1_{baseline})}{F1_{baseline}}
\end{equation*}
Since the original distribution is part of the optimization search space, i.e., when $p_o=1$, the lowest improvement we report is $0\%$, i.e., $\delta F1 \in [0, \infty)$. 
All reported values of $\delta F1$ are averaged over \textbf{three} runs of Algorithm \ref{algo:density_tree_opt}. As mentioned before, in \emph{each} such run, lines 7-11 in the algorithm are repeated thrice to obtain a robust estimate for $accuracy()$, and thus, $s_t$. 
%Further details follow.

\subsubsection{Data}
We use a variety of real-world datasets, with different dimensionalities and number of classes to test the generality of our approach. The datasets were obtained from the LIBSVM website \citep{CC01a}, and are listed in Table \ref{tab:ddatasets}.

\begin{table}
\centering
\caption{Datasets}\label{tab:ddatasets}
\hspace{0.75pt}
\begin{tabular}{lrr}
\toprule
dataset & dimensions & \# classes \\ 
\midrule
cod-rna & 8 & 2\\
ijcnn1 &  22 &  2\\
higgs &  28 &  2\\
covtype.binary &  54 &  2\\
phishing &  68 &  2\\
a1a &  123 &  2\\
pendigits &  16 &  10\\
letter &  16 &  26\\
Sensorless &  48 &  11\\
senseit\_aco &  50 &  3\\
senseit\_sei &  50 &  3\\
covtype &  54 &  7\\
connect-4 &  126 &  3\\
\bottomrule
\end{tabular}
\end{table}

\subsubsection{Models}
We use the following model families, $\mathcal{F}$, and learning algorithms, $train_\mathcal{F}()$, in our experiments:
\begin{enumerate}
    \item \emph{Decision Trees}: We use the implementation of CART in the \emph{scikit-learn} library \citep{scikit-learn}. Our notion of size here is the depth of the tree.
    
    \smallskip
    
    \emph{Sizes}: For a dataset, we first learn an optimal tree $T_{opt}$ based on the \emph{F1-score}, without any size constraints. Denote the depth of this tree by $depth(T_{opt})$. We then try our algorithm for these settings of CART's $max\_depth$ parameter: $\{1, 2, ..., min(depth(T_{opt}), 15)\}$, i.e., we experiment only up to a model size of $15$, stopping early if we encounter the optimal tree size. Stopping early makes sense since the model has attained the size needed to capture all patterns in the data; changing the input distribution is not going to help beyond this point.
        
    Note that while our notion of size is the \emph{actual} depth of the tree produced, the parameter we vary is $max\_depth$; this is because decision tree libraries do not allow specification of an exact tree depth.  This is important to remember since CART produces trees with actual depth up to as large as the specified $max\_depth$, and therefore, we might not see actual tree depths take all values in $\{1, 2, ..., min(depth(T_{opt}), 15)\}$, e.g., $max\_depth=5$ might give us a tree with $depth=5$, $max\_depth=6$ might also result in a tree with $depth=5$, but $max\_depth=7$ might give us a tree with $depth=7$. We report relative improvements at actual depths.
    
    \item \emph{Linear Probability Model (LPM)} \citep{Mood291880}: This is a linear classifier. Our notion of size is the number of terms in the model, i.e., features from the original data with non-zero coefficients. We use our own implementation based on \emph{scikit-learn}. Since LPMs inherently handle only binary class data, for a multiclass problem, we construct a \emph{one-vs-rest} model, comprising of as many binary classifiers as there are distinct labels. The given size is enforced for \emph{each} binary classifier. For instance, if we have a 3-class problem, and we specify a size of $10$, then we construct $3$ binary classifiers, each with $10$ terms.
    We did not use the more common \emph{Logistic Regression} classifier because: (1) from the perspective of interpretability, LPMs provide a better sense of variable importance \citep{Mood291880} (2) we believe our effect is equally well illustrated by either linear classifier.
    
    We use the \emph{Least Angle Regression} \citep{efron2004} algorithm, that grows the model one term at a time, to enforce the size constraint.

    \smallskip
    
    \emph{Sizes}: For a dataset with dimensionality $d$, we construct models of sizes: $\{1, 2, ..., min(d, 15)\}$. Here, the early stopping for LPM happens only for the dataset \emph{cod-rna}, which has $d =8$. All other datasets have $d>15$ (see Table \ref{tab:ddatasets}).

    \item \emph{Gradient Boosted Model (GBM)}: We use decision trees as our base classifier in the boosting. Our notion of size is the number of trees in the boosted forest for a \emph{fixed maximum depth} of the base classifiers. We use the \emph{LightGBM} library \citep{Ke:2017:LHE:3294996.3295074} for our experiments. 
    
    We run two sets of experiments with the GBM, with maximum depths fixed at $2$ and $5$. This helps us compare the impact of our technique when the model family $\mathcal{F}$ inherently differs in its effective capacity, e.g., we would expect a GBM with $10$ trees and a maximum depth of $5$ to be more accurate than a GBM with $10$ trees and a maximum depth of $2$.

    \smallskip
    
    \emph{Sizes}: If the optimal number of boosting rounds for a dataset is $r_{opt}$, we explore the model size range: $\{1, 2, ..., min(r_{opt}, 10)\}$.  We run two sets of experiments with GBM - one using base classification trees with $max\_depth=2$, and another with $max\_depth=5$. Both experiments use the same range for size/boosting rounds.
    
\end{enumerate}
The density trees themselves use the CART implementation in \emph{scikit-learn}. We use the $Beta$ distribution implementation provided by the \emph{SciPy} package \citep{scipy}.

%It is probably opportune to note that not all ML algorithms may allow direct enforcement of a model size. Usually a regulariz
\subsubsection{Parameter Settings}

    Since TPE performs optimization with \emph{box constraints}, we need to specify our search space for the various parameters in Algorithm \ref{algo:density_tree_opt}:
    \begin{enumerate}
        \item $\lambda$: this is varied in the \emph{log-space} such that $\log_{10} \lambda \in [-3, 3]$.
        \item $p_o$: We want to allow the algorithm to arbitrarily mix samples from $B$ and $(X_{train}, Y_{train})$. Hence, we set $p_o \in [0, 1]$.
        \item $N_s$: We set $N_s \in [1000, 10000]$. The lower bound ensures that we have statistically significant results. The upper bound is set to a reasonably large value.
        \item $\alpha$: For a DP, $\alpha \in \mathbb{R}_{>0}$.
        We use a lower bound of $0.1$. 
        
        We rely on the general properties of a DP to estimate an upper bound, $\alpha_{max}$.
        Given $\alpha$, for $N$ points, the expected number of components $k$ is given by:  
        \begin{align}
            &E[k|\alpha] = O(\alpha H_N) \\
            &E[k|\alpha]  \leq \alpha H_N \\
            &\alpha \geq \frac{E[k|\alpha]}{H_N }
        \end{align}
        Here, $H_N$ is the $N^{th}$ \emph{harmonic sum} (see \citet{bayes_notes}).
        
        Since our distribution is over the depth of a density tree, we already know the maximum number of components possible, $k_{max}=1+\text{depth of density tree}$. We use $N=1000$, since this is the lower bound of $N_s$, and we are interested in the upper bound of $\alpha$ (note $H_N \propto N$ - see Section \ref{ssec:harmonic}). We set $k_{max}=100$ (this is greater than any of the density tree depths in our experiments) to obtain a liberal upper bound, $\alpha_{max} = 100/H_{1000}=13.4$. Rounding up, we set $\alpha \in [0.1, 14]$ \footnote{We later observe from our experiments that this upper bound is sufficient since nearly all depth distributions have up to only $2$ dominant components (see Figures \ref{fig:dt_kde_density_depths},  \ref{fig:lpm_combined}, \ref{fig:gbdt2_combined}, \ref{fig:gbdt5_combined}).}.

        %We estimate this upper bound with $c \cdot k_{max}/H_{N'}$, where $c \geq 1 \text{ and } N'=\text{ 1 + depth of tree}$. We use this value of $N'$ since we assume that the we would have enough points (lower bound of the $N_s$ parameter) to at least sample once from each depth; also, typically $N' \ll N$, giving us a liberal upper bound. 
        
        %Thus, we set $\alpha \in [0.1, c \cdot k_{max}/H_{k_{max}}]$, where $k_{max}=1+\text{depth of tree}$. We use $c=2$.
        
        We draw a sample from the IBMM using \emph{Blackwell-MacQueen} sampling \citep{blackwell1973}. %Other equivalent methods are the \emph{Chinese Restaurant Process} \parencite{ref10.1007/BFb0099421} or the \emph{Stick-Breaking construction} \parencite{sethuraman94}.
        
        \item $\{a, b, a', b'\}$: Each of these parameters are allowed a range $[0.1, 10]$ to admit various shapes for the $Beta$ distributions.
    \end{enumerate}
    
    We need to provide a budget $T$ of iterations for the TPE to run. In the case of DT, GBM and binary class problems for LPM, $T=3000$. Since multiclass problems in LPM require learning multiple classifiers, leading to high running times, we use a lower value of $T=1000$. We arrived at these budget values by trial and error; not low enough to lead to inconclusive results, not unreasonably high to run our experiments.

\subsection{Observations and Analysis}
\label{sec:obv}
We break down our results and discussion by the $train_{\mathcal{F}}()$ used.% The final section, Section \ref{sssec:common_patterns}, discusses common patterns in the depth distribution.

\subsubsection{DT Results}
\label{sssec:dt_results}
The DT results are shown in Table \ref{tab:dec_tree_improvements}.
A series of unavailable scores, denoted by ``-'', toward the right end of the table for a dataset denotes we have already reached its optimal size.
For ex in Table \ref{tab:dec_tree_improvements}, \emph{cod-rna} has an optimal size of $10$. %A missing value before the optimal size denotes that the corresponding actual size was not obtained for any setting of the $max\_depth$ parameter. An example is $depth=2$ for \emph{cod-rna}.

For each dataset, the best improvement across different sizes is shown in bold. The horizontal line separates binary datasets from multiclass datasets.

\begin{table*}\scriptsize
\centering
\caption{Classification Results with DTs. Values indicate improvements $\delta F1$.}\label{tab:dec_tree_improvements}
\hspace{0.5pt}
\resizebox{\textwidth}{!}{\begin{tabular}{lrrrrrrrrrrrrrrrrr}
\toprule
depth =  &     1  &     2  &    3  &    4  &    5  &    6  &   7  &   8  &   9  &   10 &   11 &   12 &   13 &   14 &   15 \\
\cmidrule(lr){1-1}
datasets       &        &        &        &       &       &       &       &       &       &      &       &       &       &       &       \\
\midrule
cod-rna        &   0.22 &   0.82 &  0.36 &  \textbf{2.20} &  0.23 &  0.17 & 0.35 & 0.30 & 0.28 & 0.00 &    - &    - &    - &    - &    - \\
ijcnn1         &   3.66 &  \textbf{14.80} & 10.34 & 11.67 &  4.61 &  1.01 & 2.06 & 1.85 & 1.17 & 0.00 & 0.35 & 0.14 & 0.38 & 0.00 & 1.37 \\
higgs          &   \textbf{4.32} &   0.75 &  0.06 &  0.22 &  0.00 &  0.47 &    - &    - &    - &    - &    - &    - &    - &    - &    - \\
covtype.binary &   0.16 &   0.01 &  0.60 &  0.89 &  0.48 &  0.58 & 0.05 & 0.27 & 1.27 & 0.72 & 0.00 & \textbf{1.64} &    - &    - &    - \\
phishing       &   0.00 &   \textbf{0.76} &  0.03 &  0.09 &  0.51 &  0.13 & 0.19 & 0.52 & 0.12 & 0.13 & 0.00 & 0.00 & 0.00 & 0.00 & 0.00 \\
a1a            &   0.00 &   5.04 &  \textbf{5.77} &  5.45 &  3.60 &  1.73 & 2.97 & 0.82 & 1.63 &    - & 0.65 & 0.00 & 2.35 &    - & 0.92 \\
\midrule
pendigits      &  \textbf{10.98} &   2.80 &  4.67 &  9.45 &  4.41 &  2.53 & 0.92 & 0.06 & 0.13 & 0.00 & 0.00 & 0.00 & 0.00 & 0.00 &    - \\
letter         &   0.16 &  10.68 & 32.39 & \textbf{44.58} & 39.66 & 19.20 & 9.37 & 4.75 & 3.77 & 1.72 & 0.35 & 0.00 & 0.00 & 0.00 & 0.00 \\
Sensorless     &   0.00 &  40.93 & 70.98 & \textbf{82.53} & 39.72 & 17.07 & 7.97 & 4.77 & 2.26 & 1.44 & 0.88 & 0.33 & 0.75 & 0.84 &    - \\
senseit\_aco    &  \textbf{19.77} &   0.73 &  2.73 &  1.86 &  1.38 &  0.75 & 0.77 &    - &    - &    - &    - &    - &    - &    - &    - \\
senseit\_sei    &   1.78 &   0.63 &  \textbf{2.08} &  0.03 &  1.52 &  1.32 & 0.00 &    - &    - &    - &    - &    - &    - &    - &    - \\
covtype        &  34.44 & \textbf{115.25} & 13.54 &  9.68 &  5.38 &  3.31 & 2.02 & 2.00 & 2.73 & 3.36 & 4.30 & 2.79 & 0.00 & 1.14 & 1.03 \\
connect-4      & \textbf{180.92} &  29.47 & 15.44 & 11.99 &  3.99 & 11.12 & 2.45 & 6.14 & 3.27 & 1.54 & 4.48 & 3.09 & 2.96 & 3.00 & 1.57 \\
\bottomrule
\end{tabular}
}
\end{table*}

\begin{figure*}[h!]
    \centering
    \includegraphics[scale=0.5]{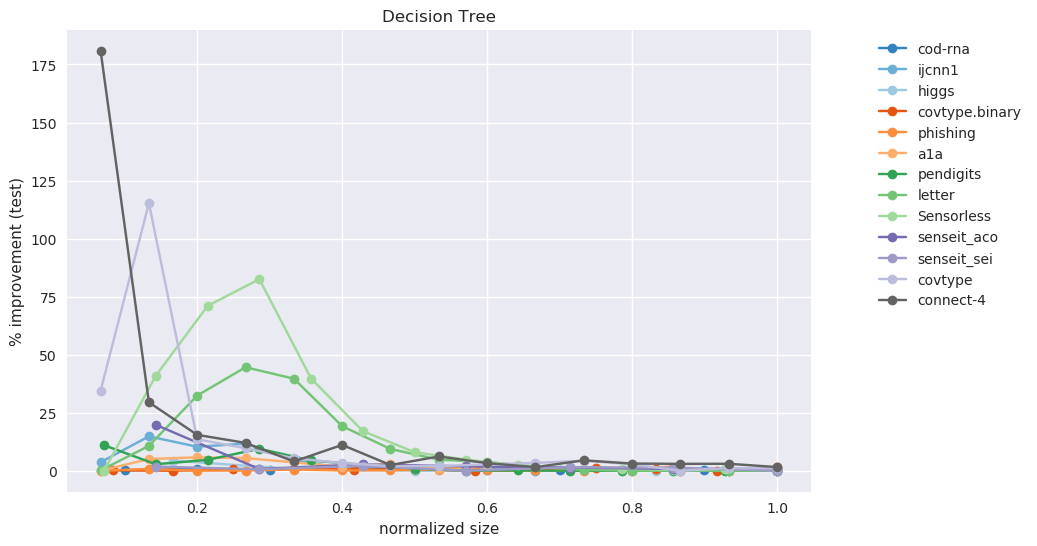}
    \caption{Improvement in F1 score on test with increasing size. Data in Table \ref{tab:dec_tree_improvements}.}
    \label{fig:dt_improvements}
\end{figure*}

This data is also visualized in Figure \ref{fig:dt_improvements}. The x-axis shows a scaled version of the actual tree depths for easy comparison: if the largest actual tree depth explored is $\eta_{max}$ for a dataset, then a size $\eta$ is represented by $\eta/\eta_{max}$. This allows us to compare a dataset like \emph{cod-rna}, which only has models up to a size of $10$, with $covtype$, where model sizes go all the way up to $15$.

We observe significant improvements in the F1-score for at least one model size for majority of the datasets. %  most that aside from the datasets \emph{cod-rna, covtype.binary, phishing}, we see significant improvement in the F1-score for at least one model size. 
The best improvements themselves vary a lot, ranging from $0.76\%$ for \emph{phishing} to $180.92\%$ for \emph{connect-4}.  
More so, these improvements seem to happen at small sizes: only one best score - for \emph{covtype.binary} - shows up on the right half of Table \ref{tab:dec_tree_improvements}. This is inline with Equations \ref{eqn:objective_breakup_a} and \ref{eqn:objective_breakup_b}: beyond a model size $\eta'$, $\delta F1=0\%$.

\begin{figure*}[h!]
    \centering
\includegraphics[scale=0.32]{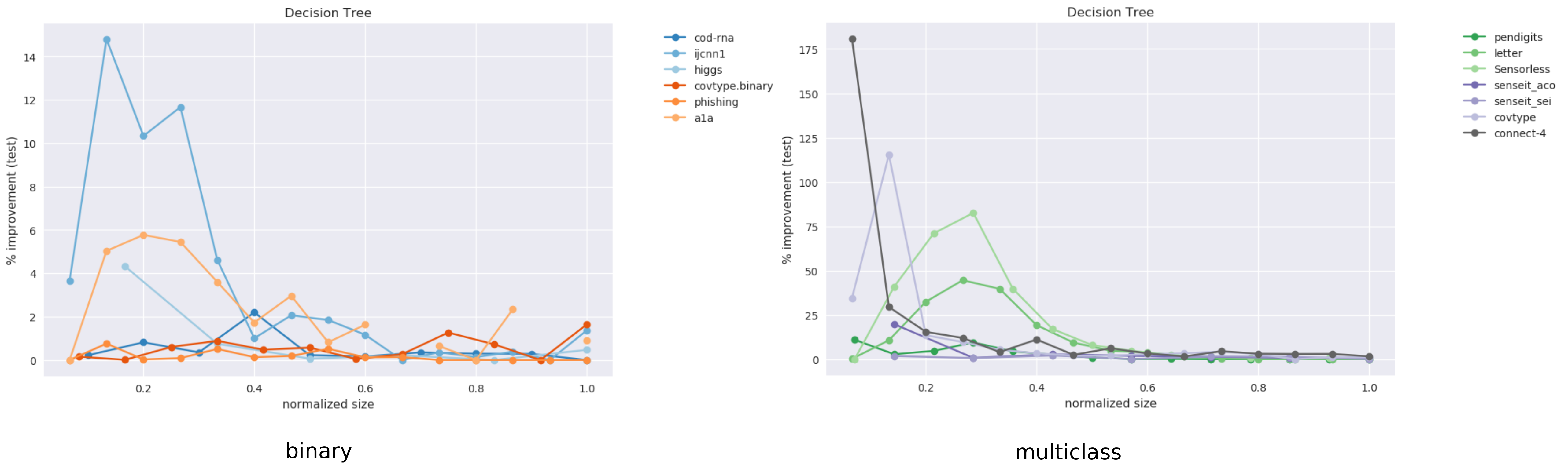}
    \caption{Performance on binary vs multi-class classification problems using CART. This is an elaboration of Figure \ref{fig:dt_improvements}.}
    \label{fig:binary_vs_multiclass}
\end{figure*}

It also seems that we do much better with multiclass data than with binary classes. Because of the large variance in improvements, this is hard to observe in Figure \ref{fig:dt_improvements}. However, if we separate the binary and multiclass results, as in Figure \ref{fig:binary_vs_multiclass}, we note that there are improvements in both the binary and multiclass cases, and the magnitude in the latter are typically higher (note the y-axes). We surmise this happens because, in general, DTs of a fixed depth have a harder problem to solve when the data is multiclass, providing our algorithm with an easier baseline to beat.

\begin{figure*}[h!]
    \centering
\includegraphics[scale=0.6]{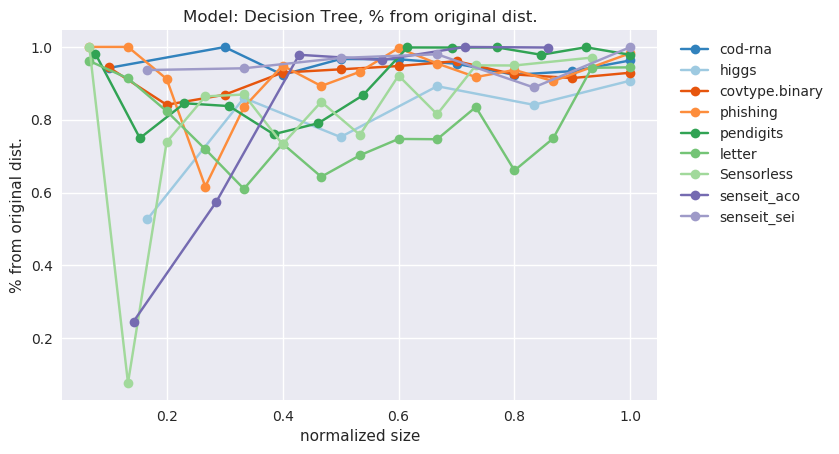}
    \caption{Variation of $p_o$ with increasing model size.}
    \label{fig:pct_orig}
\end{figure*}

Figure \ref{fig:pct_orig} shows the behavior of $p_o$, \emph{only} for the datasets where our models have grown to the optimal size (the last column in Table \ref{tab:dec_tree_improvements} for these datasets are empty). Thus, we exclude \emph{ijcnn1, a1a, covtype, connect-4}. We observe that indeed $p_o \to 1$ as our model grows to the optimal size. This empirically validates our hypothesis from Section \ref{ssec:intuition}, that \textbf{smaller models prefer a distribution different from the original distribution to learn from}, but the latter is optimal for larger models. And we gradually transition to it as model size increases.

Demonstrating this effect is a key contribution of our work.

We are also interested in knowing what the depth-distribution IBMM looks like. This is challenging to visualize for multiple datasets in one plot, since we have an optimal IBMM learned by our optimizer, for \emph{each} model size setting. We summarize this information for a dataset in the following manner:
\begin{enumerate}
    \item Pick a sample size of $N$ points to use. 
    \item We allocate points to sample from the IBMM for a particular model size, in proportion of $\delta F1$. For instance, if we have experimented with 3 model sizes, and $\delta F1$ are $7\%, 11\%$ and $2\%$, we sample $0.35N, 0.55N$ and $0.1N$ points respectively from the corresponding IBMMs.
    \item We fit a \emph{Kernel Density Estimator (KDE)} over these $N$ points, and plot the KDE curve. This plot represents the IBMM across model sizes for a dataset \emph{weighted} by the improvement seen for a size.
\end{enumerate}

$N$ should be large enough that the visualization is robust to sample variances. We use $N=10000$. %The value for the KDE \emph{bandwidth} parameter was arrived at by trial and error.

\begin{figure*}[h!]
    \centering
\includegraphics[scale=0.6]{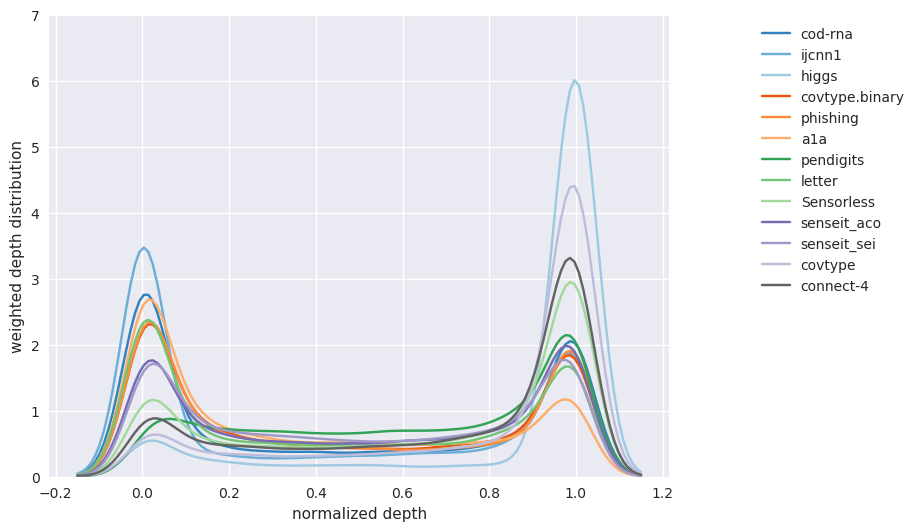}
    \caption{Distribution over levels in density tree(s). Aggregate of distribution over different model sizes.}
    \label{fig:dt_kde_density_depths}
\end{figure*}

Figure \ref{fig:dt_kde_density_depths} shows such a plot for DTs. The x-axis represents the depth of the density tree normalized to $[0, 1]$. The smoothing by the KDE causes some spillover beyond these bounds.

We observe that, in general, the depth distribution is concentrated either near the root of a density tree, where we have little or no information about class boundaries and the distribution is nearly identical to the original distribution, or at the leaves, where we have complete information of the class boundaries. An intermediate depth is relatively less used. This pattern in the depth distribution is \emph{surprisingly consistent} across all the models and datasets we have experimented with. We hypothesize this might be because of the following reasons:
\begin{enumerate}
    \item The information provided at an intermediate depth - where we have moved away from the original distribution, but have not yet completely discovered the class boundaries - might be relatively noisy to be useful.
    \item The model can selectively generalize well enough from the complete class boundary information at the leaves.
\end{enumerate}
 
Note that while fewer samples are drawn at intermediate depths, the number is not always insignificant - as an example, see \emph{pendigits} in Figure \ref{fig:dt_kde_density_depths}; hence using a distribution across the height of the density tree is still a useful strategy.

\begin{table*}\scriptsize
\centering
\caption{Classification Results with LPMs. Values indicate improvements $\delta F1$.}\label{tab:lpm_improvements}
\hspace{0.5pt}

\resizebox{\textwidth}{!}{\begin{tabular}{lrrrrrrrrrrrrrrrrr}
\toprule
\# terms =  &     1  &    2  &    3  &    4  &    5  &    6  &    7  &    8  &    9  &    10 &    11 &    12 &    13 &    14 &    15 \\
\cmidrule(lr){1-1}
datasets       &        &       &       &       &       &       &       &       &       &       &       &       &       &       &       \\
\midrule
cod-rna        &  26.87 & 28.25 & 53.71 & 57.71 & \textbf{58.08} & 35.82 & 20.05 &  4.36 &     - &     - &     - &     - &     - &     - &     - \\
ijcnn1         &  \textbf{17.25} &  8.86 &  0.51 &  1.43 &  1.86 &  1.35 &  0.94 &  1.79 &  2.04 &  1.31 &  0.61 &  1.95 &  2.01 &  0.37 &  2.48 \\
higgs          &   0.00 &  0.04 &  0.97 &  1.75 &  2.53 &  3.42 &  3.36 &  2.99 &  3.48 &  4.90 & \textbf{ 6.12} &  4.99 &  4.69 &  4.33 &  4.61 \\
covtype.binary &   0.13 &  2.22 &  4.24 &  6.73 &  9.40 & 12.29 & \textbf{14.67 }&  7.66 &  9.70 & 10.06 & 11.60 & 12.43 & 10.73 & 11.44 &  9.08 \\
phishing       &   0.00 &  0.00 &  0.00 &  0.00 &  0.11 &  0.13 & \textbf{ 0.60} &  0.00 &  0.02 &  0.27 &  0.18 &  0.39 &  0.41 &  0.58 &  0.60 \\
a1a            &   0.00 & 20.62 &\textbf{ 41.00} & 29.02 & 22.53 & 12.10 &  9.03 & 13.28 & 10.08 &  4.33 &  4.24 &  3.14 &  2.55 &  0.61 &  1.96 \\
\midrule
pendigits      &  10.08 &  8.54 & \textbf{10.09 }&  6.14 &  8.36 &  3.83 &  4.80 &  0.67 &  1.06 &  0.48 &  0.46 &  0.46 &  1.14 &  0.37 &  0.37 \\
letter         &  12.36 &  8.90 & \textbf{22.35} & 12.27 &  9.33 &  2.87 &  0.94 &  1.42 &  1.58 &  3.05 &  1.60 &  1.57 &  3.97 &  6.70 &  5.57 \\
Sensorless     &  \textbf{71.01} & 57.69 & 31.92 & 15.78 & 17.17 & 18.79 & 22.40 & 31.14 & 28.49 & 31.70 & 29.46 & 28.32 & 36.58 & 39.47 & 32.59 \\
senseit\_aco    &   7.07 &\textbf{ 56.04 }& 42.17 & 21.55 & 15.74 & 13.46 & 12.79 &  7.71 &  4.34 &  4.83 &  2.84 &  2.96 &  2.56 &  2.03 &  1.87 \\
senseit\_sei    & \textbf{143.97} & 46.19 & 20.28 &  7.56 &  2.71 &  1.01 &  1.04 &  1.75 &  0.87 &  0.79 &  1.47 &  1.60 &  1.30 &  0.27 &  0.42 \\
covtype        &  \textbf{30.10} & 20.05 &  4.68 &  3.24 &  1.24 &  6.46 &  2.57 &  4.66 &  5.43 &  5.03 &  6.92 &  6.02 &  2.29 &  9.65 &  9.03 \\
connect-4      & \textbf{117.62} & 32.09 & 20.29 & 17.47 &  7.07 &  6.41 &  6.15 &  5.67 &  6.80 &  4.72 &  3.87 &  1.82 &  1.62 &  0.12 &  2.31 \\
\bottomrule
\end{tabular}}

\end{table*}

\subsubsection{LPM Results}
The results for LPM are shown in Table \ref{tab:lpm_improvements}.
The improvements look different from what we observed for DT, which is to be expected across different model families. Notably, compared to DTs, there is no prominent disparity in  the improvements between binary class and multiclass datasets. Since the LPM builds \emph{one-vs-rest} binary classifiers in the multiclass case, and the size restriction - number of terms - applies to each individually, this intuitively makes sense. This is unlike DTs where the size constraint was applied to a single multiclass classifier. However, much like DTs, we still observe the pattern of the greatest improvements occurring at relatively smaller model sizes.

\begin{figure*}[h!]
    \centering
\includegraphics[scale=0.35]{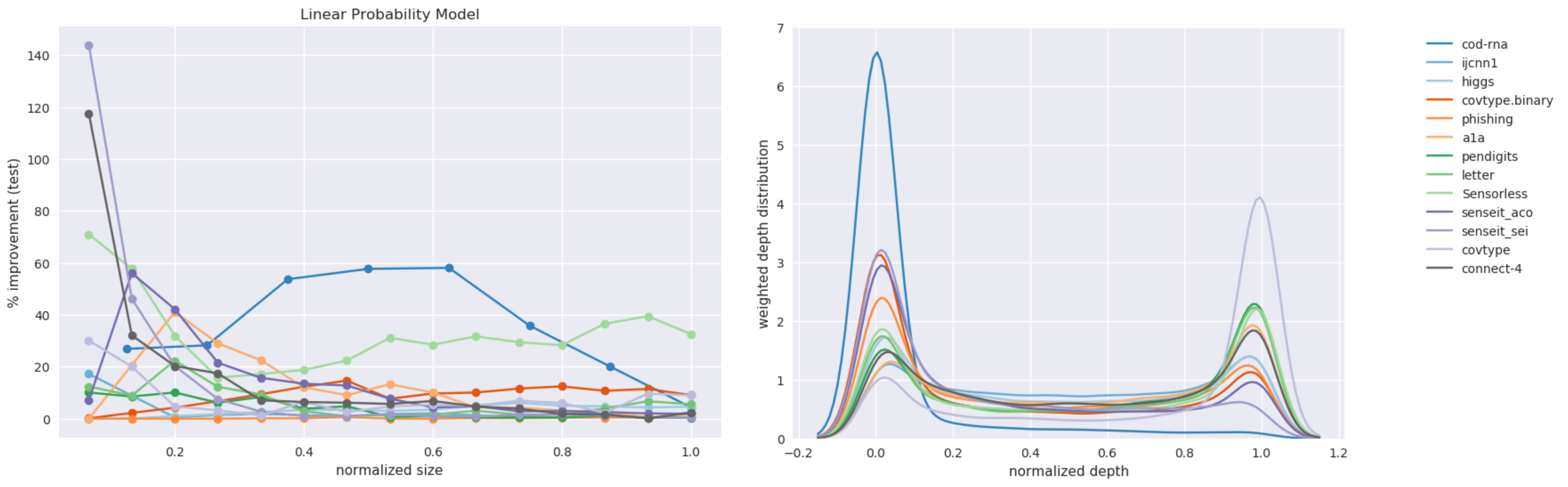}
    \caption{Linear Probability Model: improvements and the distribution over depths of the density trees.}
        \label{fig:lpm_combined}
\end{figure*}

Figure \ref{fig:lpm_combined} shows the plots for improvement in the F1-score and the weighted depth distribution. 
The depth distribution plot displays concentration near the root and the leaves, similar to the case of the DT in Figure \ref{fig:dt_kde_density_depths}.

Note that unlike the case of the DT, we haven't determined how many terms the optimal model for a dataset has; we explore up to $min(d,15)$. Nevertheless, as in the case of DTs, we note the pattern that the best improvements typically occur at smaller sizes: only \emph{higgs} exhibits its largest improvements at a relatively large model size in Table \ref{tab:lpm_improvements}.

\subsubsection{GBM Results}
An interesting question to ask is how, if at all, the \emph{bias} of the model family of $\mathcal{F}$ in Algorithm \ref{algo:density_tree_opt}, influences the improvements in accuracy. We cannot directly compare DTs with LPMs since we don't know how to order models from different families: we cannot decide how large a DT to compare to a LPM with, say, $4$ non-zero terms. 

To answer this question we look at GBMs where we identify two levers to control the model size. We consider two different GBM models - with the $max\_depth$ of base classifier trees as $2$ and $5$ respectively. The number of boosting rounds is taken as the size of the classifier and is varied from $1$ to $10$. We refer to the GBMs with base classifiers with $max\_depth=2$ and $max\_depth=5$ as representing weak and strong model families respectively.

We recognize that qualitatively there are two opposing factors at play:
\begin{enumerate}
    \item A weak model family implies it might not learn sufficiently well from the samples our technique produces. Hence, we expect to see smaller improvements than when using a stronger model family.
    \item A weak model family implies there is a lower baseline to beat. Hence, we expect to see larger improvements.
\end{enumerate}

We present an abridged version of the GBM results in Table \ref{tab:gbdt_improvements_abridged} in the interest of space. The complete results are made available in Table \ref{tab:gbdt_improvements} in the \emph{Appendix}. We present both the improvement in the $F1$ score, $\delta F1$, and its new value, $F1_{new}$.

\begin{table*}\scriptsize
\centering
\caption{Classification Results with GBMs. Both $F1_{new}$ and $\delta F1$ are shown.}\label{tab:gbdt_improvements_abridged}
\hspace{0.5pt}

\resizebox{\textwidth}{!}{\begin{tabular}{lclrrrrrrrrrrrrrrr}
\toprule
  & & boosting rounds = &     1  &    2  &    3  &    4  &    5  &    6  &    7  &    8  &    9  &    10 \\
  \cmidrule(lr){3-3}
datasets & max depth & score type       &        &       &       &       &       &       &       &       &       &              \\
\midrule
Sensorless & 2 & $F1$ & 0.76 & 0.77 & 0.78 & 0.80 & 0.80 & 0.80 & 0.81 & 0.81 & 0.81 & 0.82\\
 &  & $\delta F1$ &  \cellcolor{blue!25} 3.49 & 4.01 & 3.97 & 6.03 & 3.25 & 1.54 & 3.28 & 2.51 & 2.34 & 2.72\\
 & 5 & $F1$ & 0.91 & 0.92 & 0.93 & 0.93 & 0.94 & 0.94 & 0.94 & 0.94 & 0.95 & 0.95\\
 &  & $\delta F1$ &  \cellcolor{blue!25}0.46 & 0.43 & 0.27 & 0.44 & 0.00 & 0.30 & 0.62 & 0.47 & 0.00 & 0.31\\
\midrule
senseit\_aco & 2 & $F1$ & 0.22 & 0.25 & 0.36 & 0.38 & 0.50 & 0.59 & 0.61 & 0.62 & 0.63 & 0.64\\
 &  & $\delta F1$ & 0.00 & 11.36 & \cellcolor{red!25} 61.01 & 73.01 & \cellcolor{blue!25} 93.62 & 13.79 & 7.36 & 5.47 & 3.11 & 1.28\\
 & 5 & $F1$ & 0.22 & 0.33 & 0.45 & 0.54 & 0.60 & 0.63 & 0.65 & 0.66 & 0.67 & 0.68\\
 &  & $\delta F1$ & 0.00 & 47.50 & \cellcolor{red!25} 100.27 & 48.79 & \cellcolor{blue!25} 11.39 & 4.01 & 0.68 & 0.54 & 0.21 & 0.43\\
\midrule
senseit\_sei & 2 & $F1$ & 0.60 & 0.60 & 0.60 & 0.61 & 0.61 & 0.61 & 0.61 & 0.61 & 0.61 & 0.61\\
 &  & $\delta F1$ & 170.10 & \cellcolor{red!25} 168.56 & 171.75 & 173.56 & 172.43 & 167.41 & 95.59 & 49.62 & 26.21 & 16.73\\
 & 5 & $F1$ & 0.63 & 0.64 & 0.64 & 0.64 & 0.65 & 0.65 & 0.65 & 0.65 & 0.65 & 0.66\\
 &  & $\delta F1$ & 181.80 & \cellcolor{red!25} 186.11 & 186.60 & 186.14 & 65.48 & 28.77 & 13.00 & 4.51 & 1.06 & 0.00\\
\bottomrule
\end{tabular}}
 \vspace{1ex}

     \raggedright See Table \ref{tab:gbdt_improvements} for complete data, Fig \ref{fig:gbdt2_combined} and Fig \ref{fig:gbdt5_combined} for plots.
\end{table*}

\begin{figure*}[h!]
    \centering
    \includegraphics[scale=0.35]{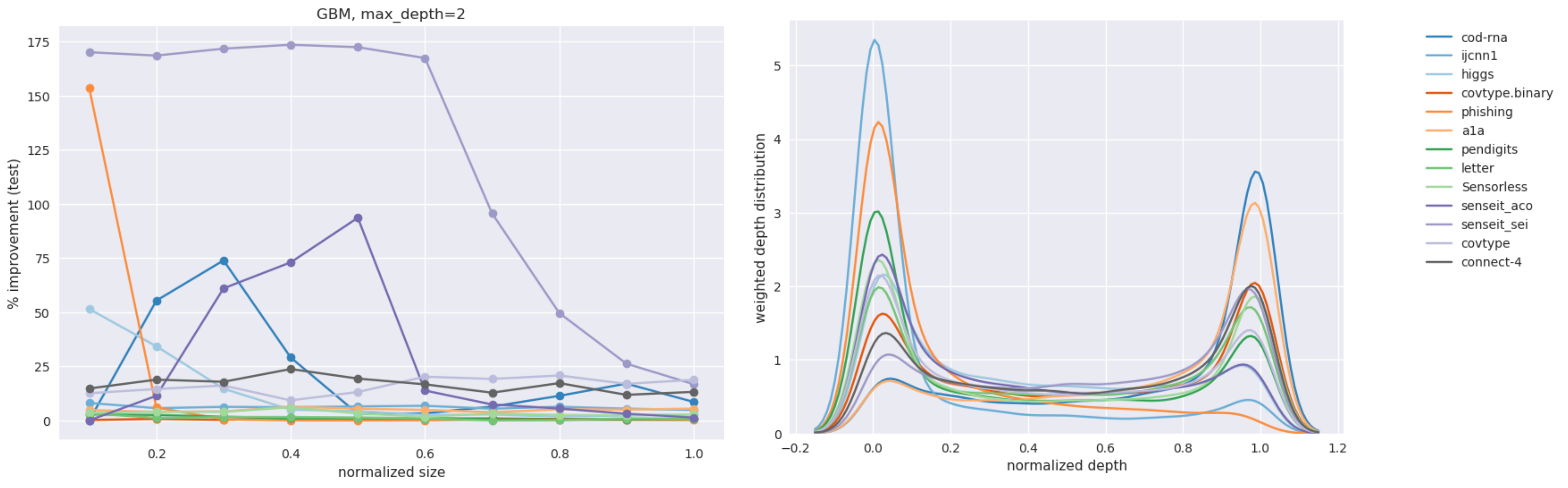}
    \caption{GBM with $max\_depth=2$. Size is the number of rounds.}
    \label{fig:gbdt2_combined}
\end{figure*}

\begin{figure*}[h!]
    \centering
    \includegraphics[scale=0.36]{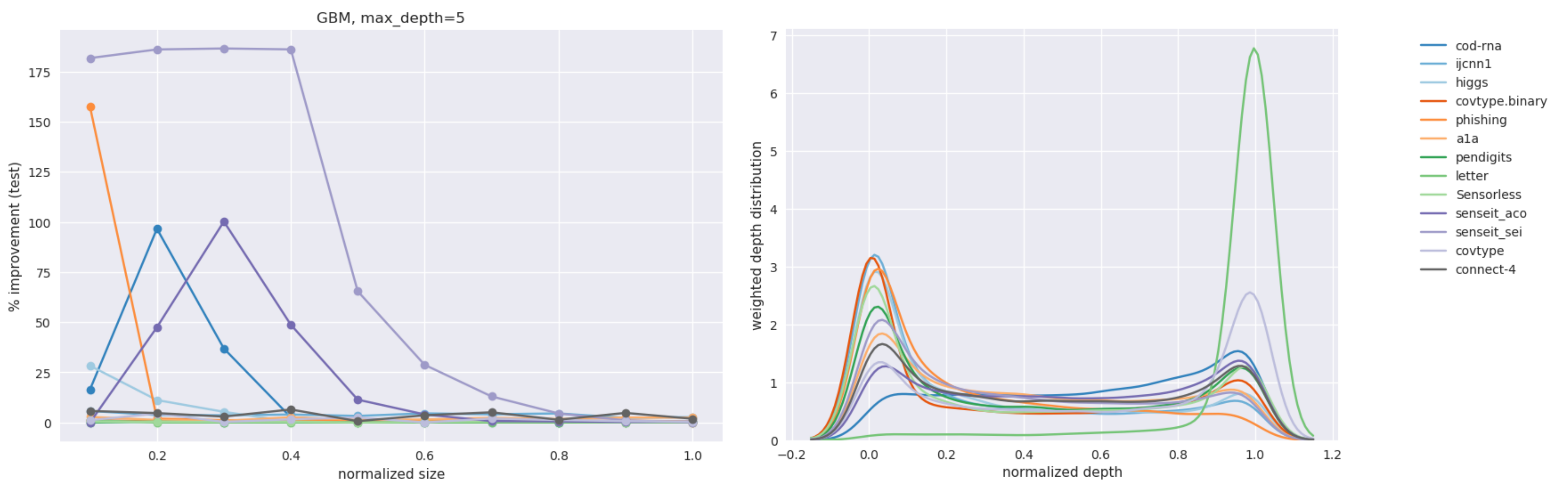}
    \caption{GBM with $max\_depth=5$.  Size is the number of rounds.}
    \label{fig:gbdt5_combined}
\end{figure*}

Figure \ref{fig:gbdt2_combined} and Figure \ref{fig:gbdt5_combined} show the improvement and depth distribution plots for the GBMs with $max\_depth=2$ and $max\_depth=5$ respectively.

The cells highlighted in blue in Table \ref{tab:gbdt_improvements_abridged} are where the GBM with $max\_depth=2$ showed a larger improvement than a GBM with $max\_depth=5$ for the same number of boosting rounds. The cells highlighted in red exhibit the opposite case. Clearly, both factors manifest themselves. Comparing the relative improvement plots in Figure \ref{fig:gbdt2_combined} and Figure \ref{fig:gbdt5_combined}, we see that improvements continue up to larger sizes when $max\_depth=2$ (also evident from Table \ref{tab:gbdt_improvements_abridged}). This is not surprising: we expect a stronger model to extract patterns from data at relatively smaller sizes, compared to a weaker model.

Observe that in Table \ref{tab:gbdt_improvements_abridged}, for the same number of boosting rounds, the new scores $F1_{new}$ for the weaker GBMs are up to as large (within some margin of error) as the scores for the stronger GBMs. This is to be expected since our sampling technique diminishes the gap between representational and effective capacities (when such a gap exists); it does not improve the representational capacity itself. Hence a weak classifier using our method is not expected to outperform a strong classifier that is also using our method. %However, this opens up an interesting line for future enquiry: is the extent to which this gap is decreased similar across different model families? In the case of our GBM experiments, where the only difference between the two experiments is the size of the base classifiers, the effectiveness might be reasonably assumed to be the same. But how about disparate model families like LPM and DTs? 

The depth distribution plots for the GBMs show a familiar pattern: high concentration at the root or the leaves. Also, similar to DTs and LPMs, the greatest improvements for a dataset mostly occur at relatively smaller model sizes - see Table \ref{tab:gbdt_improvements}.

% \subsubsection{Common Patterns}
% \label{sssec:common_patterns}
% We noticed in the previous sections the depth distributions appear remarkably similar across different $train_{\mathcal{F}}()$. Figure \ref{fig:kde_density_depths_comparison} brings them together for comparison. The y-axes are scaled to have the same range.

% \begin{figure*}[h!]
%     \centering
%     \includegraphics[scale=0.39]{kde_density_depths_comparison.pdf}
%     \caption{Weighted depth distributions: simpler classifiers tend to not use boundary information.}
%     \label{fig:kde_density_depths_comparison}
% \end{figure*}

% It appears that stronger model families, in general, tend to prefer a concentration near leaves compared to weaker model families. Compare DT and LPM, where DT is the stronger model family. There is a clear higher concentration near the leaves for the DT, and a clear higher concentration near the root for the LPM. Between the two categories of GBM models, we see a milder version of this phenomena: the models with $max\_depth=2$ prefer the root, and while the models with $max\_depth=5$ do not entirely prefer the leaves, there is a distinctive shift away from the root. We believe this happens because low bias classifiers are relatively better able to assimilate the richer information at the leaves.

\subsubsection{Summary}
Summarizing our analysis above:
\begin{enumerate}
    \item We see significant improvements in the $F1$ score across multiple combinations of model families, model sizes and datasets.
    \item Since in the DT experiments, we have multiple datasets for which we reached the optimal tree size, we were able to empirically validate the following related key hypotheses:
    \begin{enumerate}
        \item With larger model sizes the optimal distribution tends towards the original distribution. This is conveniently indicated with $p_o \to 1$ as $\eta$ increases.
        \item There is model size $\eta'$, beyond which $\delta F1 \approx 0\%$.
    \end{enumerate}
    \item For all the model families experimented with - DTs, LPMs, GBMs (results in Table \ref{tab:gbdt_improvements}) - the greatest improvements are seen for relatively smaller model sizes. 
    \item In the case of DTs, the improvements are, in general, higher with multiclass than binary datasets. We do not see this disparity for LPMs. We believe this happens because of our subjective notion of size: in the case of DTs there is a single tree to which the size constraint applies, making the baseline easier to beat for multiclass problems; while for LPMs it applies to each \emph{one-vs-rest} linear model.
    
    Its harder to characterize the behavior of the GBMs in this regard, since while the base classifiers are DTs, each of which is a multiclass classifier, a GBM maybe comprised of multiple DTs.  
    \item The GBM experiments give us the opportunity to study the effect of using model families, $\mathcal{F}$, of different strengths. 
    We make the following observations:
    \begin{enumerate}
        \item We see both these factors at work: (1)  a weaker model family has an easier baseline to beat, which may lead to higher $\delta F1$ scores relative to using a stronger model family (2) a stronger model family is likely to make better use of the optimal distribution, which may lead to higher $\delta F1$ scores relative to using a weaker model family.
        \item For a stronger model family, the benefit of using our algorithm diminishes quickly as model size grows.
        \item While the improvement $\delta F1$ for a weaker family may exceed one for a stronger family, the improved score $F1_{new}$ may, at best, match it.
        
    \end{enumerate}
    \item The depth distribution seems to favour either nodes near the root or the leaves, and this pattern is consistent across learning algorithms and datasets. 
\end{enumerate}

Given our observations, \emph{we would recommend using our approach as a pre-processing step for any size limited learning}, regardless of whether the size is appropriately small for our technique to be useful or not. If the size is large, then our method will return to the original sample anyways.

\section{Discussion}
\label{sec:discuss}

In addition to empirically validating our algorithm, the previous section also provided us with an idea of the kind of results we might expect of it. Using that as a foundation, we revisit our algorithm in this section, to consider some of our design choices and possible extensions.

\subsection{Algorithm Design Choices}
Conceptually, Algorithm \ref{algo:density_tree_opt} consists of quite a few building blocks. Although we have justified our implementation choices for them in Section \ref{sec:methodology}, it is instructive to look at some reasonable alternatives.

\begin{enumerate}
    \item Since we use our depth distribution to identify the value of a depth $\in \mathbb{Z}_{\geq 0}$, a valid question is why not use a discrete distribution, e.g., a multinomial? Our reason for using a continuous distribution is that we can use a a fixed number of optimization variables to characterize a density tree of \emph{any} depth, with just an additional step of discretization. Also, recall that the depth distribution applies to \emph{all} density trees in the forest $B$, each of which may have a different depth. A continuous distribution affords us the convenience of not having to deal with them individually. 
    \item A good candidate for the depth distribution is the \emph{Pitman-Yor} process \citep{pitman1997} - a two-parameter generalization of the DP (recall, this has one parameter: $\alpha$). Considering our results in 
    Figures \ref{fig:dt_kde_density_depths},  \ref{fig:lpm_combined}, \ref{fig:gbdt2_combined}, \ref{fig:gbdt5_combined}, where most depth distributions seem to have up to two dominant modes, we did not see a strong reason to use a more flexible distribution at the cost of introducing an optimization variable.
    \item We considered using the \emph{Kumaraswamy} distribution \citep{KUMARASWAMY198079} instead of $Beta$ for the mixture components. The advantage of the former is its \emph{cumulative distribution function} maybe be expressed as a simple formula, which leads to fast sampling. However, our tests with a Python implementation of the function showed us no significant benefit over the $Beta$ in the \emph{SciPy} package, for our use case: the depth distribution is in one dimension, and we draw samples in batches (all samples for a component are drawn simultaneously). Consequently, we decided to stick to the more conventional $Beta$ distribution\footnote{Interestingly, another recent paper on interpretability does use the Kumaraswamy distribution \citep{bastings-etal-2019-interpretable}.}. 
\end{enumerate}

%The previous section exclusively looked at experiments that validate our hypothesis; however, 

\subsection{Extensions and Applications}
\label{ssec:extn}

Our algorithm is reasonably abstracted from low level details, which enables various extensions and applications. We list some of these below:

\begin{enumerate}
    \item \textbf{Smoothing}: We had hinted at alternatives to Laplace smoothing in Section \ref{sec:preprocess_dt}. We discuss one possibility here. Assuming our density tree has $n$ nodes, we let $S \in \mathbb{R}^{n \times n}$ denote a pairwise \emph{similarity matrix} for these nodes, i.e., $[S]_{ij}$ is the similarity score between nodes $i$ and $j$. Let $P \in \mathbb{R}^{1 \times n}$ denote the base (i.e. before smoothing) probability masses for the nodes. Normalizing $P \times S^k, k \in \mathbb{Z}_{\geq 0}$ gives us a smoothed \emph{pmf} that is determined by our view of similarity between nodes. Analogous to \emph{transition matrices}, the exponent $k$ determines how diffuse the the similarity is; this can replace $\lambda$ as an optimization variable.
    
    The ability to incorporate a node similarity matrix opens up a wide range of possibilities, e.g., $S$ might be based on the \emph{Wu-Palmer} distance \citep{Wu:1994:VSL:981732.981751}, \emph{SimRank} \citep{Jeh:2002:SMS:775047.775126} or \emph{Random Walk with Restart (RWR)} \citep{Pan:2004:AMC:1014052.1014135}.

    \item  \textbf{Categorical variables}: We have not explicitly discussed the case of categorical features. There are a couple of ways to handle data with such features:
    \begin{enumerate}
        \item The density tree may directly deal with categorical variables. When sampling uniformly from a node that is defined by conditions on both continuous and categorical variables, we need to combine the outputs of a continuous uniform sampler (which we use now) and a discrete uniform sampler (i.e. multinomial with equal masses) for the respective feature types.
        \item We could create a version of the data with one-hot encoded categorical features for constructing the density tree. For input to $train_\mathcal{F}()$ at each iteration, we transform back the sampled data by identifying values for the categorical features to be the maximums in their corresponding sub-vectors. Since the optimizer already assumes a black-box $train_\mathcal{F}()$ function, this transformation would be modeled as a part of it. 
    \end{enumerate}

    \item  \textbf{Model compression}: An interesting possible use-case is model compression. Consider the column $boosting\;round=1$ for the \emph{senseit\_sei} dataset in Table \ref{tab:gbdt_improvements_abridged}. Assuming the base classifiers have grown to their $max\_depths$, the memory footprint in terms of nodes for the GBMs with $max\_depth=2$ and $max\_depth=5$ are $2^2+1=5$ and $2^5+1=33$ respectively. 
    %The $F1$ score of the latter, before applying our algorithm is $F1_{old}= 0.64/(1+1.8962)=0.22$. 
    
    Replacing the second model (larger) with the first (small) in a memory constrained system reduces footprint by $(33-5)/33 = 85\%$ at the cost of changing the $F1$ score by $(0.60-0.63)/0.63=-4.7\%$ only.
    
    Such a proposition becomes particularly attractive if we look at the baseline scores, i.e., accuracies on the original distribution. For the larger model, $F1_{baseline}= F1_{new}/(1+\delta F1/100)=0.63/(1+1.8180)=0.22$. If we replace this model with the smaller model enhanced by our algorithm, we not only reduce the footprint but actually \emph{improve} the $F1$ score by $(0.60-0.22)/0.22=173.7\%$!  
    
    We precisely state this application thus: our algorithm may be used to identify a model size $\eta_e$ (subscript ``e'' for ``equivalent'') in relation to a size $\eta > \eta_e$  such that:
    
    \begin{equation}
    \label{eqn:compression}
    accuracy(train_\mathcal{F}(p^*_{\eta_e}, \eta_e), p) \approx
        accuracy(train_\mathcal{F}(p, \eta), p)
    \end{equation}

    \item  \textbf{Segment analysis}: Our sampling operates within the bounding box $U \subset \mathbb{R}^d$; in previous sections, $U$ was defined by the entire input data. However, this is not necessary: we may use our algorithm on a subset of the data $V \subset U$, as long as $V$ is a hyperrectangle in $\mathbb{R}^{d'}$, $d' \leq d$. This makes our algorithm useful for applications like \emph{cohort analysis}, common in marketing studies, where the objective is to study the behaviour of a segment - say, based on age and income - within a larger population. Our algorithm is especially appropriate since traditionally such analyses have emphasized interpretability.
    
    %In the limit, shrinking the bounding box makes our algorithm functionally equivalent to the \emph{LIME} technique \citep{DBLP:journals/corr/RibeiroSG16}, which constructs an interpretable model for a single test point by sampling points in its neighborhood. 

    \item  \textbf{Multidimensional size}: The notion of size need not be a scalar. Our GBM experiments touch upon this possibility. The definition of size only influences how the call to $train_\mathcal{F}()$  \emph{internally executes}; Algorithm \ref{algo:density_tree_opt} itself is agnostic to this detail. This makes our technique fairly flexible. For ex, it is easy in our setup to vary both $max\_depth$ and \emph{number of boosting rounds} for GBMs. %However, we would need to define an ordering on $\eta$ that is useful for the problem.
    
    %What identifies as ``size'' decides how the results are to be interpreted, not how the sampling works.

    \item \textbf{Different optimizers}: As mentioned in Section \ref{sec:density}, the fact that our search space has no special structure implies the workings of the optimizer is decoupled from the larger sampling framework. This makes it easy to experiment with different optimizers. For ex, an interesting exercise might be to study the effect of the hybrid optimizer \emph{Bayesian Optimization with Hyperband (BOHB)} \citep{DBLP:conf/icml/FalknerKH18} when $train_{\mathcal{F}}()$ is an iterative learner; BOHB uses an early stopping strategy in tandem with Bayesian Optimization.

    \item  \textbf{Over/Under-sampling}: As the range of the sample size parameter $N_s$ is set by the user, the possibility of over/under-sampling is subsumed by our algorithm. For instance, if our dataset has $500$ points, and we believe that sampling up to $4$ times might help, we can simply set $N_s \in [500, 2000]$. Over/Under-sampling need not be explored as a separate strategy.
    
\end{enumerate}

\section{Conclusion}

Our work addresses the trade-off between interpretability and accuracy. The approach we take is to identify an optimal training distribution that often dramatically improves model accuracy for an arbitrary model family, especially when the model size is small. We believe this is the first such technique proposed. We have framed the problem of identifying this distribution as an optimization problem, and have provided a technique that is empirically shown to be useful across multiple learning algorithms and datasets. In addition to its practical utility, we believe this work is valuable in that it challenges the conventional wisdom that the optimal training distribution is the test distribution.

A unique property of our technique is that beyond a pre-processing step of constructing a DT, which we refer to as a density tree, the number of variables in the core optimization step does not depend on the dimensionality of the data; it uses a fixed set of eight variables. The density tree is used to determine a feasible space of distributions to search through, making the optimization efficient. Our choice of using DTs is innovative since while all classifiers implicitly identify boundaries, only few classifiers like DTs, rules, etc., can explicitly indicate their locations in the feature space. 
We have also discussed how our algorithm may be extended in some useful ways. 

We hope that the results presented here would motivate a larger discussion around the effect of training distributions on model accuracy.

\newpage
\appendix
\section{Appendix}

\subsection{Implementation Details}
\label{sec:impl}
Setting the lower bound of the $N_s$ parameter (see Section \ref{sec:preprocess_dt}) to ensure statistical significance is not sufficient in itself. Since our sample comes from both the density trees and the original training data, we must ensure these samples lead to statistically significant results \emph{individually}.

In order to do so the our implementation internally adjusts the quantity $p_o$. Recall that $p_o \in [0, 1]$. A low value of $p_o$ can result in a small sample of size $p_o N_s$ from the original training data, while a high value of $p_o$ might result in a small sample of size $(1-p_o) N_s$ from the density trees. Interestingly however, $p_o=0$ and $p_o=1$ should be allowed as valid values, since the sample is then drawn from only one of the sources and is therefore not small!

\begin{figure*}[h!]
    \centering
    \includegraphics[scale=0.65]{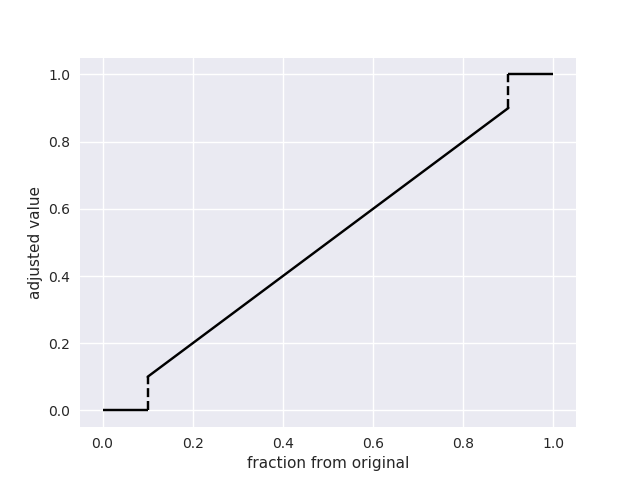}
    \caption{Adjustments to $p_o$.}
    \label{fig:p_adjustment}
\end{figure*}

The adjustment we make is shown in Figure \ref{fig:p_adjustment}. The x-axis shows the current value of $p_o$, the y-axis shows what the sampler sees. 

Below a user specified threshold for $p_o$, it is adjusted to $p_o=0$. Beyond a certain user specified threshold, it is adjusted to $p_o=1$. The lack of smoothness or differentiability of the adjustment does not impact our optimization, since a BO would construct its version of the objective function anyway.
 
%An additional consideration is the possible variance in model performance due to training on a sample from the current $\Phi_t$ in Algorithm \ref{algo:density_tree_opt}. We address this issue by taking multiple samples at each iteration, training one model per sample and then reporting the average accuracy across the models. For our experiments, we sampled $3$ times per iteration. 

\newpage
\subsection{GBM Results}
\label{sec:additional_gbm}
Table \ref{tab:gbdt_improvements} represents the improvements seen using GBMs where we have $max\_depth=2$ or $max\_depth=5$ for the base classifier trees. This is an expanded version of data presented in Table \ref{tab:gbdt_improvements_abridged}. Note here that much like DTs and LPMs, we see the largest $\delta F1$ values typically for relatively smaller model sizes.

\begin{table*}\scriptsize
\centering
\caption{GBM, $F1_{new}$ and $\delta F1$}\label{tab:gbdt_improvements}
\hspace{0.5pt}

\resizebox{\textwidth}{!}{\begin{tabular}{lclrrrrrrrrrrrrrrr}
\toprule
  & & boosting rounds = &     1  &    2  &    3  &    4  &    5  &    6  &    7  &    8  &    9  &    10 \\
  \cmidrule(lr){3-3}
datasets & max depth & score type       &        &       &       &       &       &       &       &       &       &              \\
\midrule
cod-rna & 2 & $F1$ & 0.40 & 0.62 & 0.70 & 0.70 & 0.71 & 0.71 & 0.75 & 0.78 & 0.83 & 0.85\\
 &  & $\delta F1$ & 0.00 & 55.35 & \textbf{73.91} & 29.06 & 3.03 & 3.27 & 6.38 & 11.35 & 17.00 & 8.54\\
 & 5 & $F1$ & 0.47 & 0.79 & 0.83 & 0.85 & 0.86 & 0.87 & 0.88 & 0.88 & 0.89 & 0.89\\
 &  & $\delta F1$ & 16.47 & \textbf{96.65} & 36.91 & 2.61 & 0.41 & 0.23 & 1.02 & 0.07 & 0.17 & 0.18\\
\midrule
ijcnn1 & 2 & $F1$ & 0.71 & 0.71 & 0.71 & 0.71 & 0.71 & 0.72 & 0.72 & 0.72 & 0.71 & 0.72\\
 &  & $\delta F1$ & \textbf{8.08} & 5.65 & 6.25 & 6.10 & 6.46 & 6.76 & 5.39 & 6.30 & 5.42 & 4.83\\
 & 5 & $F1$ & 0.76 & 0.77 & 0.76 & 0.77 & 0.78 & 0.78 & 0.79 & 0.79 & 0.79 & 0.79\\
 &  & $\delta F1$ & \textbf{5.73} & 3.73 & 3.59 & 3.98 & 3.28 & 4.54 & 4.09 & 4.51 & 2.31 & 2.66\\
\midrule
higgs & 2 & $F1$ & 0.61 & 0.61 & 0.63 & 0.63 & 0.63 & 0.63 & 0.63 & 0.63 & 0.64 & 0.64\\
 &  & $\delta F1$ & \textbf{51.47} & 34.24 & 14.65 & 4.93 & 4.30 & 2.54 & 2.68 & 1.63 & 2.22 & 2.65\\
 & 5 & $F1$ & 0.62 & 0.64 & 0.64 & 0.65 & 0.65 & 0.66 & 0.67 & 0.66 & 0.67 & 0.68\\
 &  & $\delta F1$ &\textbf{ 28.31} & 11.18 & 5.31 & 0.37 & 0.66 & 1.20 & 0.00 & 1.31 & 0.42 & 0.21\\
\midrule
covtype.binary & 2 & $F1$ & 0.72 & 0.72 & 0.72 & 0.73 & 0.73 & 0.73 & 0.73 & 0.73 & 0.73 & 0.73\\
 &  & $\delta F1$ & 0.21 & 0.69 & 0.24 & 0.93 & 0.77 & 0.84 &\textbf{ 0.90} & 0.52 & 0.50 & 0.72\\
 & 5 & $F1$ & 0.75 & 0.76 & 0.76 & 0.76 & 0.76 & 0.77 & 0.77 & 0.77 & 0.77 & 0.77\\
 &  & $\delta F1$ & \textbf{1.01 }& 0.60 & 0.93 & 0.81 & 0.71 & 0.64 & 0.20 & 0.12 & 0.25 & 0.00\\
\midrule
phishing & 2 & $F1$ & 0.91 & 0.91 & 0.91 & 0.91 & 0.91 & 0.91 & 0.92 & 0.92 & 0.92 & 0.92\\
 &  & $\delta F1$ & \textbf{153.47} & 5.97 & 0.48 & 0.03 & 0.00 & 0.05 & 0.38 & 0.19 & 0.23 & 0.12\\
 & 5 & $F1$ & 0.92 & 0.93 & 0.93 & 0.93 & 0.93 & 0.93 & 0.94 & 0.93 & 0.94 & 0.94\\
 &  & $\delta F1$ & \textbf{157.47 }& 1.93 & 1.13 & 1.13 & 1.23 & 0.82 & 0.83 & 0.60 & 0.91 & 0.48\\
\midrule
a1a & 2 & $F1$ & 0.71 & 0.72 & 0.72 & 0.73 & 0.73 & 0.73 & 0.74 & 0.73 & 0.73 & 0.74\\
 &  & $\delta F1$ & 4.77 & 3.93 & 4.12 & \textbf{6.30} & 5.39 & 4.84 & 3.77 & 5.10 & 4.98 & 5.43\\
 & 5 & $F1$ & 0.74 & 0.74 & 0.74 & 0.75 & 0.74 & 0.74 & 0.76 & 0.75 & 0.75 & 0.76\\
 &  & $\delta F1$ &\textbf{ 2.79} & 1.24 & 0.78 & 2.62 & 1.32 & 1.47 & 2.31 & 1.71 & 2.48 & 2.41\\
\midrule
\midrule
pendigits & 2 & $F1$ & 0.76 & 0.80 & 0.81 & 0.82 & 0.82 & 0.83 & 0.83 & 0.84 & 0.84 & 0.84\\
 &  & $\delta F1$ & \textbf{2.72 }& 2.49 & 1.66 & 0.97 & 1.04 & 0.82 & 0.52 & 0.42 & 0.48 & 0.67\\
 & 5 & $F1$ & 0.92 & 0.94 & 0.94 & 0.95 & 0.95 & 0.95 & 0.95 & 0.96 & 0.96 & 0.96\\
 &  & $\delta F1$ & 0.03 &\textbf{ 0.35 }& 0.00 & 0.20 & 0.02 & 0.04 & 0.18 & 0.00 & 0.03 & 0.12\\
\midrule
letter & 2 & $F1$ & 0.52 & 0.58 & 0.59 & 0.60 & 0.61 & 0.61 & 0.62 & 0.63 & 0.63 & 0.64\\
 &  & $\delta F1$ & \textbf{3.24 }& 1.10 & 1.53 & 1.52 & 1.30 & 0.90 & 0.00 & 0.12 & 0.91 & 0.72\\
 & 5 & $F1$ & 0.71 & 0.76 & 0.77 & 0.77 & 0.79 & 0.80 & 0.80 & 0.80 & 0.81 & 0.81\\
 &  & $\delta F1$ & \textbf{1.09} & 0.00 & 0.00 & 0.04 & 0.00 & 0.00 & 0.00 & 0.00 & 0.00 & 0.00\\
\midrule
Sensorless & 2 & $F1$ & 0.76 & 0.77 & 0.78 & 0.80 & 0.80 & 0.80 & 0.81 & 0.81 & 0.81 & 0.82\\
 &  & $\delta F1$ & 3.49 & 4.01 & 3.97 & \textbf{6.03} & 3.25 & 1.54 & 3.28 & 2.51 & 2.34 & 2.72\\
 & 5 & $F1$ & 0.91 & 0.92 & 0.93 & 0.93 & 0.94 & 0.94 & 0.94 & 0.94 & 0.95 & 0.95\\
 &  & $\delta F1$ & 0.46 & 0.43 & 0.27 & 0.44 & 0.00 & 0.30 &\textbf{ 0.62 }& 0.47 & 0.00 & 0.31\\
\midrule
senseit\_aco & 2 & $F1$ & 0.22 & 0.25 & 0.36 & 0.38 & 0.50 & 0.59 & 0.61 & 0.62 & 0.63 & 0.64\\
 &  & $\delta F1$ & 0.00 & 11.36 & 61.01 & 73.01 & \textbf{93.62 }& 13.79 & 7.36 & 5.47 & 3.11 & 1.28\\
 & 5 & $F1$ & 0.22 & 0.33 & 0.45 & 0.54 & 0.60 & 0.63 & 0.65 & 0.66 & 0.67 & 0.68\\
 &  & $\delta F1$ & 0.00 & 47.50 &\textbf{ 100.27} & 48.79 & 11.39 & 4.01 & 0.68 & 0.54 & 0.21 & 0.43\\
\midrule
senseit\_sei & 2 & $F1$ & 0.60 & 0.60 & 0.60 & 0.61 & 0.61 & 0.61 & 0.61 & 0.61 & 0.61 & 0.61\\
 &  & $\delta F1$ & 170.10 & 168.56 & 171.75 & \textbf{173.56} & 172.43 & 167.41 & 95.59 & 49.62 & 26.21 & 16.73\\
 & 5 & $F1$ & 0.63 & 0.64 & 0.64 & 0.64 & 0.65 & 0.65 & 0.65 & 0.65 & 0.65 & 0.66\\
 &  & $\delta F1$ & 181.80 & 186.11 & \textbf{186.60} & 186.14 & 65.48 & 28.77 & 13.00 & 4.51 & 1.06 & 0.00\\
\midrule
covtype & 2 & $F1$ & 0.41 & 0.41 & 0.42 & 0.41 & 0.40 & 0.40 & 0.40 & 0.40 & 0.40 & 0.39\\
 &  & $\delta F1$ & 12.56 & 14.50 & 16.13 & 9.24 & 13.12 & 20.20 & 19.16 & \textbf{20.75} & 16.87 & 18.79\\
 & 5 & $F1$ & 0.46 & 0.48 & 0.49 & 0.49 & 0.49 & 0.49 & 0.50 & 0.50 & 0.49 & 0.51\\
 &  & $\delta F1$ & 1.14 & \textbf{4.48 }& 0.27 & 1.57 & 2.52 & 0.00 & 1.96 & 1.20 & 0.65 & 0.44\\
\midrule
connect-4 & 2 & $F1$ & 0.43 & 0.44 & 0.45 & 0.45 & 0.46 & 0.46 & 0.46 & 0.47 & 0.47 & 0.47\\
 &  & $\delta F1$ & 14.75 & 18.83 & 17.84 & \textbf{23.74 }& 19.32 & 16.65 & 12.80 & 17.23 & 11.81 & 13.18\\
 & 5 & $F1$ & 0.49 & 0.49 & 0.50 & 0.52 & 0.51 & 0.52 & 0.52 & 0.53 & 0.53 & 0.53\\
 &  & $\delta F1$ & 5.74 & 4.68 & 2.94 & \textbf{6.47 }& 0.65 & 3.60 & 5.07 & 1.35 & 4.78 & 1.86\\

\bottomrule
\end{tabular}}

\end{table*}

\subsection{Harmonic Numbers}
\label{ssec:harmonic}
\begin{figure*}[h!]
    \centering
    \includegraphics[scale=0.5]{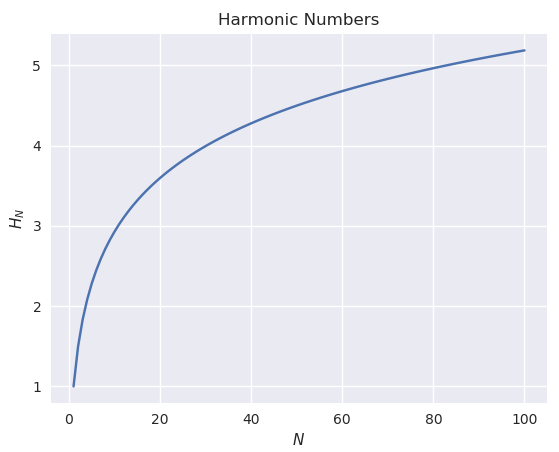}
    \caption{Variation of $H_N$ with increasing $N$.}
    \label{fig:harm}
\end{figure*}

The $N^{th}$ harmonic number is defined as:
\begin{equation}
    H_N = 1 + \frac{1}{2} + \frac{1}{3} + ... +   \frac{1}{N} = \sum_{k=1}^{N} \frac{1}{k}
\end{equation}
Clearly $H_N \propto N$, since increasing $N$ adds positive terms to $H_N$. Figure \ref{fig:harm} shows the relationship of $H_N$ and $N$ for $N = 1,2, ..., 100$

\bibliographystyle{frontiersinSCNS_ENG_HUMS} % for Science, Engineering and Humanities and Social Sciences articles, for Humanities and Social Sciences articles please include page numbers in the in-text citations
\bibliography{refs}

\begin{thebibliography}{73}
\providecommand{\natexlab}[1]{#1}
\expandafter\ifx\csname urlstyle\endcsname\relax
  \providecommand{\doi}[1]{doi:\discretionary{}{}{}#1}\else
  \providecommand{\doi}{doi:\discretionary{}{}{}\begingroup
  \urlstyle{rm}\Url}\fi
\providecommand{\selectlanguage}[1]{\relax}
\providecommand{\bibAnnoteFile}[1]{%
  \IfFileExists{#1}{\begin{quotation}\noindent\textsc{Key:} #1\\
  \textsc{Annotation:}\ \input{#1}\end{quotation}}{}}
\providecommand{\bibAnnote}[2]{%
  \begin{quotation}\noindent\textsc{Key:} #1\\
  \textsc{Annotation:}\ #2\end{quotation}}

\bibitem[{Alvi et~al.(2019)Alvi, Ru, Calliess, Roberts, and
  Osborne}]{pmlr-v97-alvi19a}
Alvi, A., Ru, B., Calliess, J.-P., Roberts, S., and Osborne, M.~A. (2019).
\newblock Asynchronous batch {B}ayesian optimisation with improved local
  penalisation.
\newblock In \emph{Proceedings of the 36th International Conference on Machine
  Learning}, eds. K.~Chaudhuri and R.~Salakhutdinov (Long Beach, California,
  USA: PMLR), vol.~97 of \emph{Proceedings of Machine Learning Research},
  253--262
\bibAnnoteFile{pmlr-v97-alvi19a}

\bibitem[{Ancona et~al.(2019)Ancona, Oztireli, and Gross}]{pmlr-v97-ancona19a}
Ancona, M., Oztireli, C., and Gross, M. (2019).
\newblock Explaining deep neural networks with a polynomial time algorithm for
  shapley value approximation.
\newblock In \emph{Proceedings of the 36th International Conference on Machine
  Learning}, eds. K.~Chaudhuri and R.~Salakhutdinov (Long Beach, California,
  USA: PMLR), vol.~97 of \emph{Proceedings of Machine Learning Research},
  272--281
\bibAnnoteFile{pmlr-v97-ancona19a}

\bibitem[{Angelino et~al.(2017)Angelino, Larus-Stone, Alabi, Seltzer, and
  Rudin}]{Angelino:2017:LCO:3097983.3098047}
Angelino, E., Larus-Stone, N., Alabi, D., Seltzer, M., and Rudin, C. (2017).
\newblock Learning certifiably optimal rule lists.
\newblock In \emph{Proceedings of the 23rd ACM SIGKDD International Conference
  on Knowledge Discovery and Data Mining} (New York, NY, USA: ACM), KDD '17,
  35--44.
\newblock \doi{10.1145/3097983.3098047}
\bibAnnoteFile{Angelino:2017:LCO:3097983.3098047}

\bibitem[{{Bachem} et~al.(2017){Bachem}, {Lucic}, and
  {Krause}}]{2017arXiv170306476B}
{Bachem}, O., {Lucic}, M., and {Krause}, A. (2017).
\newblock {Practical Coreset Constructions for Machine Learning}.
\newblock Preprint at
  \url{https://ui.adsabs.harvard.edu/abs/2017arXiv170306476B}
\bibAnnoteFile{2017arXiv170306476B}

\bibitem[{Bastings et~al.(2019)Bastings, Aziz, and
  Titov}]{bastings-etal-2019-interpretable}
Bastings, J., Aziz, W., and Titov, I. (2019).
\newblock Interpretable neural predictions with differentiable binary
  variables.
\newblock In \emph{Proceedings of the 57th Annual Meeting of the Association
  for Computational Linguistics} (Florence, Italy: Association for
  Computational Linguistics), 2963--2977.
\newblock \doi{10.18653/v1/P19-1284}
\bibAnnoteFile{bastings-etal-2019-interpretable}

\bibitem[{Bergstra et~al.(2011)Bergstra, Bardenet, Bengio, and
  K{\'e}gl}]{Bergstra:2011:AHO:2986459.2986743}
Bergstra, J., Bardenet, R., Bengio, Y., and K{\'e}gl, B. (2011).
\newblock Algorithms for hyper-parameter optimization.
\newblock In \emph{Proceedings of the 24th International Conference on Neural
  Information Processing Systems} (USA: Curran Associates Inc.), NIPS'11,
  2546--2554
\bibAnnoteFile{Bergstra:2011:AHO:2986459.2986743}

\bibitem[{Bergstra et~al.(2013)Bergstra, Yamins, and
  Cox}]{Bergstra:2013:MSM:3042817.3042832}
Bergstra, J., Yamins, D., and Cox, D.~D. (2013).
\newblock Making a science of model search: Hyperparameter optimization in
  hundreds of dimensions for vision architectures.
\newblock In \emph{Proceedings of the 30th International Conference on
  International Conference on Machine Learning - Volume 28} (JMLR.org),
  ICML'13, I--115--I--123
\bibAnnoteFile{Bergstra:2013:MSM:3042817.3042832}

\bibitem[{Blackwell and MacQueen(1973)}]{blackwell1973}
Blackwell, D. and MacQueen, J.~B. (1973).
\newblock Ferguson distributions via polya urn schemes.
\newblock \emph{Ann. Statist.} 1, 353--355.
\newblock \doi{10.1214/aos/1176342372}
\bibAnnoteFile{blackwell1973}

\bibitem[{Blaser and Fryzlewicz(2016)}]{JMLR:v17:blaser16a}
Blaser, R. and Fryzlewicz, P. (2016).
\newblock Random rotation ensembles.
\newblock \emph{Journal of Machine Learning Research} 17, 1--26
\bibAnnoteFile{JMLR:v17:blaser16a}

\bibitem[{{Blei}(2007)}]{bayes_notes}
{Blei}, D. (2007).
\newblock {COS 597C Notes, Bayesian Nonparametrics}.
\newblock
  \url{https://www.cs.princeton.edu/courses/archive/fall07/cos597C/scribe/20070921.pdf}
\bibAnnoteFile{bayes_notes}

\bibitem[{Breiman et~al.(1984)}]{cart93}
Breiman, L. et~al. (1984).
\newblock \emph{{Classification and Regression Trees}} (New York: Chapman \&
  Hall)
\bibAnnoteFile{cart93}

\bibitem[{Brochu et~al.(2010)Brochu, Cora, and de~Freitas}]{Brochu2010ATO}
Brochu, E., Cora, V.~M., and de~Freitas, N. (2010).
\newblock A tutorial on bayesian optimization of expensive cost functions, with
  application to active user modeling and hierarchical reinforcement learning.
\newblock \emph{CoRR} abs/1012.2599
\bibAnnoteFile{Brochu2010ATO}

\bibitem[{Caruana et~al.(2015)Caruana, Lou, Gehrke, Koch, Sturm, and
  Elhadad}]{Caruana:2015:IMH:2783258.2788613}
Caruana, R., Lou, Y., Gehrke, J., Koch, P., Sturm, M., and Elhadad, N. (2015).
\newblock Intelligible models for healthcare: Predicting pneumonia risk and
  hospital 30-day readmission.
\newblock In \emph{Proceedings of the 21th ACM SIGKDD International Conference
  on Knowledge Discovery and Data Mining} (New York, NY, USA: ACM), KDD '15,
  1721--1730.
\newblock \doi{10.1145/2783258.2788613}
\bibAnnoteFile{Caruana:2015:IMH:2783258.2788613}

\bibitem[{Chang and Lin(2011)}]{CC01a}
Chang, C.-C. and Lin, C.-J. (2011).
\newblock {LIBSVM}: A library for support vector machines.
\newblock \emph{ACM Transactions on Intelligent Systems and Technology} 2,
  27:1--27:27.
\newblock Software available at \url{http://www.csie.ntu.edu.tw/~cjlin/libsvm}
\bibAnnoteFile{CC01a}

\bibitem[{Dai et~al.(2019)Dai, Yu, Low, and Jaillet}]{pmlr-v97-dai19a}
Dai, Z., Yu, H., Low, B. K.~H., and Jaillet, P. (2019).
\newblock {B}ayesian optimization meets {B}ayesian optimal stopping.
\newblock In \emph{Proceedings of the 36th International Conference on Machine
  Learning}, eds. K.~Chaudhuri and R.~Salakhutdinov (Long Beach, California,
  USA: PMLR), vol.~97 of \emph{Proceedings of Machine Learning Research},
  1496--1506
\bibAnnoteFile{pmlr-v97-dai19a}

\bibitem[{Dasgupta(2011)}]{Dasgupta:2011:TFA:1959886.1960197}
Dasgupta, S. (2011).
\newblock Two faces of active learning.
\newblock \emph{Theor. Comput. Sci.} 412, 1767--1781.
\newblock \doi{10.1016/j.tcs.2010.12.054}
\bibAnnoteFile{Dasgupta:2011:TFA:1959886.1960197}

\bibitem[{Efron et~al.(2004)Efron, Hastie, Johnstone, and
  Tibshirani}]{efron2004}
Efron, B., Hastie, T., Johnstone, I., and Tibshirani, R. (2004).
\newblock Least angle regression.
\newblock \emph{Ann. Statist.} 32, 407--499.
\newblock \doi{10.1214/009053604000000067}
\bibAnnoteFile{efron2004}

\bibitem[{Falkner et~al.(2018)Falkner, Klein, and
  Hutter}]{DBLP:conf/icml/FalknerKH18}
Falkner, S., Klein, A., and Hutter, F. (2018).
\newblock Bohb: Robust and efficient hyperparameter optimization at scale.
\newblock In \emph{ICML}. 1436--1445
\bibAnnoteFile{DBLP:conf/icml/FalknerKH18}

\bibitem[{Gelbart et~al.(2014)Gelbart, Snoek, and
  Adams}]{Gelbart:2014:BOU:3020751.3020778}
Gelbart, M.~A., Snoek, J., and Adams, R.~P. (2014).
\newblock Bayesian optimization with unknown constraints.
\newblock In \emph{Proceedings of the Thirtieth Conference on Uncertainty in
  Artificial Intelligence} (Arlington, Virginia, United States: AUAI Press),
  UAI'14, 250--259
\bibAnnoteFile{Gelbart:2014:BOU:3020751.3020778}

\bibitem[{Gelfand and Mitter(1989)}]{Gelfand1989SimulatedAW}
Gelfand, S.~B. and Mitter, S.~K. (1989).
\newblock Simulated annealing with noisy or imprecise energy measurements.
\newblock \emph{Journal of Optimization Theory and Applications} 62, 49--62
\bibAnnoteFile{Gelfand1989SimulatedAW}

\bibitem[{Goodman and Flaxman(2017)}]{Goodman2017EuropeanUR}
Goodman, B. and Flaxman, S. (2017).
\newblock European union regulations on algorithmic decision-making and a
  "right to explanation".
\newblock \emph{AI Magazine} 38, 50--57
\bibAnnoteFile{Goodman2017EuropeanUR}

\bibitem[{Grill et~al.(2015)Grill, Valko, Munos, and Munos}]{NIPS2015_5721}
Grill, J.-B., Valko, M., Munos, R., and Munos, R. (2015).
\newblock Black-box optimization of noisy functions with unknown smoothness.
\newblock In \emph{Advances in Neural Information Processing Systems 28}, eds.
  C.~Cortes, N.~D. Lawrence, D.~D. Lee, M.~Sugiyama, and R.~Garnett (Curran
  Associates, Inc.). 667--675
\bibAnnoteFile{NIPS2015_5721}

\bibitem[{Gutjahr and Pflug(1996)}]{Gutjahr1996}
Gutjahr, W.~J. and Pflug, G.~C. (1996).
\newblock Simulated annealing for noisy cost functions.
\newblock \emph{Journal of Global Optimization} 8, 1--13.
\newblock \doi{10.1007/BF00229298}
\bibAnnoteFile{Gutjahr1996}

\bibitem[{Hansen and Kern(2004)}]{hansen2004ecm}
Hansen, N. and Kern, S. (2004).
\newblock Evaluating the {CMA} evolution strategy on multimodal test functions.
\newblock In \emph{Parallel Problem Solving from Nature {PPSN VIII}}, eds.
  X.~Yao et~al. (Springer), vol. 3242 of \emph{LNCS}, 282--291
\bibAnnoteFile{hansen2004ecm}

\bibitem[{Hansen and Ostermeier(2001)}]{Hansen:2001:CDS:1108839.1108843}
Hansen, N. and Ostermeier, A. (2001).
\newblock Completely derandomized self-adaptation in evolution strategies.
\newblock \emph{Evol. Comput.} 9, 159--195.
\newblock \doi{10.1162/106365601750190398}
\bibAnnoteFile{Hansen:2001:CDS:1108839.1108843}

\bibitem[{Hastie et~al.(2009)Hastie, Tibshirani, and
  Friedman}]{hastie_09_elements-of.statistical-learning}
Hastie, T., Tibshirani, R., and Friedman, J. (2009).
\newblock \emph{The Elements of Statistical Learning: Data Mining, Inference
  and Prediction} (Springer), 2 edn.
\bibAnnoteFile{hastie_09_elements-of.statistical-learning}

\bibitem[{Herman(2017)}]{DBLP:journals/corr/abs-1711-07414}
Herman, B. (2017).
\newblock The promise and peril of human evaluation for model interpretability.
\newblock Presented at NIPS 2017 Symposium on Interpretable Machine Learning.
  Available at: \url{https://arxiv.org/abs/1711.09889v3}
\bibAnnoteFile{DBLP:journals/corr/abs-1711-07414}

\bibitem[{Hern\'{a}ndez-Lobato et~al.(2016)Hern\'{a}ndez-Lobato, Gelbart,
  Adams, Hoffman, and Ghahramani}]{Hernandez-Lobato:2016:GFC:2946645.3053442}
Hern\'{a}ndez-Lobato, J.~M., Gelbart, M.~A., Adams, R.~P., Hoffman, M.~W., and
  Ghahramani, Z. (2016).
\newblock A general framework for constrained bayesian optimization using
  information-based search.
\newblock \emph{J. Mach. Learn. Res.} 17, 5549--5601
\bibAnnoteFile{Hernandez-Lobato:2016:GFC:2946645.3053442}

\bibitem[{Hutter et~al.(2011)Hutter, Hoos, and
  Leyton-Brown}]{Hutter:2011:SMO:2177360.2177404}
Hutter, F., Hoos, H.~H., and Leyton-Brown, K. (2011).
\newblock Sequential model-based optimization for general algorithm
  configuration.
\newblock In \emph{Proceedings of the 5th International Conference on Learning
  and Intelligent Optimization} (Berlin, Heidelberg: Springer-Verlag), LION'05,
  507--523.
\newblock \doi{10.1007/978-3-642-25566-3_40}
\bibAnnoteFile{Hutter:2011:SMO:2177360.2177404}

\bibitem[{Japkowicz and Stephen(2002)}]{Japkowicz:2002:CIP:1293951.1293954}
Japkowicz, N. and Stephen, S. (2002).
\newblock The class imbalance problem: A systematic study.
\newblock \emph{Intell. Data Anal.} 6, 429--449
\bibAnnoteFile{Japkowicz:2002:CIP:1293951.1293954}

\bibitem[{Jeh and Widom(2002)}]{Jeh:2002:SMS:775047.775126}
Jeh, G. and Widom, J. (2002).
\newblock Simrank: A measure of structural-context similarity.
\newblock In \emph{Proceedings of the Eighth ACM SIGKDD International
  Conference on Knowledge Discovery and Data Mining} (New York, NY, USA: ACM),
  KDD '02, 538--543.
\newblock \doi{10.1145/775047.775126}
\bibAnnoteFile{Jeh:2002:SMS:775047.775126}

\bibitem[{Jones et~al.(2001)Jones, Oliphant, Peterson et~al.}]{scipy}
Jones, E., Oliphant, T., Peterson, P., et~al. (2001).
\newblock {SciPy}: Open source scientific tools for {Python}.
\newblock \url{http://www.scipy.org/}
\bibAnnoteFile{scipy}

\bibitem[{Jurafsky and Martin(2019)}]{jurafsky2019speech}
Jurafsky, D. and Martin, J. (2019).
\newblock Speech and language processing.
\newblock Preprint on webpage at
  \url{https://web.stanford.edu/~jurafsky/slp3/ed3book.pdf}
\bibAnnoteFile{jurafsky2019speech}

\bibitem[{Ke et~al.(2017)Ke, Meng, Finley, Wang, Chen, Ma
  et~al.}]{Ke:2017:LHE:3294996.3295074}
Ke, G., Meng, Q., Finley, T., Wang, T., Chen, W., Ma, W., et~al. (2017).
\newblock Lightgbm: A highly efficient gradient boosting decision tree.
\newblock In \emph{Proceedings of the 31st International Conference on Neural
  Information Processing Systems} (USA: Curran Associates Inc.), NIPS'17,
  3149--3157
\bibAnnoteFile{Ke:2017:LHE:3294996.3295074}

\bibitem[{{Kennedy} and {Eberhart}(1995)}]{PSO}
{Kennedy}, J. and {Eberhart}, R. (1995).
\newblock Particle swarm optimization.
\newblock In \emph{Proceedings of ICNN'95 - International Conference on Neural
  Networks}. vol.~4, 1942--1948 vol.4.
\newblock \doi{10.1109/ICNN.1995.488968}
\bibAnnoteFile{PSO}

\bibitem[{Kirkpatrick et~al.(1983)Kirkpatrick, Gelatt, and
  Vecchi}]{Kirkpatrick1983}
Kirkpatrick, S., Gelatt, C.~D., and Vecchi, M.~P. (1983).
\newblock Optimization by simulated annealing.
\newblock \emph{Science} 220, 671--680.
\newblock \doi{10.1126/science.220.4598.671}
\bibAnnoteFile{Kirkpatrick1983}

\bibitem[{Koh and Liang(2017)}]{pmlr-v70-koh17a}
Koh, P.~W. and Liang, P. (2017).
\newblock Understanding black-box predictions via influence functions.
\newblock In \emph{Proceedings of the 34th International Conference on Machine
  Learning}, eds. D.~Precup and Y.~W. Teh (International Convention Centre,
  Sydney, Australia: PMLR), vol.~70 of \emph{Proceedings of Machine Learning
  Research}, 1885--1894
\bibAnnoteFile{pmlr-v70-koh17a}

\bibitem[{Kumaraswamy(1980)}]{KUMARASWAMY198079}
Kumaraswamy, P. (1980).
\newblock A generalized probability density function for double-bounded random
  processes.
\newblock \emph{Journal of Hydrology} 46, 79 -- 88.
\newblock \doi{https://doi.org/10.1016/0022-1694(80)90036-0}
\bibAnnoteFile{KUMARASWAMY198079}

\bibitem[{Lakkaraju et~al.(2016)Lakkaraju, Bach, and
  Leskovec}]{Lakkaraju:2016:IDS:2939672.2939874}
Lakkaraju, H., Bach, S.~H., and Leskovec, J. (2016).
\newblock Interpretable decision sets: A joint framework for description and
  prediction.
\newblock In \emph{Proceedings of the 22Nd ACM SIGKDD International Conference
  on Knowledge Discovery and Data Mining} (New York, NY, USA: ACM), KDD '16,
  1675--1684.
\newblock \doi{10.1145/2939672.2939874}
\bibAnnoteFile{Lakkaraju:2016:IDS:2939672.2939874}

\bibitem[{Letham et~al.(2017)Letham, Karrer, Ottoni, and Bakshy}]{BO_noisy}
Letham, B., Karrer, B., Ottoni, G., and Bakshy, E. (2017).
\newblock Constrained bayesian optimization with noisy experiments.
\newblock \emph{Bayesian Analysis} \doi{10.1214/18-BA1110}
\bibAnnoteFile{BO_noisy}

\bibitem[{Letham et~al.(2013)Letham, Rudin, McCormick, and
  Madigan}]{Letham2013InterpretableCU}
Letham, B., Rudin, C., McCormick, T.~H., and Madigan, D. (2013).
\newblock Interpretable classifiers using rules and bayesian analysis: Building
  a better stroke prediction model.
\newblock \emph{CoRR} abs/1511.01644
\bibAnnoteFile{Letham2013InterpretableCU}

\bibitem[{Levesque et~al.(2017)Levesque, Durand, Gagn{\'e}, and
  Sabourin}]{Levesque2017BayesianOF}
Levesque, J.-C., Durand, A., Gagn{\'e}, C., and Sabourin, R. (2017).
\newblock Bayesian optimization for conditional hyperparameter spaces.
\newblock \emph{2017 International Joint Conference on Neural Networks (IJCNN)}
  , 286--293
\bibAnnoteFile{Levesque2017BayesianOF}

\bibitem[{Li et~al.(2017{\natexlab{a}})Li, Gupta, Rana, Nguyen, Venkatesh, and
  Shilton}]{ijcai2017-291}
Li, C., Gupta, S., Rana, S., Nguyen, V., Venkatesh, S., and Shilton, A.
  (2017{\natexlab{a}}).
\newblock High dimensional bayesian optimization using dropout.
\newblock In \emph{Proceedings of the Twenty-Sixth International Joint
  Conference on Artificial Intelligence, {IJCAI-17}}. 2096--2102.
\newblock \doi{10.24963/ijcai.2017/291}
\bibAnnoteFile{ijcai2017-291}

\bibitem[{Li et~al.(2017{\natexlab{b}})Li, Jamieson, DeSalvo, Rostamizadeh, and
  Talwalkar}]{Li:2017:HNB:3122009.3242042}
Li, L., Jamieson, K., DeSalvo, G., Rostamizadeh, A., and Talwalkar, A.
  (2017{\natexlab{b}}).
\newblock Hyperband: A novel bandit-based approach to hyperparameter
  optimization.
\newblock \emph{J. Mach. Learn. Res.} 18, 6765--6816
\bibAnnoteFile{Li:2017:HNB:3122009.3242042}

\bibitem[{Lim and Hastie(2015)}]{Lim2015}
Lim, M. and Hastie, T. (2015).
\newblock Learning interactions via hierarchical group-lasso regularization.
\newblock \emph{J Comput Graph Stat} 24, 627--654.
\newblock \doi{10.1080/10618600.2014.938812}.
\newblock 26759522[pmid]
\bibAnnoteFile{Lim2015}

\bibitem[{Lipton(2018)}]{Lipton:2018:MMI:3236386.3241340}
Lipton, Z.~C. (2018).
\newblock The mythos of model interpretability.
\newblock \emph{Queue} 16, 30:31--30:57.
\newblock \doi{10.1145/3236386.3241340}
\bibAnnoteFile{Lipton:2018:MMI:3236386.3241340}

\bibitem[{Lou et~al.(2013)Lou, Caruana, Gehrke, and
  Hooker}]{Lou:2013:AIM:2487575.2487579}
Lou, Y., Caruana, R., Gehrke, J., and Hooker, G. (2013).
\newblock Accurate intelligible models with pairwise interactions.
\newblock In \emph{Proceedings of the 19th ACM SIGKDD International Conference
  on Knowledge Discovery and Data Mining} (New York, NY, USA: ACM), KDD '13,
  623--631.
\newblock \doi{10.1145/2487575.2487579}
\bibAnnoteFile{Lou:2013:AIM:2487575.2487579}

\bibitem[{Lundberg and Lee(2017)}]{NIPS2017_7062}
Lundberg, S.~M. and Lee, S.-I. (2017).
\newblock A unified approach to interpreting model predictions.
\newblock In \emph{Advances in Neural Information Processing Systems 30}, eds.
  I.~Guyon, U.~V. Luxburg, S.~Bengio, H.~Wallach, R.~Fergus, S.~Vishwanathan,
  and R.~Garnett (Curran Associates, Inc.). 4765--4774
\bibAnnoteFile{NIPS2017_7062}

\bibitem[{Malkomes and Garnett(2018)}]{abo_NIPS2018_7838}
Malkomes, G. and Garnett, R. (2018).
\newblock Automating bayesian optimization with bayesian optimization.
\newblock In \emph{Advances in Neural Information Processing Systems 31}, eds.
  S.~Bengio, H.~Wallach, H.~Larochelle, K.~Grauman, N.~Cesa-Bianchi, and
  R.~Garnett (Curran Associates, Inc.). 5984--5994
\bibAnnoteFile{abo_NIPS2018_7838}

\bibitem[{Mood(2010)}]{Mood291880}
Mood, C. (2010).
\newblock Logistic regression : Why we cannot do what we think we can do, and
  what we can do about it.
\newblock \emph{European Sociological Review} 26, 67--82.
\newblock \doi{10.1093/esr/jcp006}
\bibAnnoteFile{Mood291880}

\bibitem[{Munteanu and Schwiegelshohn(2018)}]{Munteanu2018}
Munteanu, A. and Schwiegelshohn, C. (2018).
\newblock Coresets-methods and history: A theoreticians design pattern for
  approximation and streaming algorithms.
\newblock \emph{KI - K{\"u}nstliche Intelligenz} 32, 37--53.
\newblock \doi{10.1007/s13218-017-0519-3}
\bibAnnoteFile{Munteanu2018}

\bibitem[{Nayebi et~al.(2019)Nayebi, Munteanu, and
  Poloczek}]{pmlr-v97-nayebi19a}
Nayebi, A., Munteanu, A., and Poloczek, M. (2019).
\newblock A framework for {B}ayesian optimization in embedded subspaces.
\newblock In \emph{Proceedings of the 36th International Conference on Machine
  Learning}, eds. K.~Chaudhuri and R.~Salakhutdinov (Long Beach, California,
  USA: PMLR), vol.~97 of \emph{Proceedings of Machine Learning Research},
  4752--4761
\bibAnnoteFile{pmlr-v97-nayebi19a}

\bibitem[{Olkin and Trikalinos(2014)}]{article_alt_beta}
Olkin, I. and Trikalinos, T. (2014).
\newblock Constructions for a bivariate beta distribution.
\newblock \emph{Statistics \& Probability Letters} 96.
\newblock \doi{10.1016/j.spl.2014.09.013}
\bibAnnoteFile{article_alt_beta}

\bibitem[{Pan et~al.(2004)Pan, Yang, Faloutsos, and
  Duygulu}]{Pan:2004:AMC:1014052.1014135}
Pan, J.-Y., Yang, H.-J., Faloutsos, C., and Duygulu, P. (2004).
\newblock Automatic multimedia cross-modal correlation discovery.
\newblock In \emph{Proceedings of the Tenth ACM SIGKDD International Conference
  on Knowledge Discovery and Data Mining} (New York, NY, USA: ACM), KDD '04,
  653--658.
\newblock \doi{10.1145/1014052.1014135}
\bibAnnoteFile{Pan:2004:AMC:1014052.1014135}

\bibitem[{Parsopoulos and Vrahatis(2001)}]{Parsopoulos01particleswarm}
Parsopoulos, K.~E. and Vrahatis, M.~N. (2001).
\newblock Particle swarm optimizer in noisy and continuously changing
  environments.
\newblock In \emph{M.H. Hamza (Ed.), Arti cial Intelligence and Soft Computing,
  IASTED/ACTA} (IASTED/ACTA Press), 289--294
\bibAnnoteFile{Parsopoulos01particleswarm}

\bibitem[{Pedregosa et~al.(2011)Pedregosa, Varoquaux, Gramfort, Michel,
  Thirion, Grisel et~al.}]{scikit-learn}
Pedregosa, F., Varoquaux, G., Gramfort, A., Michel, V., Thirion, B., Grisel,
  O., et~al. (2011).
\newblock Scikit-learn: Machine learning in {P}ython.
\newblock \emph{Journal of Machine Learning Research} 12, 2825--2830
\bibAnnoteFile{scikit-learn}

\bibitem[{Perrone et~al.(2018)Perrone, Jenatton, Seeger, and
  Archambeau}]{NIPS2018_7917}
Perrone, V., Jenatton, R., Seeger, M.~W., and Archambeau, C. (2018).
\newblock Scalable hyperparameter transfer learning.
\newblock In \emph{Advances in Neural Information Processing Systems 31}, eds.
  S.~Bengio, H.~Wallach, H.~Larochelle, K.~Grauman, N.~Cesa-Bianchi, and
  R.~Garnett (Curran Associates, Inc.). 6845--6855
\bibAnnoteFile{NIPS2018_7917}

\bibitem[{Pitman and Yor(1997)}]{pitman1997}
Pitman, J. and Yor, M. (1997).
\newblock The two-parameter poisson-dirichlet distribution derived from a
  stable subordinator.
\newblock \emph{Ann. Probab.} 25, 855--900.
\newblock \doi{10.1214/aop/1024404422}
\bibAnnoteFile{pitman1997}

\bibitem[{Quinlan(1993)}]{Quinlan:1993:CPM:152181}
Quinlan, J.~R. (1993).
\newblock \emph{C4.5: Programs for Machine Learning} (San Francisco, CA, USA:
  Morgan Kaufmann Publishers Inc.)
\bibAnnoteFile{Quinlan:1993:CPM:152181}

\bibitem[{{Quinlan}(2004)}]{Quinlan:c5}
[Dataset] {Quinlan}, J.~R. (2004).
\newblock C5.0.
\newblock \url{https://rulequest.com/}
\bibAnnoteFile{Quinlan:c5}

\bibitem[{Rana et~al.(2017)Rana, Li, Gupta, Nguyen, and
  Venkatesh}]{pmlr-v70-rana17a}
Rana, S., Li, C., Gupta, S., Nguyen, V., and Venkatesh, S. (2017).
\newblock High dimensional {B}ayesian optimization with elastic {G}aussian
  process.
\newblock In \emph{Proceedings of the 34th International Conference on Machine
  Learning}, eds. D.~Precup and Y.~W. Teh (International Convention Centre,
  Sydney, Australia: PMLR), vol.~70 of \emph{Proceedings of Machine Learning
  Research}, 2883--2891
\bibAnnoteFile{pmlr-v70-rana17a}

\bibitem[{Rasmussen(1999)}]{Rasmussen:1999:IGM:3009657.3009736}
Rasmussen, C.~E. (1999).
\newblock The infinite gaussian mixture model.
\newblock In \emph{Proceedings of the 12th International Conference on Neural
  Information Processing Systems} (Cambridge, MA, USA: MIT Press), NIPS'99,
  554--560
\bibAnnoteFile{Rasmussen:1999:IGM:3009657.3009736}

\bibitem[{Ribeiro et~al.(2016)Ribeiro, Singh, and
  Guestrin}]{Ribeiro:2016:WIT:2939672.2939778}
Ribeiro, M.~T., Singh, S., and Guestrin, C. (2016).
\newblock ``why should i trust you?'': Explaining the predictions of any
  classifier.
\newblock In \emph{Proceedings of the 22Nd ACM SIGKDD International Conference
  on Knowledge Discovery and Data Mining} (New York, NY, USA: ACM), KDD '16,
  1135--1144.
\newblock \doi{10.1145/2939672.2939778}
\bibAnnoteFile{Ribeiro:2016:WIT:2939672.2939778}

\bibitem[{Ribeiro et~al.(2018)Ribeiro, Singh, and Guestrin}]{AAAI1816982}
[Dataset] Ribeiro, M.~T., Singh, S., and Guestrin, C. (2018).
\newblock Anchors: High-precision model-agnostic explanations
\bibAnnoteFile{AAAI1816982}

\bibitem[{Rodriguez et~al.(2006)Rodriguez, Kuncheva, and
  Alonso}]{Rodriguez:2006:RFN:1159167.1159358}
Rodriguez, J.~J., Kuncheva, L.~I., and Alonso, C.~J. (2006).
\newblock Rotation forest: A new classifier ensemble method.
\newblock \emph{IEEE Trans. Pattern Anal. Mach. Intell.} 28, 1619--1630.
\newblock \doi{10.1109/TPAMI.2006.211}
\bibAnnoteFile{Rodriguez:2006:RFN:1159167.1159358}

\bibitem[{{Selvaraju} et~al.(2017){Selvaraju}, {Cogswell}, {Das}, {Vedantam},
  {Parikh}, and {Batra}}]{8237336}
{Selvaraju}, R.~R., {Cogswell}, M., {Das}, A., {Vedantam}, R., {Parikh}, D.,
  and {Batra}, D. (2017).
\newblock Grad-cam: Visual explanations from deep networks via gradient-based
  localization.
\newblock In \emph{2017 IEEE International Conference on Computer Vision
  (ICCV)}. 618--626.
\newblock \doi{10.1109/ICCV.2017.74}
\bibAnnoteFile{8237336}

\bibitem[{Settles(2009)}]{settles2009active}
Settles, B. (2009).
\newblock \emph{Active Learning Literature Survey}.
\newblock Computer Sciences Technical Report 1648, University of
  Wisconsin--Madison
\bibAnnoteFile{settles2009active}

\bibitem[{{Shahriari} et~al.(2016){Shahriari}, {Swersky}, {Wang}, {Adams}, and
  {de Freitas}}]{7352306}
{Shahriari}, B., {Swersky}, K., {Wang}, Z., {Adams}, R.~P., and {de Freitas},
  N. (2016).
\newblock Taking the human out of the loop: A review of bayesian optimization.
\newblock \emph{Proceedings of the IEEE} 104, 148--175.
\newblock \doi{10.1109/JPROC.2015.2494218}
\bibAnnoteFile{7352306}

\bibitem[{Snoek et~al.(2012)Snoek, Larochelle, and Adams}]{NIPS2012_4522}
Snoek, J., Larochelle, H., and Adams, R.~P. (2012).
\newblock Practical bayesian optimization of machine learning algorithms.
\newblock In \emph{Advances in Neural Information Processing Systems 25}, eds.
  F.~Pereira, C.~J.~C. Burges, L.~Bottou, and K.~Q. Weinberger (Curran
  Associates, Inc.). 2951--2959
\bibAnnoteFile{NIPS2012_4522}

\bibitem[{Snoek et~al.(2015)Snoek, Rippel, Swersky, Kiros, Satish, Sundaram
  et~al.}]{Snoek:2015:SBO:3045118.3045349}
Snoek, J., Rippel, O., Swersky, K., Kiros, R., Satish, N., Sundaram, N., et~al.
  (2015).
\newblock Scalable bayesian optimization using deep neural networks.
\newblock In \emph{Proceedings of the 32Nd International Conference on
  International Conference on Machine Learning - Volume 37} (JMLR.org),
  ICML'15, 2171--2180
\bibAnnoteFile{Snoek:2015:SBO:3045118.3045349}

\bibitem[{Ustun and Rudin(2016)}]{Ustun2016}
Ustun, B. and Rudin, C. (2016).
\newblock Supersparse linear integer models for optimized medical scoring
  systems.
\newblock \emph{Machine Learning} 102, 349--391.
\newblock \doi{10.1007/s10994-015-5528-6}
\bibAnnoteFile{Ustun2016}

\bibitem[{Wang et~al.(2013)Wang, Zoghi, Hutter, Matheson, and
  De~Freitas}]{Wang:2013:BOH:2540128.2540383}
Wang, Z., Zoghi, M., Hutter, F., Matheson, D., and De~Freitas, N. (2013).
\newblock Bayesian optimization in high dimensions via random embeddings.
\newblock In \emph{Proceedings of the Twenty-Third International Joint
  Conference on Artificial Intelligence} (AAAI Press), IJCAI '13, 1778--1784
\bibAnnoteFile{Wang:2013:BOH:2540128.2540383}

\bibitem[{Wu and Palmer(1994)}]{Wu:1994:VSL:981732.981751}
Wu, Z. and Palmer, M. (1994).
\newblock Verbs semantics and lexical selection.
\newblock In \emph{Proceedings of the 32Nd Annual Meeting on Association for
  Computational Linguistics} (Stroudsburg, PA, USA: Association for
  Computational Linguistics), ACL '94, 133--138.
\newblock \doi{10.3115/981732.981751}
\bibAnnoteFile{Wu:1994:VSL:981732.981751}

\end{thebibliography}

\end{document}